\documentclass[11pt]{article}

\RequirePackage{amsthm,amsmath,amsfonts,amssymb,bbm}
\RequirePackage[numbers]{natbib}
\RequirePackage[colorlinks,citecolor=blue,urlcolor=blue]{hyperref}
\RequirePackage{graphicx}
\usepackage{enumerate}
\usepackage{enumitem}
\usepackage{caption}
\usepackage{subfigure}
\usepackage{booktabs}
\usepackage{multirow}
\usepackage{xcolor}
\usepackage[ruled,vlined]{algorithm2e}

\theoremstyle{plain}

\newtheorem{theorem}{Theorem}

\theoremstyle{remark}

\newtheorem{assumption}{Assumption}

\newtheorem{remark}{Remark}

\renewcommand{\hat}{\widehat}

\newcommand{\R}{\mathbb{R}}
\newcommand{\E}{\mathbb{E}}
\def\P{\mathbb{P}}
\newcommand{\onestar}{$\star$}
\newcommand{\twostar}{$\star\star$}
\newcommand{\threestar}{$\star\star\star$}

\long\def\revise#1{#1}

\newcommand*\samethanks[1][\value{footnote}]{\footnotemark[#1]}

\title{High Confidence Level Inference is Almost Free using Parallel Stochastic Optimization}
\author{Wanrong Zhu\thanks{Department of Statistics, University of California, Irvine}, Zhipeng Lou\thanks{Department of Mathematics, University of California, San Diego}, Ziyang Wei\thanks{Department of Statistics, University of Chicago}, and Wei Biao Wu\samethanks[3]}
\date{}

\begin{document}

\maketitle

\begin{abstract}
	Uncertainty quantification for stochastic approximation (SA) solutions has gained popularity recently. 
This paper introduces a novel inference method focused on constructing high level confidence intervals using SA solutions in an online setting. 
Specifically, we propose to use a small number of independent multi-runs to acquire distribution information and construct a $t$-based confidence interval.  
Our method requires minimal additional computation and memory beyond the standard updating of SA solutions, making the inference process almost cost-free.
We provide a rigorous theoretical guarantee for the confidence interval, demonstrating that the coverage is approximately exact with an explicit convergence rate and allowing for  high confidence level inference.
In particular, a new Gaussian approximation result is developed for the online estimators to characterize the coverage properties of our confidence intervals in terms of relative errors. 
Additionally, our method also allows for leveraging parallel computing to further accelerate calculations using multiple cores. It is easy to implement and can be integrated with existing stochastic algorithms without the need for complicated modifications.
\end{abstract}

\section{Introduction} 

Consider the statistical inference problem for model parameters where the true model parameter  $x^{*}\in \R^{d}$ can be characterized as the minimizer of an objective function  $F(x)$ from $\R^{d}$ to $\R$, i.e,
\begin{equation}\label{eq:obj}
 	x^{*} =  \arg\min_{x\in \R^{d}} \ F(x).
\end{equation}
The objective function $F(x)$ is defined as $F(x) = \E_{\xi \sim \Pi}{f(x, \xi)}$, where $f(x, \xi)$ represents a noisy measurement of $F(x)$, and $\xi$ is a random variable following the distribution $\Pi$.  

As modern datasets grow increasingly large and are often collected in an online manner, classical deterministic optimization which requires storing all data is becoming less appealing or even infeasible for such problems. One solution is to employ stochastic approximation (SA), updating the estimate of the minimizer based on the stochastic gradient/subgradient. This approach is characterized by low memory and computational efficiency, making it suitable for an online setting with sequential data and decision-making.
One of the most popular algorithms is the stochastic gradient descent (SGD), also known as the Robbins-Monro algorithm \citep{robbins1951stochastic}.
A diverse range of variants have been developed to accelerate convergence or reduce variance in different scenarios \citep{ruppert1988efficient, polyak1992acceleration, duchi2011adaptive, rakhlin2011making, kingma2014adam, toulis2017asymptotic, woodworth2020local}. Beyond convergence analysis of the final (point) estimate, there is a growing emphasis on uncertainty quantification and statistical inference within the context of SA algorithms, which forms a significant part of recent research endeavors \citep{Chen2020AoS, fang2018online, lee2022fast, su2023higrad, wei2025online, wei2026refining}.

With streaming data $\{\xi_{i}\}_{i\ge1}$, and assuming we obtain iterates/outputs of an SA algorithm,  the primary goal of this paper is to enhance statistical inference by constructing confidence intervals based on these iterates in an online setting. Specifically, for a given vector $\upsilon \in \mathbb{R}^{d}$, we aim to construct a valid $(1 - \alpha) \times 100\%$ confidence interval $\hat{\textnormal{CI}}$ for the linear functional $\upsilon^{\top} x^{*}$, that is 
\begin{equation}\label{directdiff}
	\mathbb{P}(\upsilon^{\top} x^{*} \in \hat{\textnormal{CI}}) - (1 - \alpha) 
	= \alpha - \mathbb{P}(\upsilon^{\top} x^{*} \not \in \hat{\textnormal{CI}})
	\approx 0,
\end{equation}
where $\alpha\in(0,1)$.  To fit in an online setting, the proposed confidence interval can be updated recursively as new data becomes available. It utilizes only previous SA iterates, requiring minimal extra computation for the inference purpose beyond the original computation, thus allowing for easy integration into existing codebases.

In this paper, we consider a \emph{high level of confidence}, i.e., $\alpha\approx 0$, as uncertainty quantification is particularly important in applications involving high-stakes decisions, where a nearly 100\% confidence interval is required. Moreover, with datasets growing increasingly large, the demand for higher-level confidence intervals becomes more prevalent. Additionally, in applications involving multiple simultaneous tests, such as high-dimensional parameter analysis, correction techniques like the Bonferroni method are employed. This leads to each individual test maintaining a sufficiently high confidence level (related to dimension).  In such cases, the guarantee in \eqref{directdiff} may not be sufficient.
In particular, we shall construct confidence intervals $\hat{\textnormal{CI}}$ such that the relative error
\begin{equation}\label{ratiodiff}
	\Delta_{\alpha}:= \left|\frac{ \mathbb{P}(\upsilon^{\top} x^{*} \in \hat{\textnormal{CI}}) - (1 - \alpha)}{\alpha} \right|
	= \left|\frac{ \mathbb{P}(\upsilon^{\top} x^{*} \not \in \hat{\textnormal{CI}})}{\alpha} - 1 \right|
\end{equation}
is small. Note that (\ref{ratiodiff}) offers a much more refined assessment than   (\ref{directdiff}). For example, if $\alpha = 10^{-4}$ and $\mathbb{P}(\upsilon^{\top} x^{*} \not \in \hat{\textnormal{CI}}) = 10^{-3}$, then (\ref{directdiff}) is not severely violated, while $\Delta_{\alpha}$ in (\ref{ratiodiff}) is very different from $0$. 
In this context, it is crucial to recognize that even a slight undercoverage can be significant due to low tolerance for error. Conversely, an extremely wide confidence interval that nearly always covers can become uninformative, underscoring the importance of precision in interval construction. Hence, employing a method that provides confidence intervals with rapid convergence to the desired coverage level is essential. In Section \ref{sec:thm},  we derive the upper bound (with explicit rate) of the relative error of coverage  for the constructed confidence intervals and explicitly detail the dependence on $\alpha$ in the upper bound. The results indicate that our method remains valid even when $\alpha$ is potentially very small or decreases with the total sample size or the number of hypotheses.

\subsection{Background: existing confidence interval construction} 
Practical inference methods are based on the limiting distribution of SA solutions. Consider the vanilla SGD iterates with the recursion form:
\[ 
	x_{i} = x_{i - 1} - \eta_{i} \nabla f(x_{i - 1}, \xi_{i}), \quad i = 1, 2, \ldots, 
\]
where $\nabla{f} (x, \xi)$ is the gradient vector of $f(x, \xi)$ with respect to the first variable $x$, and $\eta_{i}$ is the step size at the $i$-th step.     In the celebrated work of \cite{polyak1992acceleration}, it is shown that, under suitable conditions, the averaged SGD (ASGD), $\bar{x}_{n} = n^{-1} \sum_{i = 1}^{n} x_{i}$, exhibits   asymptotic normality, that is, 
\begin{equation}\label{eq_MCLT_SGD}
	\sqrt{n}(\bar{x}_{n}-x^{*})\Rightarrow \mathcal{N}(0, \Sigma), 
\end{equation}
where $\Sigma =  A^{-1}SA^{-1}$ is the sandwich form covariance matrix with $A = \nabla^{2}F(x^{*})$ and $S = \mathbb{E}\left([\nabla f(x^{*}, \xi)][\nabla f(x^{*}, \xi)]^{T}\right)$.
Note that the asymptotic normality result is the same as that for offline M-estimators \citep{van2000asymptotic}, and achieves asymptotic minimax optimality \citep{hajek1972local, duchi2021asymptotic}. Similar asymptotic normality results have been established for other variants of SGD with   adjusted asymptotic covariance matrices \citep{li2022root, wei2026general, na2025asymptotic}.

These asymptotic normality results form the foundation of statistical inference in an online setting. As the limiting covariance matrix is unknown in practice, to perform practical inference, there are three primary methods for constructing confidence intervals.

\begin{itemize}
	\item The first method relies on recursively estimating the limiting covariance matrix $\Sigma$. \cite{Chen2020AoS} proposes the plug-in method to estimate  $A$ and  $S$ separately using sample averages and then applying them in the sandwich form.  \cite{zhu2023online} proposed the online batch-means method, which only utilizes SGD iterates and is more computationally efficient. Both methods provide consistent estimators for the asymptotic covariance matrix $\Sigma$ of ASGD solutions. 
	With a consistent covariance estimate $\hat{\Sigma}_{n}$,  one can construct confidence intervals for $\upsilon^{\top} x^{*}$  as
$$
		\hat{\textnormal{CI}}_{n, \textnormal{cov}} = \left[\upsilon^{\top} \bar{x}_{n} -z_{1-\alpha/2}\sqrt{\frac{\upsilon^{\top} \hat{\Sigma}_{n}\upsilon}{n}},\  \upsilon^{\top} \bar{x}_{n}+z_{1-\alpha/2}\sqrt{\frac{\upsilon^{\top} \hat{\Sigma}_{n}\upsilon}{n}}\right],
$$
	where $z_{1-\alpha/2}$ is the $(1-\alpha/2)\times100\%$ quantile of the standard normal distribution.
	\item The second method takes advantage of statistical pivotal statistics. One example is the random scaling method. Instead of consistently estimating the asymptotic covariance matrix, \cite{lee2022fast} leverages the asymptotic normality result by constructing asymptotic pivotal statistics after self-normalization. Specifically, they  studentize $\sqrt{n}(\bar{x}_n-x^*)$ via  the random scaling matrix
	\[\hat V_{rs, n} = \frac{1}{n}\sum_{s=1}^{n}\left\{\frac{1}{\sqrt{n}}\sum_{i=1}^{s}(x_{i} - \bar{x}_{n})\right\}\left\{\frac{1}{\sqrt{n}}\sum_{i=1}^{s}(x_{i} - \bar{x}_{n})\right\}^{T}.\]
	The resulting statistic is  asymptotically pivotal and the confidence interval for $\upsilon^{\top} x^{*}$  is then constructed as
	\begin{equation}\label{eqn:rs}
		\hat{\textnormal{CI}}_{n, \textnormal{rs}} = \left[\upsilon^{\top}\bar{x}_{n} - q_{rs, 1-\alpha/2}\sqrt{\frac{\upsilon^{\top}\hat V_{rs,n}\upsilon}{n}},\ \upsilon^{\top}\bar{x}_{n} + q_{rs, 1-\alpha/2}\sqrt{\frac{\upsilon^{\top}\hat V_{rs,n}\upsilon}{n}}\right],
	\end{equation}
	where $q_{rs, 1-\alpha/2}$ is the $(1-\alpha/2)\times 100\%$ percentile for  $W_{1}(1)/[\int_{0}^{1}\{W_{1}(r) - rW_{1}(1)\}^2dr]^{1/2}$ with $W_{1}(r)$ stands for a standard Brownian motion.
	\item An alternative method for inference is via bootstrap. One can apply bootstrap perturbations and modify the original SGD path. Then, the asymptotic distribution of the online estimate as well as other quantities such as variance or quantiles can be estimated using a large number of bootstrapped sequences \citep{fang2018online, li2018statistical}.
\end{itemize} 
Similar ideas have also been applied for inference when using different algorithms or dealing with online decision-making problems~\citep{luo2022covariance, li2022statistical, chen2021statistical, ramprasad2022online, su2023higrad}. Note that all the three methods above have their advantages and applicable use cases. The first and the third methods can provide consistent estimators of the limiting covariance matrix. However, the cost of using bootstrap (the third method) involves heavy computation or complicated modification to existing code base, and we will not consider this method. The online covariance matrix estimation (the second method) is a difficult task in SGD settings. The plug-in estimator requires Hessian information which is typically unavailable, and involves matrix computation that requires an $O(d^3)$ computational cost, which is not desirable for large dimensions. The online batch-means methods do not require extra information such as the Hessian and are computationally efficient, but they come at the cost of slow convergence. 
The random scaling method does not provide a consistent covariance matrix estimator, yet in terms of confidence interval construction, it is computationally comparable to the online batch-means method while offering better coverage. However, the critical values of the self-normalized statistics are not easy to obtain for arbitrary $\alpha$,  we simulate the value via MCMC in Section \ref{sec:exp}.  

In terms of theoretical guarantees, although all three methods demonstrate asymptotically valid coverage of confidence intervals, the convergence guarantee without a specific rate and without explicit dependence on $\alpha$ is not sufficient, as demonstrated above and in Section \ref{sec:thm}. This limitation is not significant at moderate confidence levels but becomes substantial at higher levels, leading to unstable coverage. This may result in either undercoverage (failing to meet the standard) or overcoverage (producing an excessively wide confidence interval, thereby diminishing the interval's meaningfulness). A detailed comparison and discussion of these methods can be found in \cite{lee2022fast}. We also make a brief summary in Table \ref{table:methods_comparison} comparing the above online inference methods.

\begin{table}[ht]
	\centering 
    \begin{tabular}{lcccc}
\hline
Method & Plug-In & Online BM & Random Scale & This paper \\ 
\hline
Consistent covariance estimator? & $\checkmark$ & $\checkmark$ & $\times$ & $\times$ \\
Avoid Hessian? & $\times$ & $\checkmark$ & $\checkmark$ & $\checkmark$ \\
CI coverage convergence rate? & $\times$ & $\times$ & $\times$ & $\checkmark$ \\
Empirical CI coverage & \threestar & \onestar & \twostar & \threestar \\
Computation time & \onestar & \twostar & \twostar & \threestar \\ \hline
\end{tabular}
\caption{Comparison of methods for online statistical inference: Plug-In \citep{Chen2020AoS}, Online BM \citep{zhu2023online}, and Random Scale \citep{lee2022fast}. Symbols denote: $\checkmark$ for yes, $\times$ for no. More stars indicate better performance/efficiency.}
\label{table:methods_comparison}
\end{table}

\subsection{Contribution}
We propose a new inference framework for stochastic optimization algorithms based on a small number of parallel runs. The key idea is to treat parallel trajectories of stochastic approximation as approximately independent replicates and construct a $t$-based confidence interval using their empirical variability. Our main contributions are summarized as follows.
\begin{itemize}[leftmargin=*,itemsep=0pt,topsep=0pt]
\revise{
\item \textbf{Rigorous theoretical guarantees.}
We prove that the proposed parallel-run inference procedure achieves asymptotically exact coverage and derive explicit convergence rates for the relative coverage error, yielding  theoretical guarantees in high-confidence regimes that are not provided in existing inference approaches for stochastic optimization.

\item \textbf{Algorithm-agnostic and computationally efficient.}
The method applies to any stochastic optimization algorithm whose iterates satisfy an asymptotic normality property and is easy to implement.  Inference can be performed only at selected checkpoints with negligible computational or memory overhead relative to standard SGD.

\item \textbf{Empirical performance.}
Experiments in Section~\ref{sec:exp} show that the proposed method achieves more reliable coverage than competing approaches while remaining computationally efficient.

\item \textbf{Compatibility with parallel computing.}
The method naturally fits modern large-scale and federated learning systems where data are distributed across different clients \citep{zinkevich2010parallelized, dean2012large, li2020federated, karimireddy2020scaffold, mcmahan2017communication, ghosh2020efficient}. In our work, the requirement for parallel processing is seen not as a burden but as a beneficial tool.
}
\end{itemize}


\section{Inference with parallel runs of stochastic algorithms}
In this section, we introduce the parallel run inference method for constructing confidence intervals. The method involves  $K$ parallel runs of a predetermined stochastic algorithm, calculating the sample variance of the linear functional of interest from $K$ parallel runs, and self-normalizing to obtain asymptotic pivotal  $t$-statistics and the corresponding confidence interval.

\subsection{Parallel computing}
Consider a general stochastic algorithm characterized by the update rule $h_{i}$ at the $i$-th step and $K$ parallel run sequences. For  the $k$-th sequence where $k = 1, \ldots, K$, we begin with a random initialization $\hat{x}_{0}^{(k)}$. The estimate for the $k$-th machine at the $i$-th iterate is denoted by $\hat{x}_{i}^{(k)}$. The recursive update is given by
\begin{equation}
	\label{eqn:single_path}
	\hat{x}_{i}^{(k)} = h_{i}(\xi_{i}^{(k)}, \mathcal{F}_{i-1}^{(k)}), \quad i = 1, 2, \ldots,
\end{equation}
where $\mathcal{F}_{i-1}^{(k)} = \sigma (\xi_{i - 1}^{(k)}, \xi_{i - 2}^{(k)}, \ldots)$ encapsulates information from previous steps, such as $\hat{x}_{i-1}^{(k)}$ or other intermediate estimates according to the algorithm. For example, in the case of ASGD, we have
\begin{equation}\label{eqn:asgd}
	\begin{cases}
		x_{i}^{(k)} = x_{i-1}^{(k)}  - \eta_{i}\nabla f(x_{i-1}^{(k)}, \xi_{i}^{(k)} ),\\ 
		\hat{x}_{i}^{(k)} = \{(i - 1) \hat{x}_{i - 1}^{(k)} + x_{i}^{(k)}\} / i, 
	\end{cases}
\end{equation} 
where $\nabla f(x_{i-1}^{(k)}, \xi_{i}^{(k)} )$ is the derivative of the objective function $f$ with respect to the first variable, and step size is usually chosen as $\eta_{i} = \eta \times i^{-\beta}$ for some $\beta\in (0.5, 1]$. 

If we seek output for estimation or inference at the $n$-th step (with $N = nK$ total samples), we can aggregate the results from the $K$ machines by averaging the estimates or predictions. Specifically, we define the parallel average of the $K$ sequences as
\begin{equation}
	\label{eqn:parallel_average}
	\bar{x}_{K, n} = \frac{1}{K}\sum_{k=1}^{K}\hat{x}_{n}^{(k)}.
\end{equation}

In a practical online setting, sequential data $\{\xi_{i}\}_{i=1, 2, \ldots}$ can be distributed across $K$ different machines, with $\xi_{i}^{(k)} = \xi_{k+K(i-1)}$. Alternatively, in an offline setting, the dataset can be randomly divided into $K$ batches. Note that when the initialization $\hat{x}_{0}^{(k)}$ is the same for all $k = 1, \ldots, K$, the output from $K$ sequences $\hat{x}_{i}^{(k)} $, $k= 1, \ldots, K$, will be independent and identically distributed (i.i.d.), given that the data components $\xi_{i}^{(k)}$ are i.i.d.

Note that, unlike local/federated SGD as discussed in~\cite{yu2019parallel, li2022statistical, woodworth2020local}, we do not communicate local solutions/gradients to obtain a common averaged iterate for parallel runs at intermediate iterations. Our method is more akin to model averaging, often referred to as one-shot averaging \citep{zinkevich2010parallelized}.
On one hand, model averaging without communication costs can still achieve good convergence when $K$ is small or moderate. On the other hand, it ensures the $K$ sequences are i.i.d.~and enables us to construct an asymptotically pivotal $t$-statistic and demonstrate strong convergence in a later section. Additionally, the straightforward parallel running and model averaging make it easier to apply to different stochastic algorithms. In some cases, local updates with more frequent periodic averaging would improve the statistical efficiency of the algorithm and communication costs may not be a problem. The inference procedure may still hold with refined proof. However, discussing the difference between vanilla SGD, parallel SGD and local SGD is beyond the scope of this paper. 

\subsection{Asymptotic $t$-distribution}  
In this context, various stochastic approximation algorithms can be employed to run parallel sequences. To derive a valid $t$-distribution, it is essential to consider cases where asymptotic normality is applicable to the estimate $\hat{x}_{n}^{(k)}$ in each sequence. Specifically, for each $k = 1, \ldots, K$, 
\[  
	\sqrt{n} (\hat{x}_{n}^{(k)} - x^{*}) \Rightarrow \mathcal{N}(0, \Sigma), \enspace \mathrm{as} \enspace n \to \infty. 
\]
In the case of ASGD as denoted in \eqref{eqn:asgd}, the celebrated work of  \cite{polyak1992acceleration} demonstrated the asymptotic normality with the sandwich form $\Sigma$ as mentioned before. Other algorithms, such as various versions of weighted-averaged SGD \citep{wei2026general}, Root-SGD \citep{li2022root}, and StoSQP \citep{na2025asymptotic}, have also been shown to possess this asymptotic normality property, albeit with adjusted limiting covariance matrices. 

\begin{figure} 
	\centering
\includegraphics[width=0.7\textwidth]{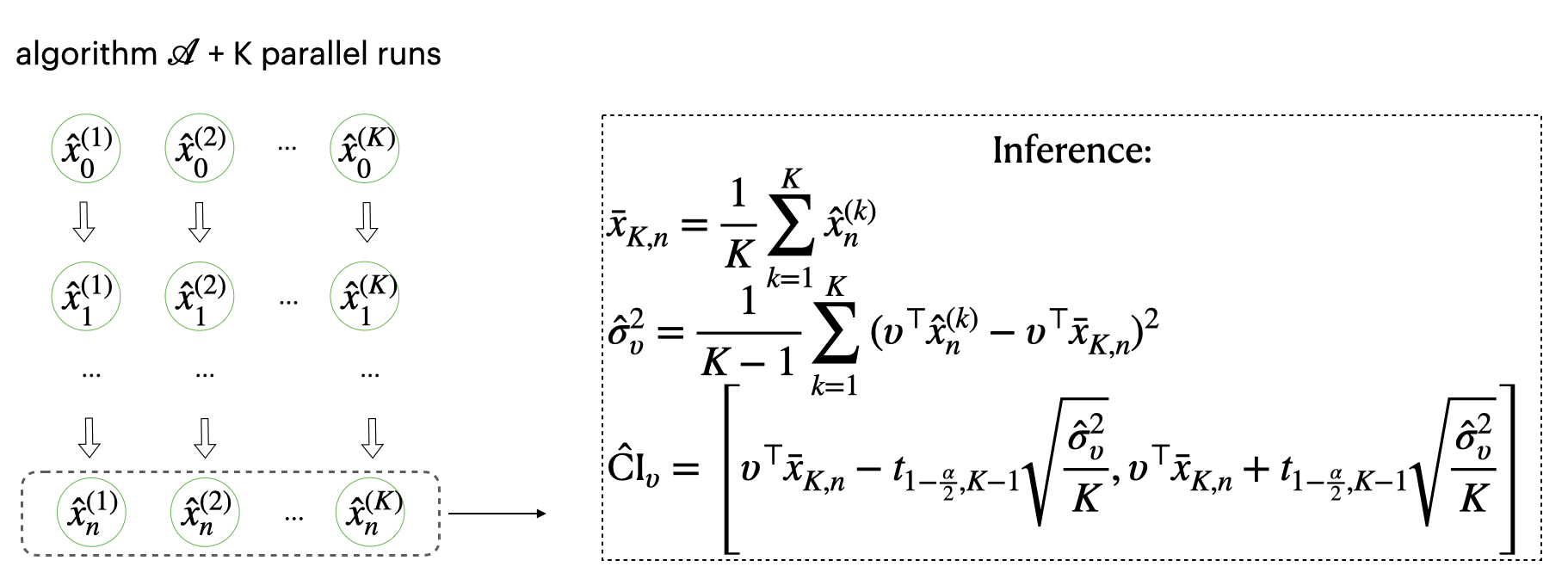}
	\caption{Realizations of parallel computing and inference.}
	\label{fig:illustration2} 
\end{figure}

For any vector $\upsilon\in\R^{d}$, considering inference for the linear functional $\upsilon^{\top}x^{*}$ at the $n$-th iteration (with $N = nK$ total samples), define the sample variance $\hat{\sigma}_{\upsilon}^{2}$ as
\[	\hat{\sigma}_{\upsilon}^{2} = \frac{1}{K - 1}\sum_{k = 1}^{K} (\upsilon^{\top} \hat{x}_{n}^{(k)} - \upsilon^{\top} \bar{x}_{K, n})^{2},\]
where $\bar{x}_{K, n} $ is the sample average defined in \eqref{eqn:parallel_average}.  It is worth noting that 
$\hat{\sigma}_{\upsilon}^{2}$ is not a consistent estimator for the variance of $\upsilon^{\top} x^{*}$. However, we can studentize $\sqrt{K}(\upsilon^{\top} \bar{x}_{K, n} - \upsilon^{\top} x^{*})$ with $\hat{\sigma}_{\upsilon}$ to obtain a  $t$-statistic which is asymptotically pivotal.  Assuming the validity of the asymptotic normality result, together with the i.i.d.~property of $\{\hat{x}_{n}^{(k)}\}_{k = 1, \ldots, K}$, we can derive a $t$-type distribution, that is,
\begin{equation}
	\label{eqn:t_v}
	\hat{t}_{\upsilon} := \frac{\sqrt{K}(\upsilon^{\top} \bar{x}_{K, n} - \upsilon^{\top} x^{*})}{\hat{\sigma}_{\upsilon}} \Rightarrow t_{K - 1}.
\end{equation}  
Based on~\eqref{eqn:t_v}, we can construct a $(1-\alpha)\times 100\%$ confidence interval for $\upsilon^{\top} x^{*}$ as follows, 
\begin{equation}
	\label{eq_CI_ASGD}
	\hat{\textnormal{CI}}_{\upsilon}  = \left[\upsilon^{\top} \bar{x}_{K, n} - \frac{t_{1-\alpha/2,K-1} \hat{\sigma}_{\upsilon}}{\sqrt{K}}, \enspace \upsilon^{\top}\bar{x}_{K, n} + \frac{t_{1-\alpha/2,K-1} \hat{\sigma}_{\upsilon}}{\sqrt{K}}\right],
\end{equation}
where $t_{1-\alpha/2,K-1}$ is the $(1-\alpha/2)\times100\%$ percentile for the $t_{K-1}$ distribution. 
The proposed confidence interval in~\eqref{eq_CI_ASGD} is fairly easy and efficient to construct. In particular, when $K = 2$, $\hat{t}_{\upsilon}$ is asymptotically distributed as the standard Cauchy distribution with upper quantile $t_{1 - \alpha/2, 1} = \tan ((1 - \alpha) \pi /2)$. The entire procedure is summarized in Algorithm \ref{alg:parallel}. See Figure \ref{fig:illustration2} for an illustration. 

\begin{algorithm}[H]
	\SetAlgoLined
	\caption{Online Parallel Inference}
	\label{alg:parallel}
	Input: stochastic algorithm $h$, number of parallel runs $K$ \\
	\For{$i = 1, 2, \ldots$}{
		\For{$k = 1, \ldots, K$}{
			\textnormal{Update } 	$\hat{x}_{i}^{(k)} = h_{i}(\xi_{i}^{(k)}, \mathcal{F}_{i-1}^{(k)}) $
			($\xi_{i}^{(k)}$ is data received)\;
		}
		\textnormal{Output if necessary}:\\
		$\bar{x}_{K, i} \leftarrow \frac{1}{K} \sum_{k=1}^{K} \hat{x}_{k,i}$\;
		$\hat{\sigma}_{\upsilon}^{2} \leftarrow \frac{1}{K - 1} \sum_{k = 1}^{K} \left(\upsilon^{\top} \hat{x}_{k, i} - \upsilon^{\top} \bar{x}_{K, i}\right)^{2}$\;
		$\hat{\textnormal{CI}}_{\upsilon} \leftarrow \left[ \upsilon^{\top} \bar{x}_{K, i} - \hat{\sigma}_{\upsilon} t_{1-\alpha/2,K-1} / \sqrt K, \,\, 
		\upsilon^{\top} \bar{x}_{K, i} + \hat{\sigma}_{\upsilon} t_{1-\alpha/2,K-1} / \sqrt K \right]$
	}
\end{algorithm}

\revise{We focus on constructing scalar confidence intervals for linear projections. Specifically, our framework naturally accommodates inference on individual components by setting the projection vector $\nu$ to be a standard basis vector.  
By evaluating multiple directions, one can obtain inference on any collection of linear functionals of interest. In principle, sufficiently many projections can characterize the parameter space as comprehensively as an ellipsoidal set, while preserving the transparency and simplicity of univariate $t$–type inference. Nevertheless, our method can be extended to construct full ellipsoidal confidence sets using a Hotelling $T^2$ statistic; we provide the explicit construction and corresponding convergence results in Appendix \ref{sec:app_set}.}

\begin{remark}[Almost cost-free]
	We observe that the inference step in our method can be performed whenever necessary with minimal calculation and memory requirements, without needing any modifications to existing stochastic algorithms. This makes it almost cost-free and can be easily integrated into existing codebases. In contrast, all other methods typically demand considerable extra effort for inference. This may involve complex modifications, as seen in \cite{su2023higrad}, or entail storing and updating a $d\times d$ matrix at each iteration, as required by covariance-matrix-estimation-based methods or the random scaling method. In these cases, the computing and memory costs for inference purposes may exceed those involved in the SGD update itself.
\end{remark}

\begin{figure} 
    	\centering
    	\includegraphics[width=0.5\textwidth]{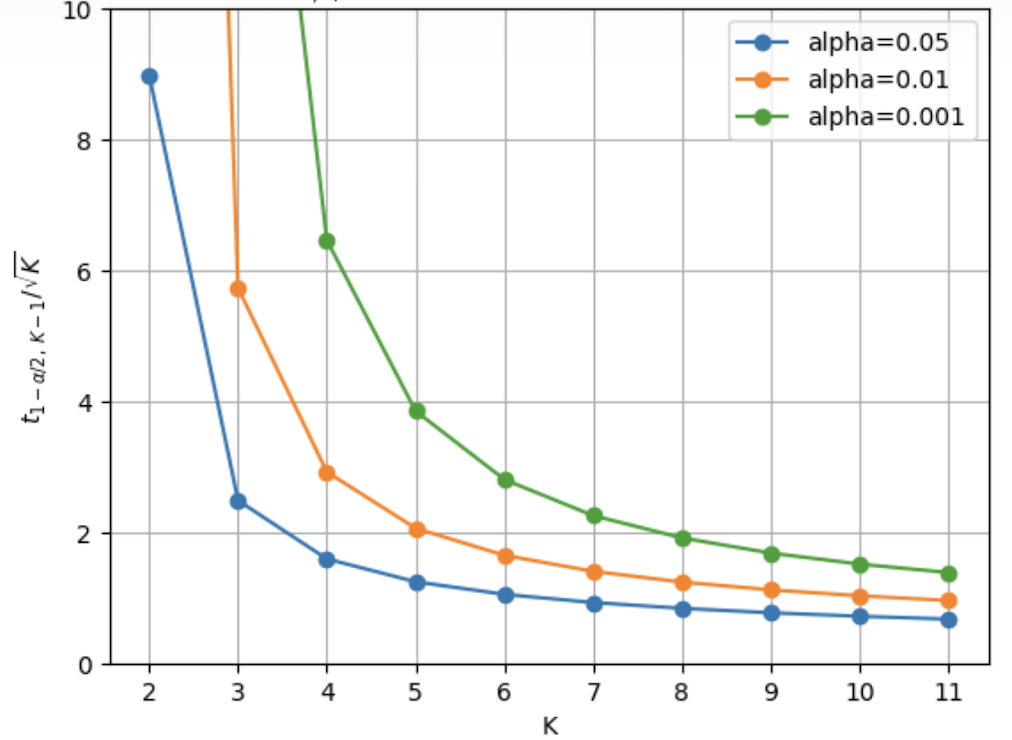}
    	\caption{$t_{1-\alpha/2,K-1}/\sqrt{K}$ VS. $K$ for various $\alpha$}
    	\label{fig:K} 
    \end{figure}

\revise{ \begin{remark}[Choice of $K$]
 The selection of $K$ involves a fundamental trade-off between validity (accurate coverage) and  statistical efficiency (shorter intervals).  When the total data budget $N$ is fixed, smaller $K$ increases the per-machine sample size $n=N/K$, improving the Gaussian approximation and thus reducing coverage error. This consideration becomes especially critical when estimation is difficult, such as in high-dimensional settings where a larger sample size $n$ is typically required as $d$ increases to maintain the validity of the inference.
   The interval length is proportional to $t_{1-\alpha/2, K-1}/\sqrt{K}$. This factor decreases as $K$ increases, so increasing $K$ tends to shorten the interval.  However, the marginal gain from increasing $K$ diminishes rapidly because the  $t$-quantile stabilizes as $K$ grows.   
In practice, we recommend  $K \approx 6$ as a starting point. This is a heuristic based on the ``elbow" of the efficiency curve. As shown in Figure \ref{fig:K}, the marginal gain in interval shrinkage becomes negligible once $K>6$. Therefore, there is little incentive to increase $K$ further, as doing so would sacrifice validity (by decreasing $n$) without meaningful gains in efficiency. Users may adjust $K$ based on their specific priorities and total data budget $N$.
\end{remark} 
}


\section{Theoretical guarantee}
\label{sec:thm}
In this section, we provide the theoretical guarantee for the confidence interval~\eqref{eq_CI_ASGD} constructed using the $t$-distribution.  Recall that we consider a \emph{high level of confidence} where the noncoverage level $\alpha$ can be potentially very small or decrease with the total sample size (or dimension).  This level of validation requires a more stringent guarantee than just showing that 
\begin{align}
	\label{eq_convergence_difference}
	\mathbb{P}(\upsilon^{\top} x^{*} \in \hat{\textnormal{CI}}) - (1 - \alpha) \rightarrow 0,
\end{align}
which can be derived from the convergence of relevant statistics in distribution as shown in other works. Our focus is to establish the bound of the relative error of coverage
\[\Delta_N := \sup_{\alpha(N)\le \alpha < 1}\left|\frac{ \mathbb{P}(\upsilon^{\top} x^{*} \in \hat{\textnormal{CI}}) - (1 - \alpha)}{\alpha} \right|,\]  
where $\alpha(N)$ goes to zero at an appropriate rate. Compared with~\eqref{eq_convergence_difference}, this bound offers a more rigorous assessment. It is critical in cases where we require high precision in our confidence assessments, ensuring that the constructed interval genuinely reflects the desired confidence level. For example, suppose we use the Bonferroni method to construct simultaneous confidence intervals for $m$ parameters at overall level $0.95$ with large $m$, then the CI for each individual parameter should be at level $1 - 0.05/m$. In this case $\alpha = 0.05/m$ and a small $\Delta_{N}$  is needed, while  \eqref{eq_convergence_difference} itself is not sufficient. Also, it is important to make the dependence on level $\alpha$ explicit since we may consider a decreasing $\alpha$. 

To derive the upper bound of the relative error, it is important to obtain the rate of convergence of the $t$-statistic.  In the rest of this section, we will first explore the application of ASGD in each parallel run and then extend our results to a broader class of stochastic algorithms that meet certain mild assumptions.

\subsection{Convergence characterization for ASGD}
\label{sec:thm_asgd}
Among various stochastic approximation algorithms, SGD is notably convenient and popular. Its variant, ASGD, is also widely used and has been the subject of extensive study. Beyond the well-known asymptotic normality results related to convergence in distribution, the rate of convergence to normality is of growing interest and has been studied in the literature. Notably, \cite{anastasiou2019normal} derived the non-asymptotic rate of convergence to a normal distribution using non-asymptotic rates of the martingale Central Limit Theorem (CLT), and~\citep{shao2022berry,wei2025gaussian} established a Berry–Esseen type bound for the Kolmogorov distance between the cumulative distribution functions of the ASGD estimator and its Gaussian analogue.

In this subsection, to better characterize the distributional approximation of the ASGD estimator, we develop a new Gaussian approximation of which the asymptotic normality is a direct consequence. Before presenting the main approximation result, we first introduce some regularity assumptions on the objective function and basic definitions. 
For any vector $\nu = (\nu_{1}, \ldots, \nu_{m})^{\top} \in \mathbb{R}^{m}$, we use $\|\nu\| = \sqrt{\sum_{\ell = 1}^{m} \nu_{\ell}^{2}}$ to denote its Euclidean norm.


\begin{assumption}
	\label{Assumption_convex_F}
	There exist positive constants $\tau$ and $L$ such that 
	\begin{align*}
		(x - x')^{\top} (\nabla F(x) - \nabla F(x')) &\geq \tau \|x - x'\|^{2}, \cr   
		\|\nabla F(x) - \nabla F(x')\| &\leq L \|x - x'\|. 
	\end{align*}
\end{assumption}

\begin{assumption}
	\label{Assumption_Delta_Lip}
	Denote $\Delta(x, \xi) = \nabla F(x) - \nabla f(x, \xi)$ for $x \in \mathbb{R}^{d}$ and $\xi \sim \Pi$. Given $q > 4$, we have $\mathbb{E}_{\xi} \|\Delta(x^{*}, \xi)\|^{q} < \infty$ and there exists some positive constant $\gamma$ such that for any $x, x' \in \mathbb{R}^{d}$, 
	\begin{align*}
		\left(\E_{\xi}\|\Delta(x, \xi) - \Delta(x', \xi)\|^{q}\right)^{1/q} \leq \gamma \|x - x'\|. 
	\end{align*}
\end{assumption} 
\begin{assumption}
	\label{Assumption_Hessian_Lip} 
	There exists some positive constant $\mathcal{L}$ such that for $x \in \mathbb{R}^{d}$, 
	\begin{align*}
		\|\nabla F(x) - \nabla^{2} F(x^{*}) (x - x^{*})\| \leq \mathcal{L} \|x - x^{*}\|^{2}. 
	\end{align*}
\end{assumption}

Assumptions~\ref{Assumption_convex_F}--\ref{Assumption_Hessian_Lip} are common and fairly mild in the context of convex optimization based on the SGD algorithm and its variants~\citep{Chen2020AoS, zhu2023online}. \revise{We note that these assumptions are introduced only for Theorem \ref{Theorem_strong_approximation} in the ASGD setting and are not required for the general results of our framework.} 

Theorem \ref{Theorem_strong_approximation} below establishes a Gaussian approximation result for the ASGD estimator. Before stating the theorem, we first define a covariance matrix.
For $n \geq 1$, we define 
\begin{equation}
	\label{eqn:Gamma}
	\Gamma_{n} = \frac{1}{n} \sum_{k = 1}^{n} U_{k} S U_{k}^{\top}, \enspace \mathrm{where}\ S = \mathrm{Cov}\{\nabla f(x^{*}, \xi)\}, \ \enspace U_{k} = \sum_{i = k}^{n} Y_{k}^{i}\eta_{k}
\end{equation}
with $Y_{k}^{k} = {\bf{I}}_{d},  Y_{k}^{i} = \prod_{l = k + 1}^{i} ({\bf{I}}_{d} - \eta_{l} \nabla^{2} F(x^{*})), i>k$. Note that $\Gamma_{n}$ is the covariance matrix of the linear form
\begin{equation}\label{eq:Mn}
    M_{n} = - n^{-1/2} \sum_{k = 1}^{n} U_{k} \nabla f(x^{*}, \xi_{k}),
\end{equation}
which is an asymptotic linear approximation of $\sqrt{n} (\bar{x}_{n} - x^{*})$, $\bar{x}_{n} = n^{-1} \sum_{i=1}^{n}x_{i}$.

\begin{theorem}
\label{Theorem_strong_approximation}
	Assume that $\{x_{i}\}_{i=1}^{n}$ is a SGD sequence  defined by:
	\begin{align*}
		x_{i} = x_{i - 1} - \eta_{i} \nabla f(x_{i - 1}, \xi_{i}), \quad i = 1, 2, \ldots, 
	\end{align*}
	where $\eta_{i} = \eta \times i^{-\beta}$ for some constant $\beta\in(1/2, 1)$. Under Assumptions~\ref{Assumption_convex_F}--\ref{Assumption_Hessian_Lip}, we have 
    \begin{equation}
    \label{eq:lineaapprox}
        \E\|\sqrt{n} (\bar{x}_{n} - x^{*}) - M_{n}\|^{2} \lesssim \max\left(n^{1 - 2 \beta}, \frac{\|x_{0} - x^{*}\|^{2}}{n}\right),
    \end{equation}
    where $M_{n}$ is defined in \eqref{eq:Mn}.
    Moreover, on a sufficiently rich probability space, there exists a $d$-dimensional random vector $W_{n} \overset{\mathcal{D}}{=} M_{n}$ and a centered Gaussian random vector $Z_{n} \sim \mathcal{N} (0, \Gamma_{n})$, where $\Gamma_{n}$ is defined in \eqref{eqn:Gamma}, such that 
	\begin{align}
	\label{eq_strong_approximation_W_Z}
		\E \|M_{n} - Z_{n}\|^{2} \lesssim \frac{\log n}{n^{1 - 2/q}}. 
	\end{align} 
\end{theorem}

\begin{remark}
Note that (\ref{eq:lineaapprox}) and (\ref{eq_strong_approximation_W_Z}) implies that the 2-Wasserstein distance
\begin{equation}
    W_2( \sqrt n (\bar{x}_{n} - x^{*}), \, \mathcal{N} (0, \Gamma_{n})) 
    \lesssim \max\left(n^{1 - 2 \beta}, \frac{\|x_{0} - x^{*}\|^{2}}{n}, \frac{\log n}{n^{1 - 2/q}}\right)^{1/2}. 
\end{equation}        
This motivates us to propose Assumption \ref{ass:normal_convergence_rate} without requiring the presence of the linear approximation term $M_n$. 
\end{remark}
\begin{remark}
	Theorem~\ref{Theorem_strong_approximation} reveals that the ASGD estimator can be approximated by a centered Gaussian random vector 
    and the approximation error is asymptotically negligible as long as $\beta > 1/2$ and $|x_{0} - x^{*}| \ll n^{1/2}$. To the best of our knowledge, this is the first result establishing a Gaussian approximation with explicit error bounds for the averaged SGD estimator, strengthening the classical asymptotic normality theory for stochastic approximation. It is worth noting the SGD iterates $x_{i} \in \mathbb{R}^{d}, i = 1, 2, \ldots$, are neither independent nor stationary. Hence the existing strong invariance principle results for the partial sums of independent random elements (e.g., \cite{Komlos1975I, Komlos1976II, CsorgHo1975, Einmahl1987}) or general stationary sequences (e.g., \cite{Wu2007strong, Liu2009strong, Berkes2014komlos}) are not directly applicable here. To handle the nonstationary property of the sequence $\{x_{i}\}_{i \geq 1}$, we shall invoke the recently established strong approximation result for non-stationary time series \cite{mies2023sequential}. 
\end{remark} 
\begin{remark}
	It is worth noting the covariance matrix $\Gamma_{n}$ defined in~\eqref{eqn:Gamma} and hence the distribution of the coupled Gaussian vector $Z_{n}$ do not depend on the initial estimate $x_{0}$. Therefore, our result in Theorem~\ref{Theorem_strong_approximation} can also be viewed as a {\it quenched} Gaussian approximation in the sense that the impact of the initial point $x_0$ diminishes, as asserted by the second term ${\|x_{0} - x^{*}\|^{2}}/{n}$ in the upper bound~\eqref{eq:lineaapprox}. As a direct consequence, the distribution of the random vector $\sqrt{n}(\bar{x}_{n} - x^{*})$ can be approximated by that of $\mathcal{N}(0, \Gamma_{n})$. Moreover, the multivariate central limit theorem~\eqref{eq_MCLT_SGD} can be easily derived as $\Gamma_{n}$ converges to the sandwich form covariance matrix $\Sigma = A^{-1} S A^{-1}$; see, for example~\cite {polyak1992acceleration}. It is important to mention that our procedure does not rely on the convergence of $\Gamma_{n}$ to $\Sigma$, which can be slow and introduce additional approximation error in practical implementations. Our constructed $t$-statistic is asymptotically pivotal as long as~\eqref{eq:lineaapprox} and~\eqref{eq_strong_approximation_W_Z} hold. Particularly, the simulation studies in Section~\ref{sec:exp} demonstrate that our procedure has better finite-sample performance than the oracle procedure based on the multivariate central limit theorem~\eqref{eq_MCLT_SGD} with the population covariance matrix $\Sigma$ given.  
\end{remark}

The rate of convergence to normality plays a crucial role in assessing the high-probability approximation of the $t$-distribution. As we will discuss in a later section through a general theorem, the convergence of the $t$-statistic and the upper bound of the relative error relies on the convergence rate (of a single parallel run sequence) to normality.

\subsection{Main results}
\revise{To derive general results of our framework that apply to a broad class of stochastic optimization methods and settings---without imposing specific regularity conditions on the objective function (such as convexity or smoothness)---we introduce the following assumption.}
\begin{assumption}[Convergence rate to normality]
\label{ass:normal_convergence_rate}
For a chosen stochastic algorithm and number of parallel runs $K$, let $\hat{x}_{n}^{(k)} (k = 1, \ldots, K)$ denote the result at the $n$-th iteration of the $k$-th parallel run used in calculating the parallel average $\bar{x}_{K, n}$ in \eqref{eqn:parallel_average}. There exists a centered Gaussian random vector $Z_{n} \sim \mathcal{N} (0, \Sigma_{n})$ (for some $\Sigma_{n}\in\R^{d\times d}$)
such that 
$$
\left(\E\|\sqrt{n}(\hat{x}_{n}^{(k)} - x^{*}) - Z_{n}\|^2\right)^{1/2} \lesssim \delta(n),
$$
where the approximation rate $\delta(n) \rightarrow 0$.
\end{assumption} 


\revise{Many stochastic gradient algorithms  are known to exhibit the asymptotic normality.
Examples include weighted-averaged SGD with various weighting schemes \citep{shamir2013stochastic, rakhlin2011making, wei2026general} , modified SGD with adaptive-scaled gradient or momentum \citep{li2022root,tang2023acceleration} , constant step-size SGD in certain non-convex regimes \citep{Yunonconvex},   zeroth-order SGD \citep{Chen01102024}, and  second order methods such as stochastic sequential quadratic programming for constrained stochastic nonlinear optimization problems \citep{na2025asymptotic} and also SGD in online decision making setting \citep{chen2021statistical}. For any algorithm that satisfies asymptotic normality property, there exists a corresponding approximation error function $\delta(n)\rightarrow 0$.  
In the previous section, we provide an explicit expression for $\delta(n)$ in the case of ASGD. 
Theorem \ref{Theorem_strong_approximation} demonstrates that if we employ ASGD as defined in \eqref{eqn:asgd}, Assumption \ref{ass:normal_convergence_rate} is satisfied with $\Sigma_{n} = \Gamma_{n}$ defined in~\eqref{eqn:Gamma} and 
\begin{align*}
\delta(n)=\max\left(n^{1/2 - \beta}, \frac{\|x_{0} - x^{*}\|}{\sqrt{n}}, \frac{\sqrt{\log n}}{n^{1/2 - 1/q}}\right).
\end{align*} 
 For the other algorithms mentioned above, analogous rates exist in principle, although deriving them requires detailed algorithm-specific analysis. Importantly, the exact convergence rate is not needed in order to apply the method.}

With this assumption, we are ready to show that the statistic in \eqref{eqn:t_v} is asymptotically pivotal with a specific convergence rate.

\begin{theorem}
\label{thm:sup}
Suppose we run Algorithm~\ref{alg:parallel} and Assumption~\ref{ass:normal_convergence_rate} holds. For any $\upsilon$ and $\hat{t}_{\upsilon}$ defined in \eqref{eqn:t_v}, we have
$$    \sup_{z \in \mathbb{R}} \left|\mathbb{P}\left(\hat{t}_{\upsilon}\ge z\right) - \mathbb{P}(T_{K-1}\ge z)\right| \lesssim (\delta(N/K))^{1/4},
$$
where $T_{K-1}$ is a random variable following the $t$ distribution with degrees of freedom $K-1$, $N$ is the total sample size and $K$ is the number of parallel runs. Consequently, for any confidence level $\alpha \in (0, 1)$, 
\begin{equation}\label{eq:relative_error_rate}
\left|\frac{\mathbb{P}\left(|\hat{t}_{\upsilon}|\ge t_{1-\alpha/2, K-1}\right)}{\alpha} - 1\right|\lesssim\alpha^{-1}\delta(N/K)^{1/4},
\end{equation} 
where $t_{1-\alpha/2,K-1}$ is the $(1-\alpha/2)\times100\%$ percentile for the $t_{K-1}$ distribution and the constant in $\lesssim$ does not depend on $\alpha$.
For $\alpha(N)$ goes to zero with $\delta(N/K)^{1/4} \ll \alpha(N)$, the relative error of coverage goes to zero when  $\alpha \ge\alpha(N)$, i.e., 
\[\Delta_N = \sup_{\alpha(N)\le \alpha < 1}\left|\frac{ \mathbb{P}(\upsilon^{\top} x^{*} \in \hat{\textnormal{CI}}) - (1 - \alpha)}{\alpha} \right|\rightarrow 0.\] 
\end{theorem}

The results suggest that any stochastic algorithm demonstrating appropriate convergence in certain scenarios can be selected, provided its single sequence exhibits convergence towards normality.  The convergence of the $t$-statistic, as well as the relative error in the coverage of the confidence interval, can be bounded based on the rate of convergence to normality. Furthermore, this study comprehensively examines the reliance on the value of $\alpha$. The uniform convergence of $\Delta_{N}$ indicates that an extremely small $\alpha$, or decreasing $\alpha$, is feasible.

\revise{In the case of ASGD, plugging in the corresponding $\delta$ into \eqref{eq:relative_error_rate}, the condition for the relative error to vanish is 
\[
\alpha(N)\gg
\left[
\max\left( (N/K)^{1/2-\beta},\;\frac{\sqrt{K}\|x_{0} - x^{*}\|}{\sqrt{N}},  \frac{\sqrt{\log (N/K)}}{(N/K)^{1/2-1/q}}\right)
\right]^{1/4}.
\]
This explicitly characterizes the rate at which $\alpha(N)$ is allowed to converge to $0$ in the ASGD setting.
}

\section{Experiment}\label{sec:exp}

 \revise{In this section, we evaluate the empirical performance of the proposed inference procedure across a range of settings with increasing complexity. We begin with standard convex models, where the classical theory applies, and then move to non-convex and nonlinear problems to demonstrate the generality of our approach. In particular, we consider (i) linear and logistic regression, (ii) a non-convex optimization problem where the asymptotic normality of SGD has been established in prior work, and (iii) an online source localization problem as a representative real-world application with a nonlinear and non-convex objective. All reported results are averaged over 10000 independent trials.}

\subsection{Convex objectives}
\label{sec:exp_convex}
\revise{We begin by investigating the parallel inference method under classical convex settings --- linear and logistic regression models, employing ASGD with a decaying learning rate $\eta_i = \eta i^{-\beta}$. These settings allow for a direct comparison with existing approaches under well-understood conditions. }
 
\noindent{\bf  Set up.} The true coefficient of interest $x^*$ is a $d$-dimensional vector with $x^* =  (0, 1, 2, \dots, d-1)/d$. We generate a sequence of i.i.d.~random samples $\{(a_i, b_i)\}_{i=1}^{n}$, where $a_i$ denotes the explanatory variable generated from $\mathcal{N}(0,{\bf{I}}_d)$, and $b_i$ denotes the response variable. In the linear regression model, we have 
\[b_i  = a_i^Tx^* + \epsilon_{i},\]
where $\epsilon_{i}$ follows $\mathcal{N}(0,1)$ independently.
The corresponding  loss function is
$f(x,\xi_i )=(a_i^T x-b_i)^2/2$.
In the logistic regression model, $b_i \in \{0,1\}$ is generated from a Bernoulli distribution, where 
\[\mathbb{P}(b_i=1|a_i)=\frac{1}{1+\exp (-a^T_ix^*)},\]
and the loss function is logit loss as
$f(x,\xi_i)=(1-b_i)a_i^Tx+\log (1+\exp (-a^T_ix))$. 
We employ ASGD for our parallel method, with $\beta=0.505$, and $\eta=0.5$, consistent with the settings used in \cite{lee2022fast,zhu2023online}. 
We consider the case of marginal inference of coordinates, that is, the vector $\upsilon$ in the linear functional is chosen as the canonical basis. To analyze the empirical performance, we record the coverage of the constructed confidence intervals, the relative error of coverage $\Delta_{\alpha}$ as defined in \eqref{ratiodiff}, the length of the confidence intervals, and the running time.  

\begin{figure}
\centering 
\subfigure[]{\includegraphics[width=0.4\textwidth]{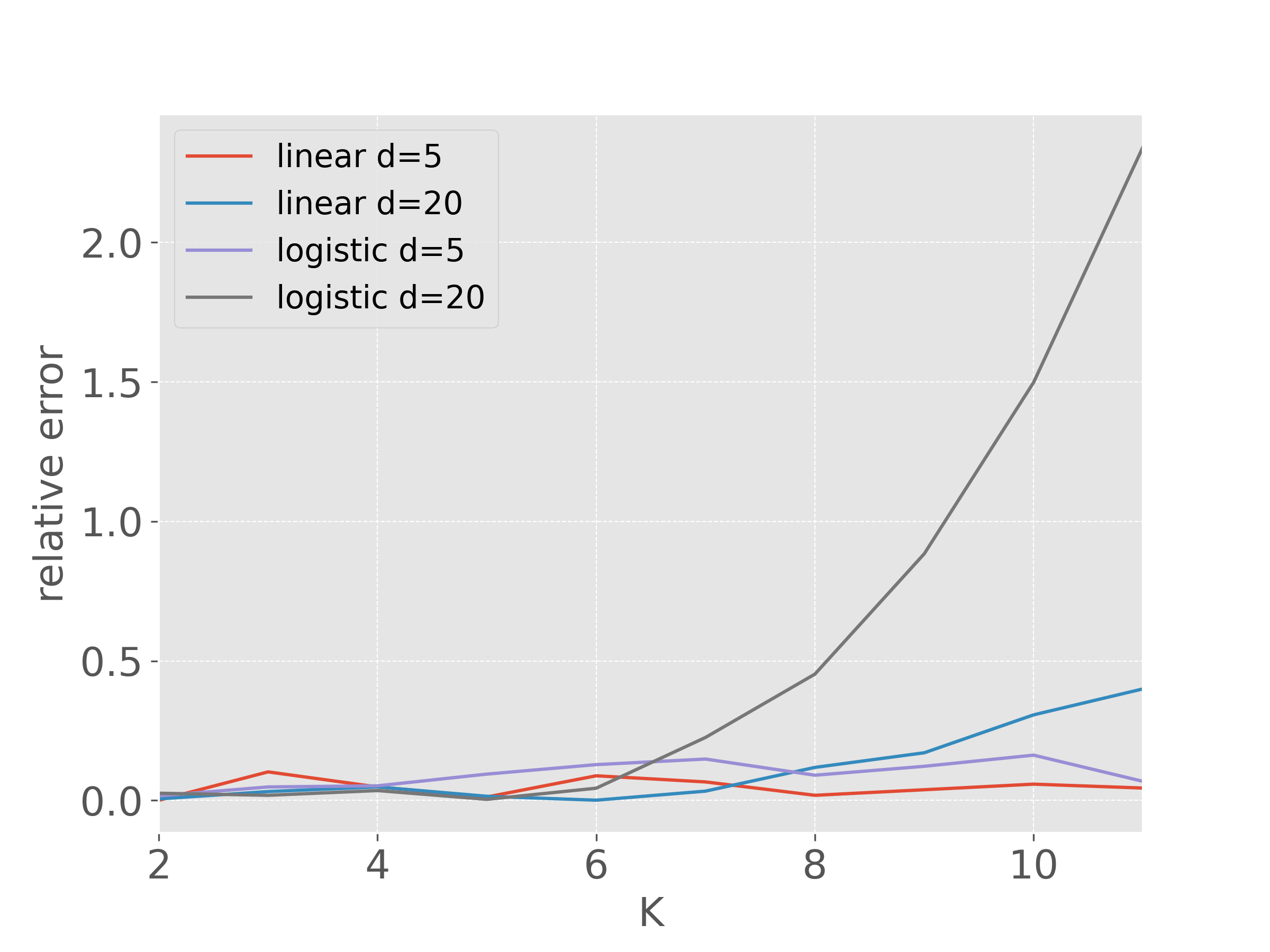}}
\subfigure[]{\includegraphics[width=0.4\textwidth]{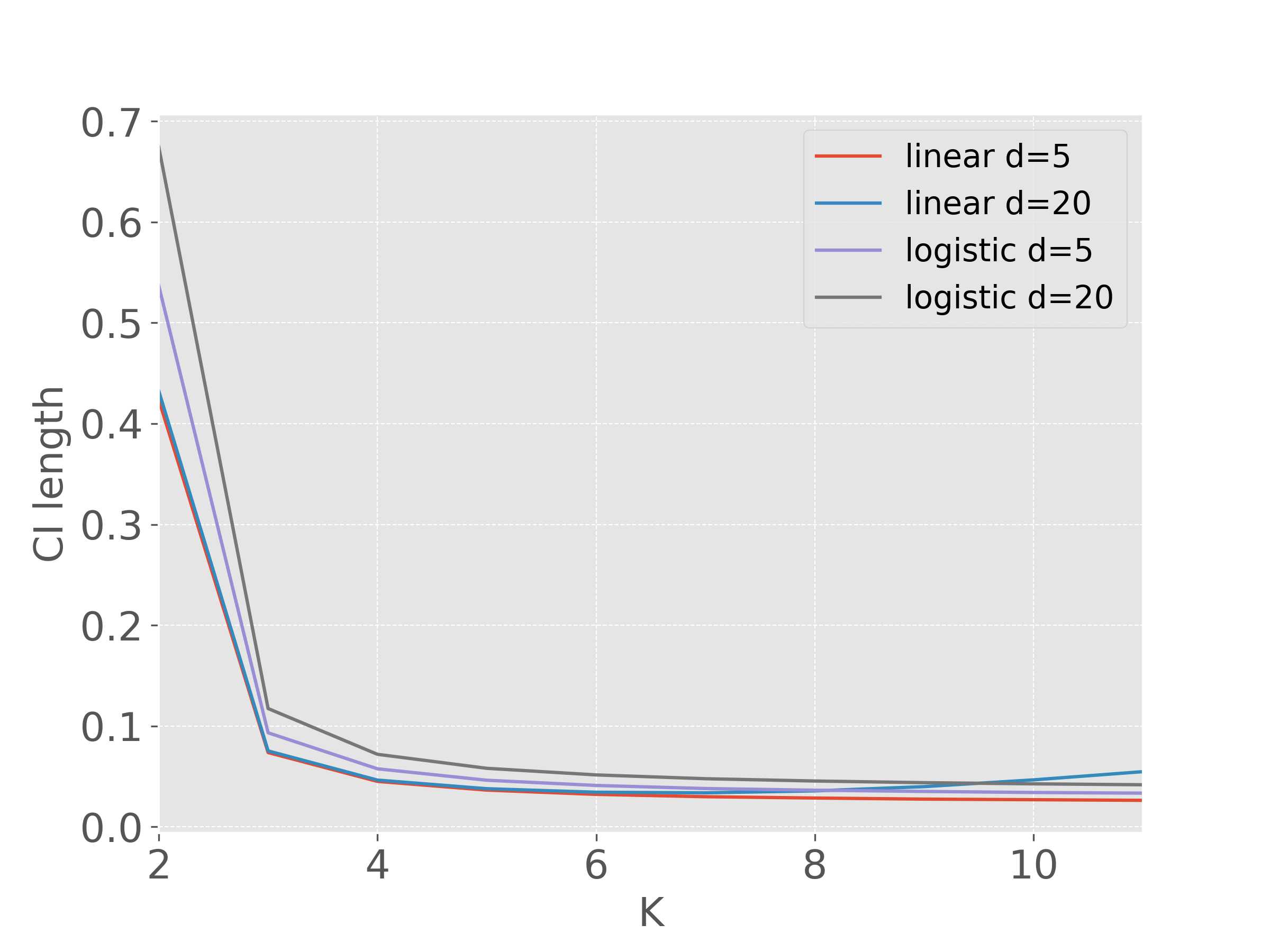}}
\caption{Effect of $K$. Plot (a): relative error of coverage; plot (b): the length of confidence interval. The nominal coverage probability is 0.99. The total sample size $N$ is $60000$ for linear models and $200000$ for logistic models.}
\label{fig:compare_K} 
\end{figure}

\noindent{\bf  Choice of $K$.}
We first examine the effect of  $K$. We construct $99\%$  confidence intervals ($\alpha = 0.01$) with a total sample size  of $60000$ ($N = nK = 60000$)  for linear regression, and  with a total sample size of $200000$  ($N = nK = 200000$) for logistic regression.  \revise{Regarding coverage, a small $K$ can reduce coverage bias. As shown in Figure~\ref{fig:compare_K}, while the choice of $K$ is relatively insensitive in simple estimation problems (e.g., linear regression), its impact becomes significant in challenging scenarios. For example, in logistic regression, the sensitivity to $K$ is much more pronounced when $d = 20$ compared to $d = 5$. As $d$ increases, a larger per-machine sample size $n$ is typically required for the Gaussian approximation to hold; consequently, a smaller $K$ is preferred to ensure that each machine has sufficient data for valid $t$-inference. Regarding interval length, the results are consistent with our previous discussion: a larger $K$ results in shorter confidence intervals, but the marginal gain decreases rapidly as $K$ increases. Overall, when $K$ falls within a reasonable range (e.g., between 3 and 8), the results appear satisfactory and are not overly sensitive to the specific choice. Based on this, we use $K = 6$ for the following simulation results. }

\noindent{\bf  Comparison with random scaling.}  
We compare the finite sample performance of our proposed inference method, referred to as the parallel method, with that of the state-of-the-art method: the random scaling method \citep{lee2022fast}, which also leverages an asymptotic pivotal statistic. The confidence interval constructed by the random scaling method is given in \eqref{eqn:rs}, and we obtain critical values through Monte Carlo simulation. We did not include comparisons with other methods such as the Plug-in \citep{Chen2020AoS} or Online Batch-means \citep{zhu2023online}, as the random scaling method has already demonstrated comparable coverage to the Plug-in method, superior coverage compared to Online Batch-means, and faster computing times. For both methods, we apply the ASGD algorithm with $\beta = 0.505$ and $\eta = 0.5$. The number of parallel runs, $K$, is set to $6$ for the parallel method. We consider constructing confidence intervals every 600 samples. Overall, the performance of the parallel method is satisfactory and better than that of the random scaling method, with faster convergence, comparable confidence interval lengths, and less computation.


\begin{figure}
\subfigure[$\alpha = 0.05$]{
	\includegraphics[width=0.33\textwidth]{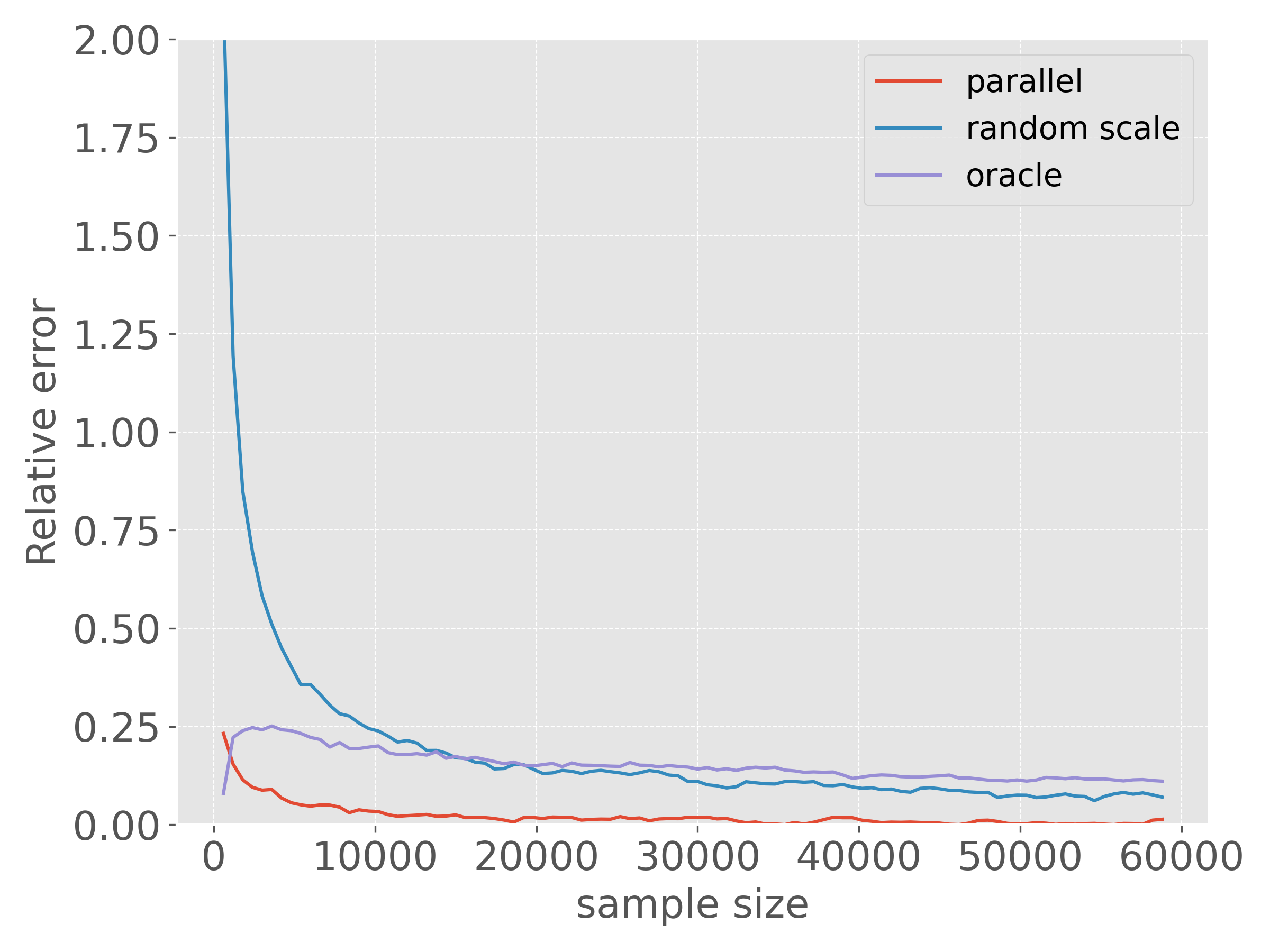} 
	\includegraphics[width=0.33\textwidth]{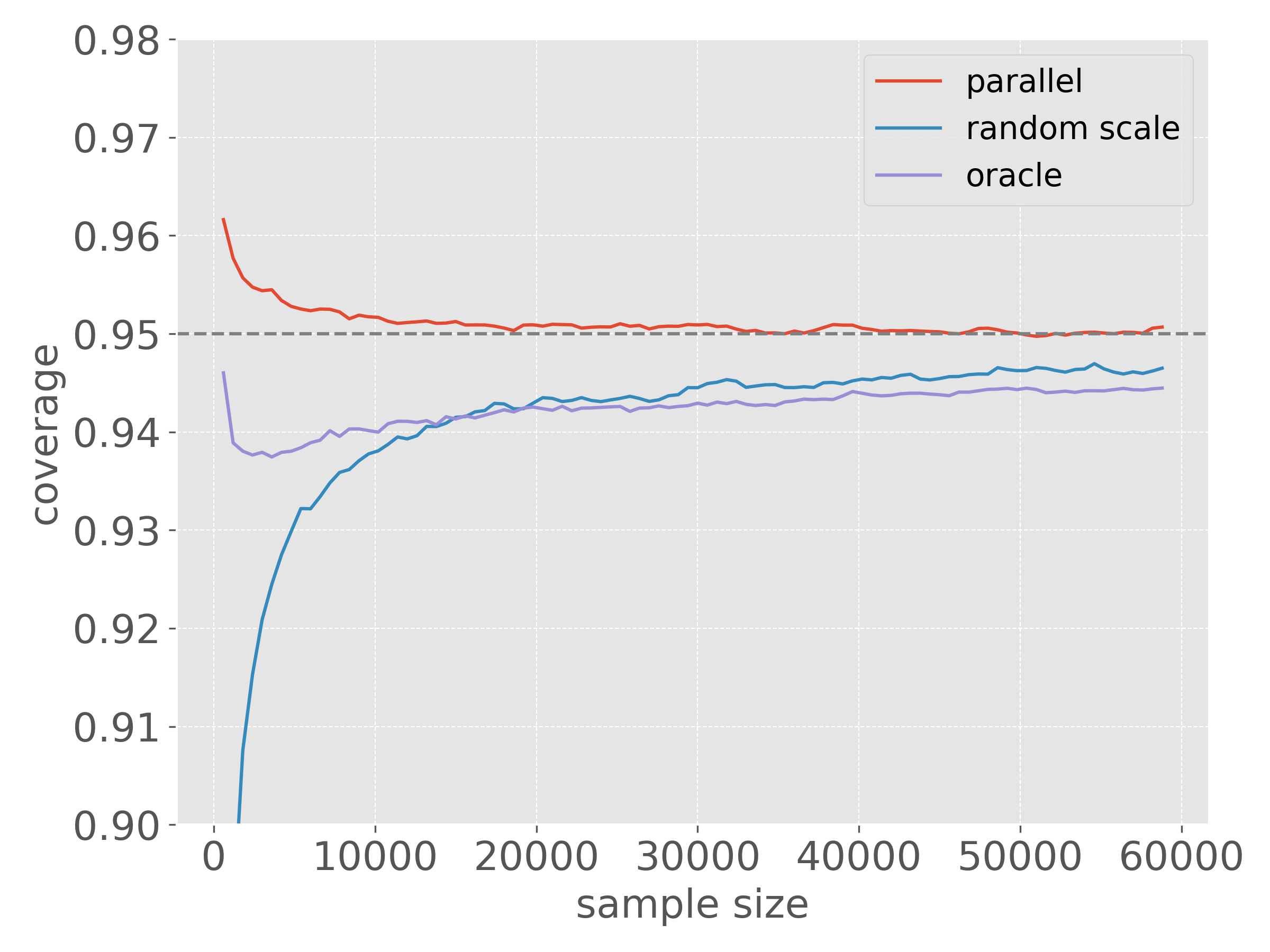} 
	\includegraphics[width=0.33\textwidth]{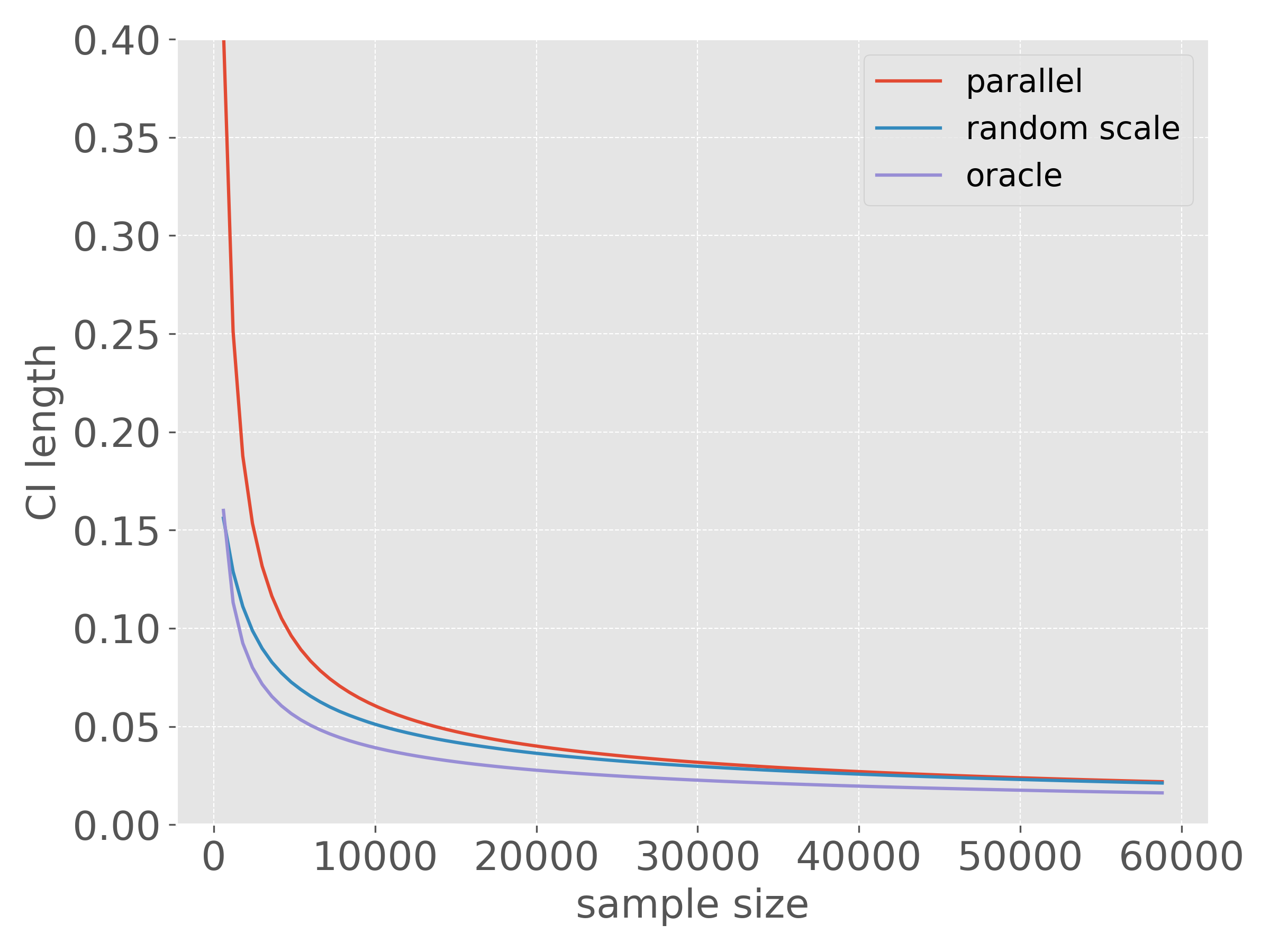}
}
\subfigure[$\alpha = 0.01$]{
	\includegraphics[width=0.33\textwidth]{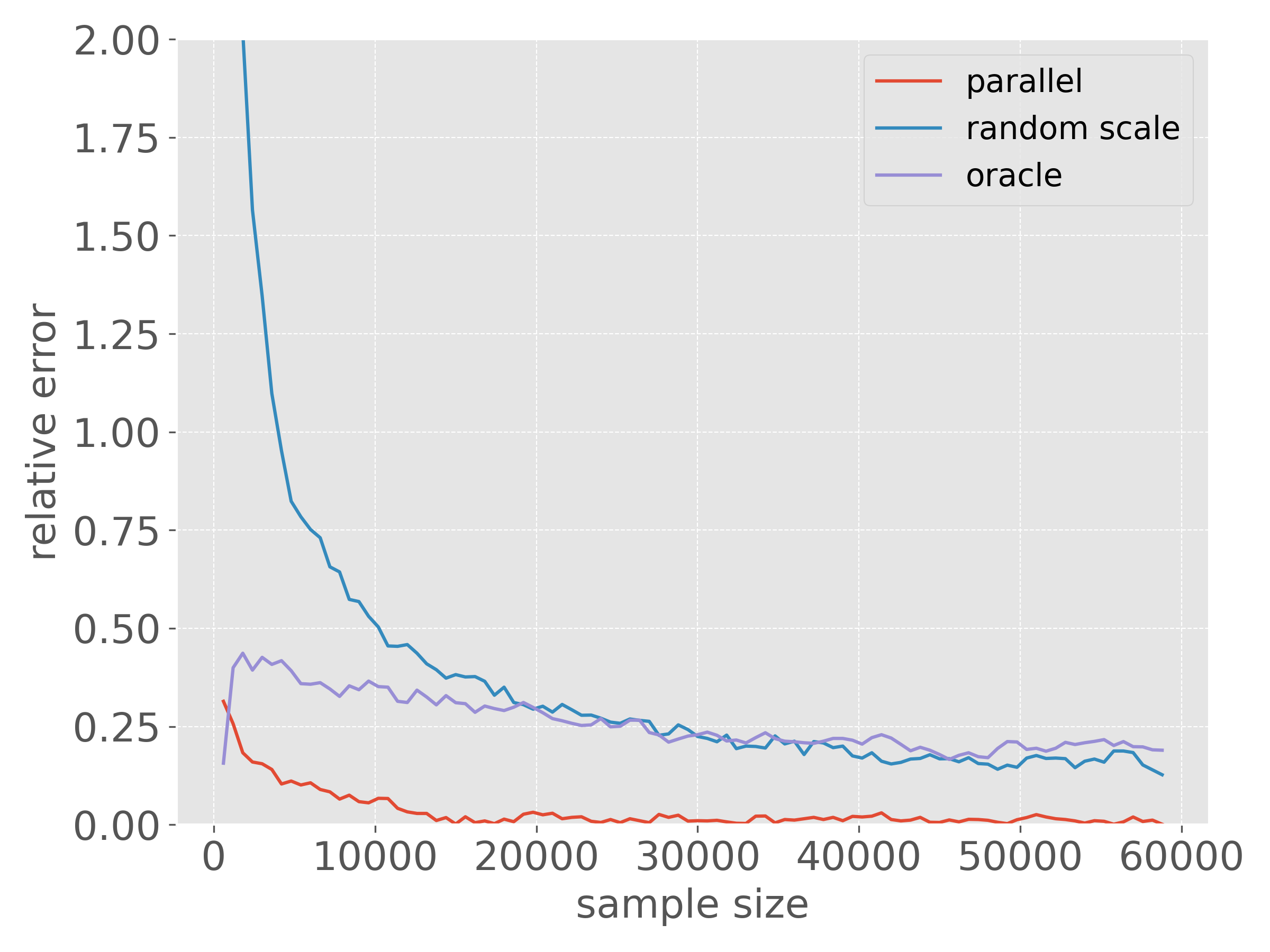} 
	\includegraphics[width=0.33\textwidth]{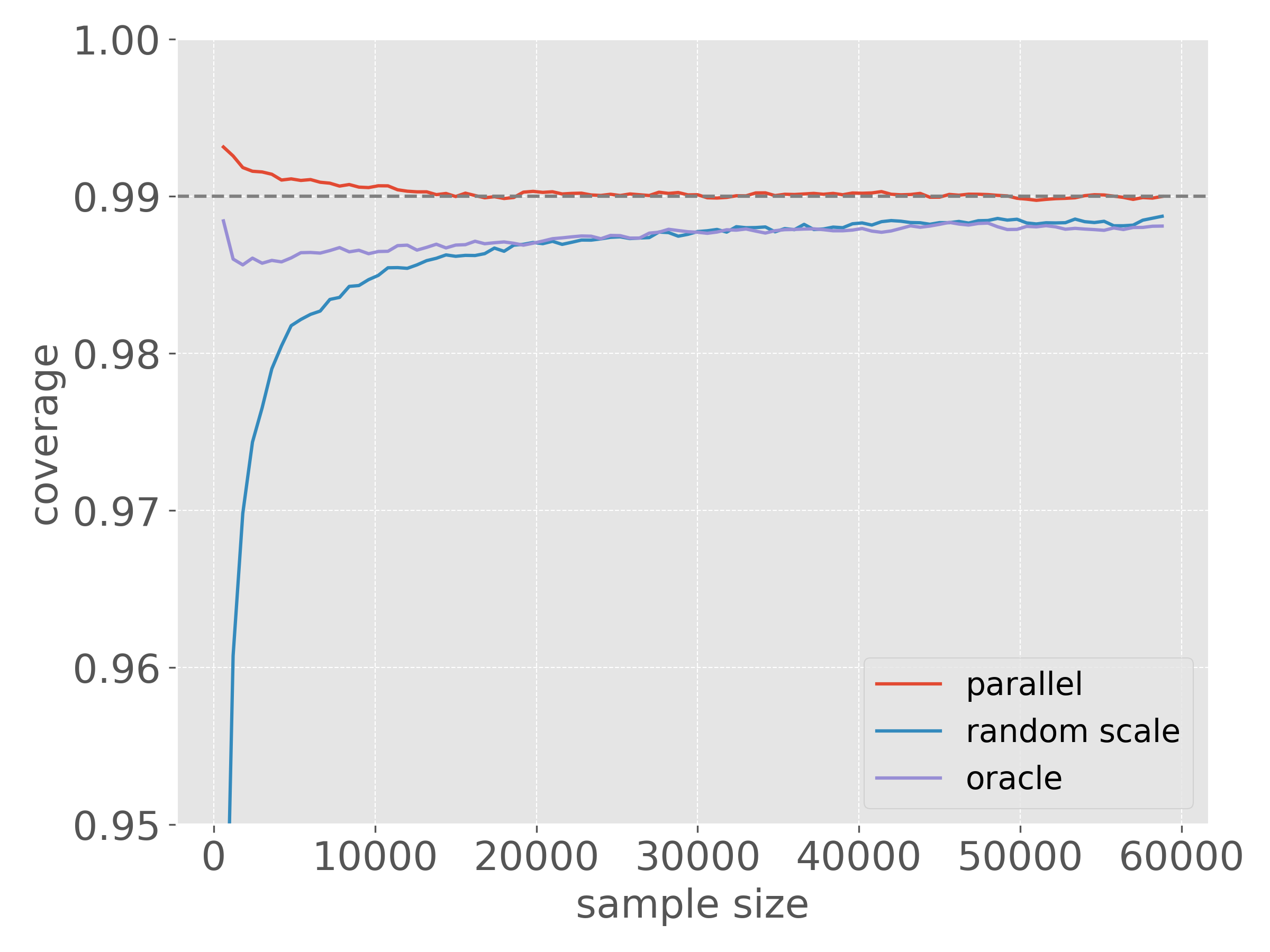} 
	\includegraphics[width=0.33\textwidth]{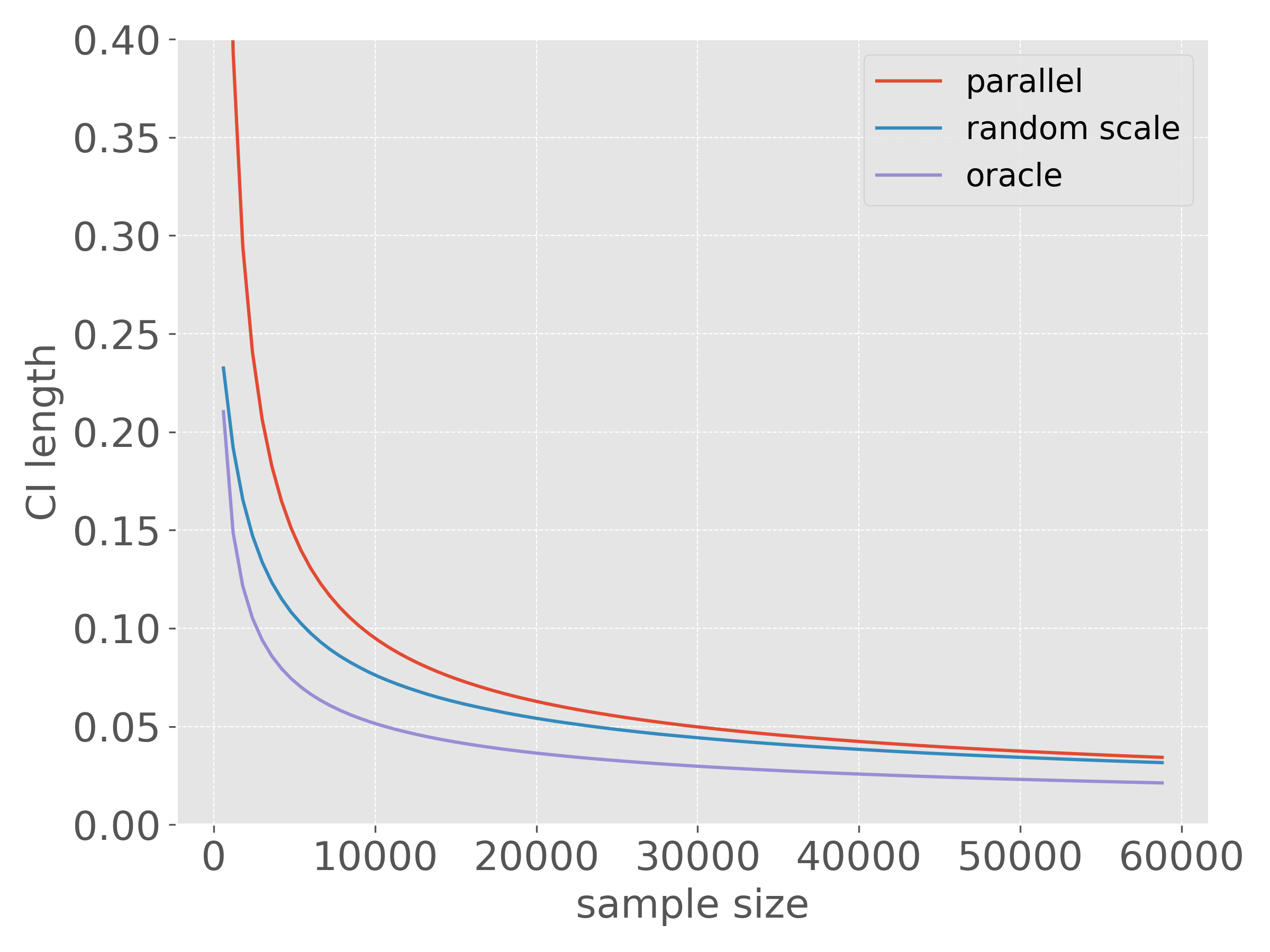}
}
\subfigure[$\alpha = 0.001$]{
	\includegraphics[width=0.33\textwidth]{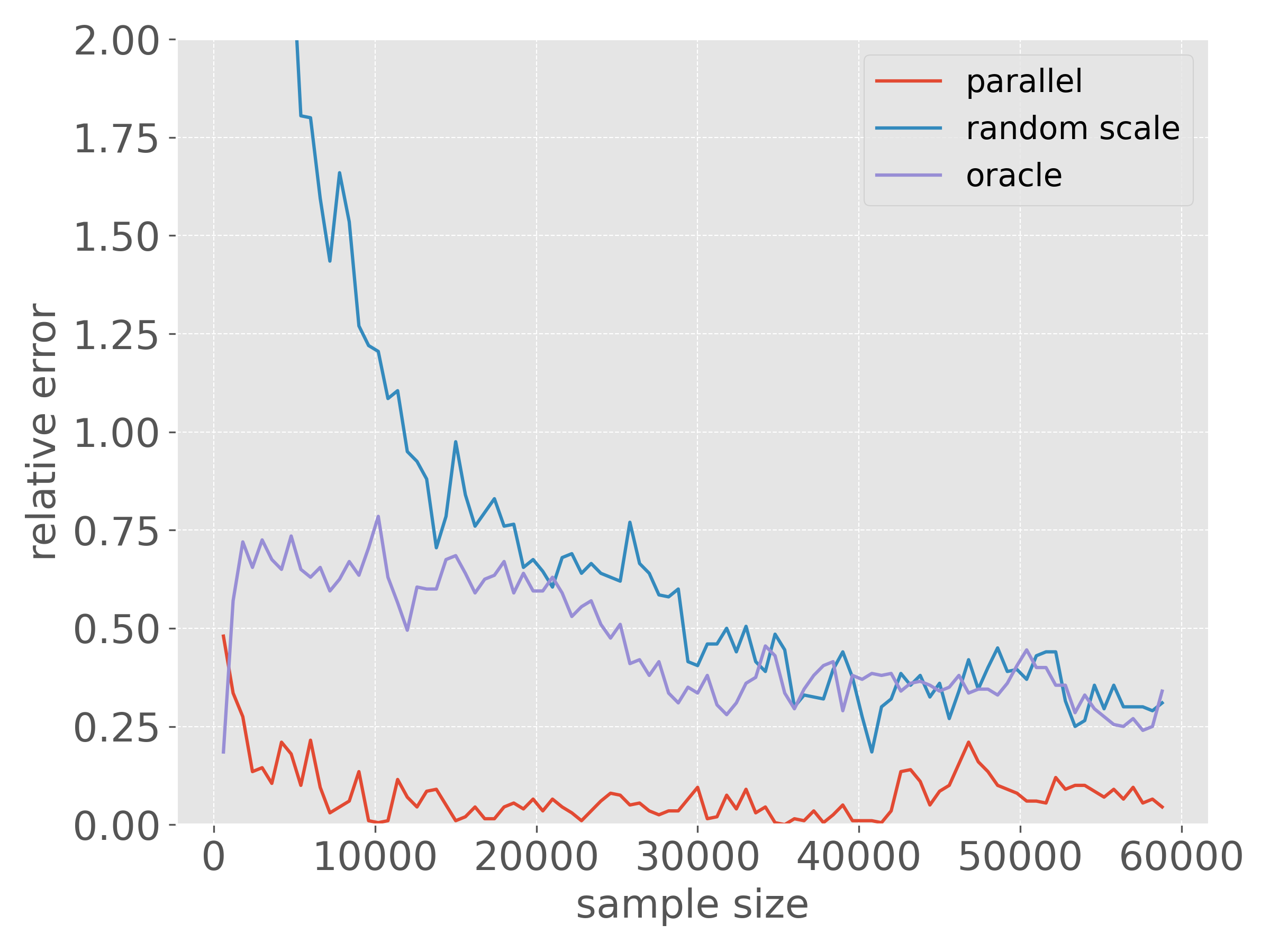} 
	\includegraphics[width=0.33\textwidth]{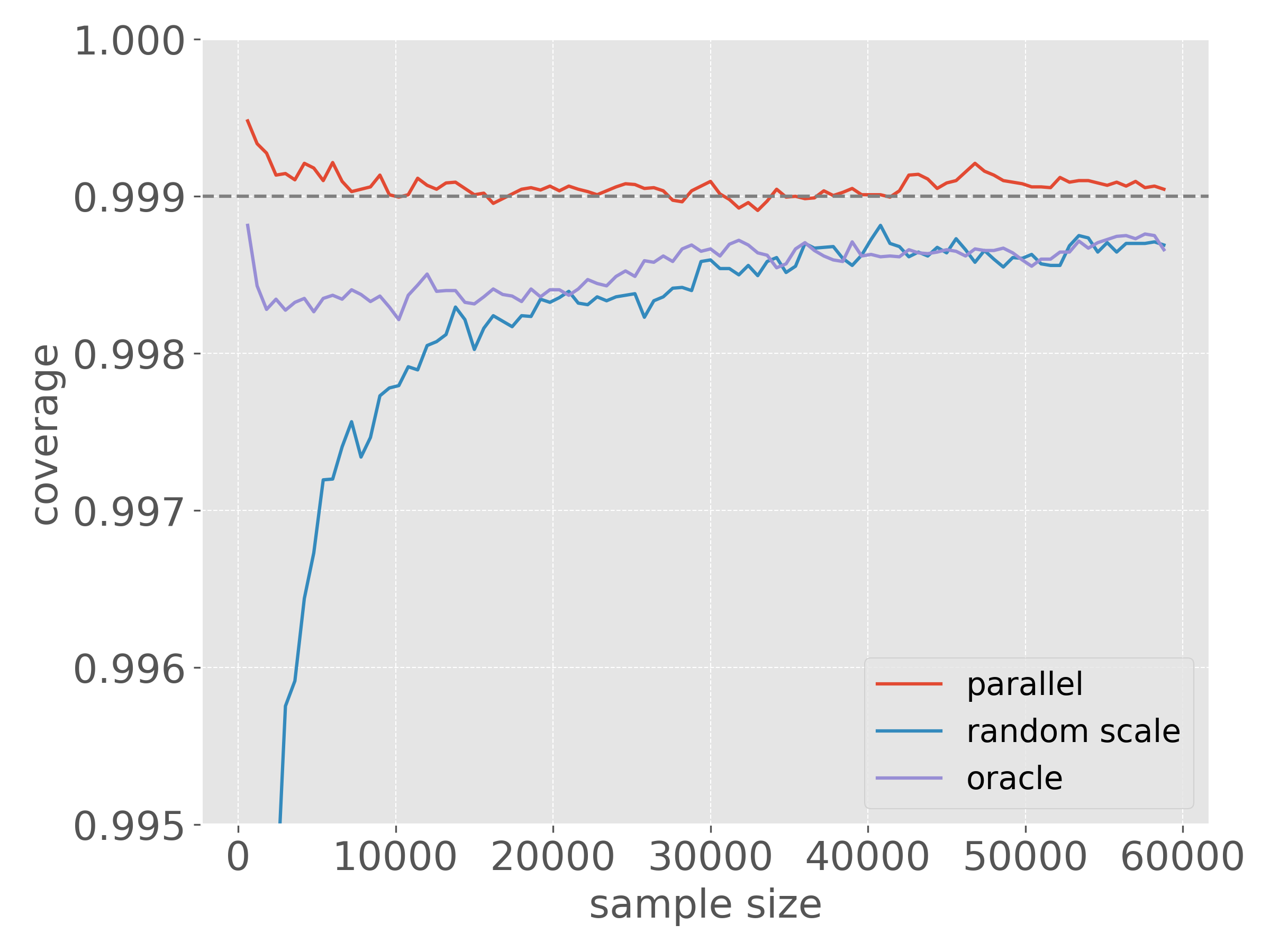} 
	\includegraphics[width=0.33\textwidth]{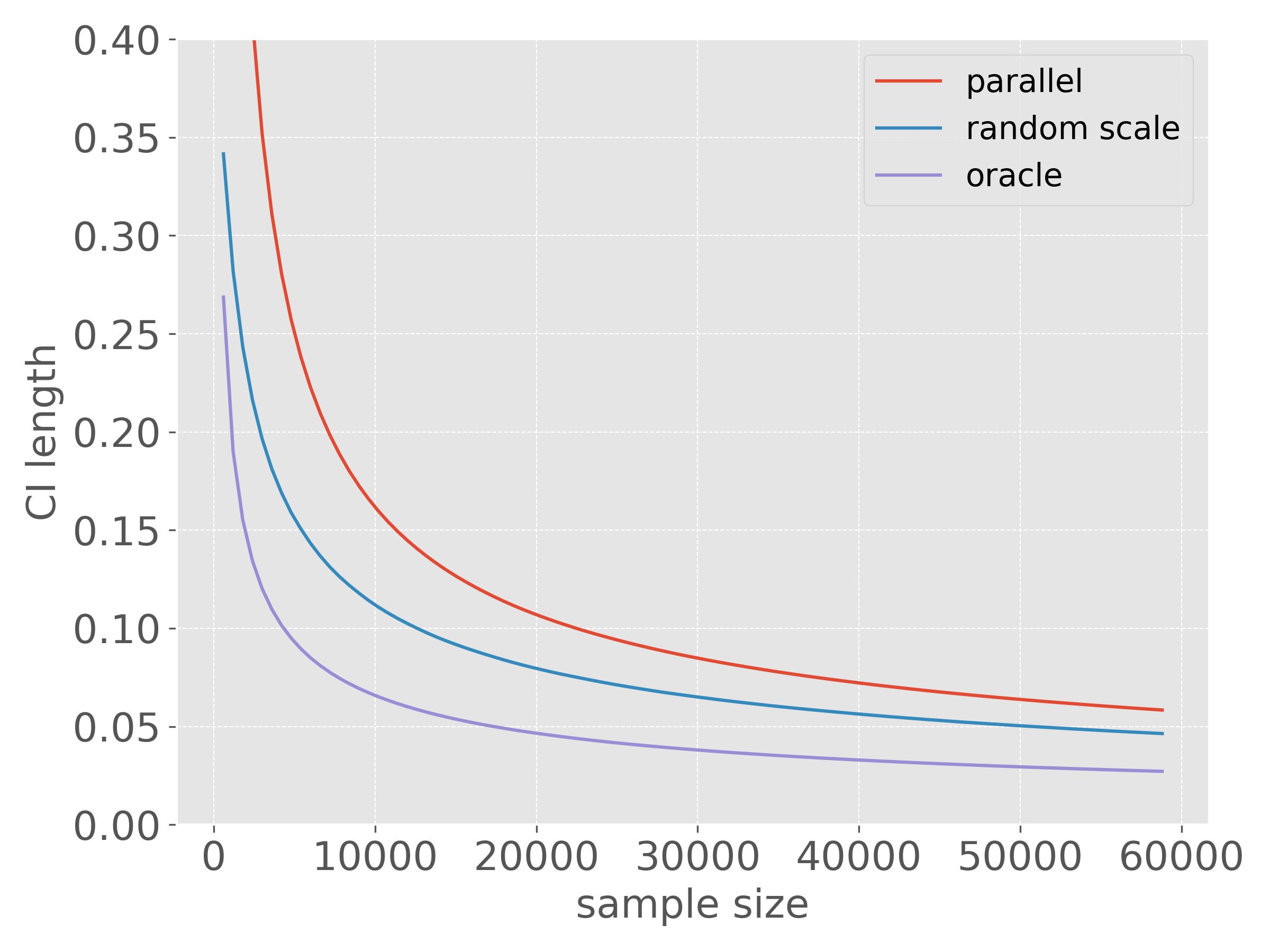}
}
\caption{Linear Regression $d = 20$. Left: relative error of coverage; Middle: empirical coverage; Right: length of confidence intervals.}
\label{fig:linear_d20}
\end{figure}

\begin{figure}
\subfigure[$\alpha = 0.05$]{
	\includegraphics[width=0.33\textwidth]{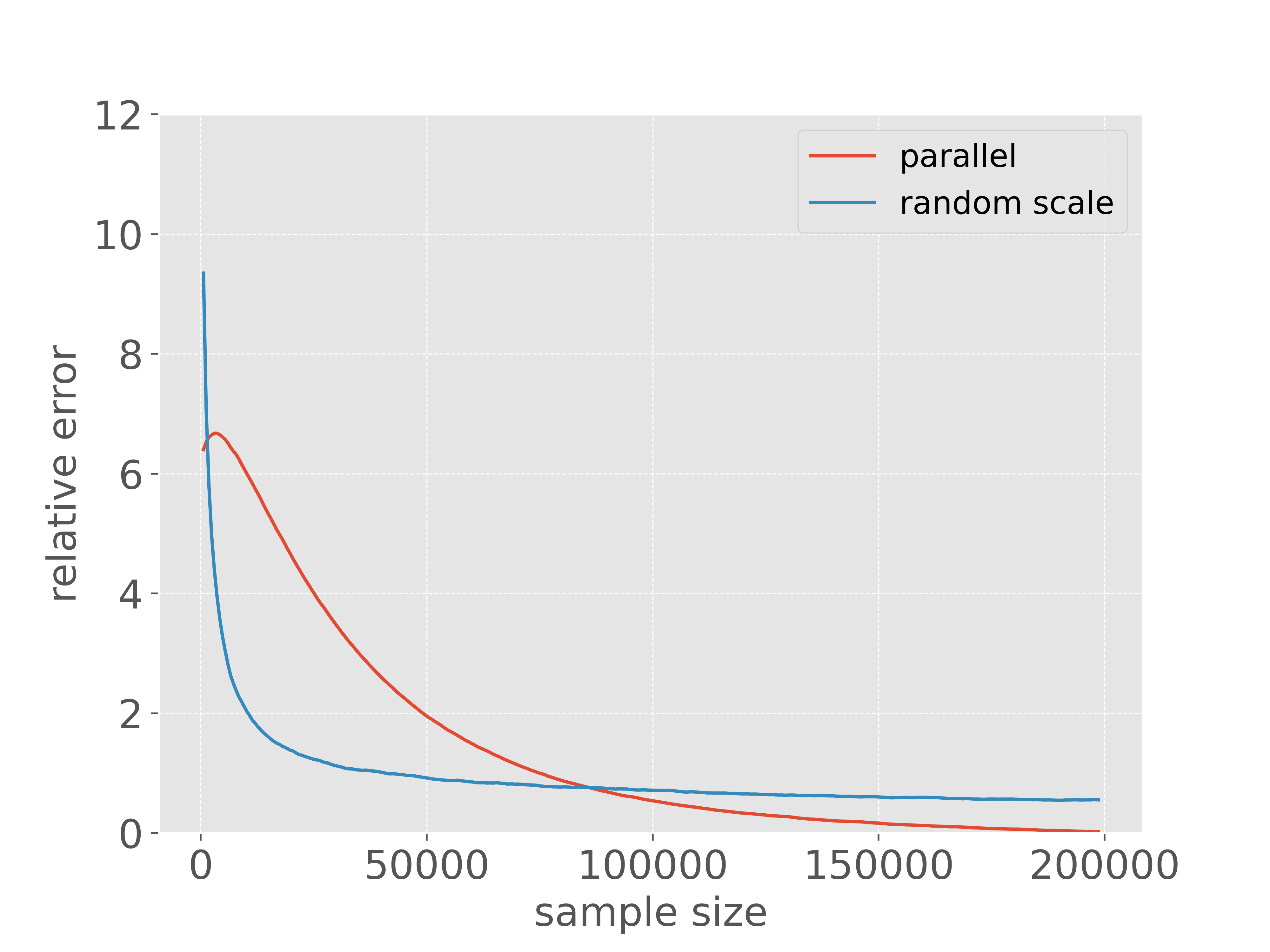} 
	\includegraphics[width=0.33\textwidth]{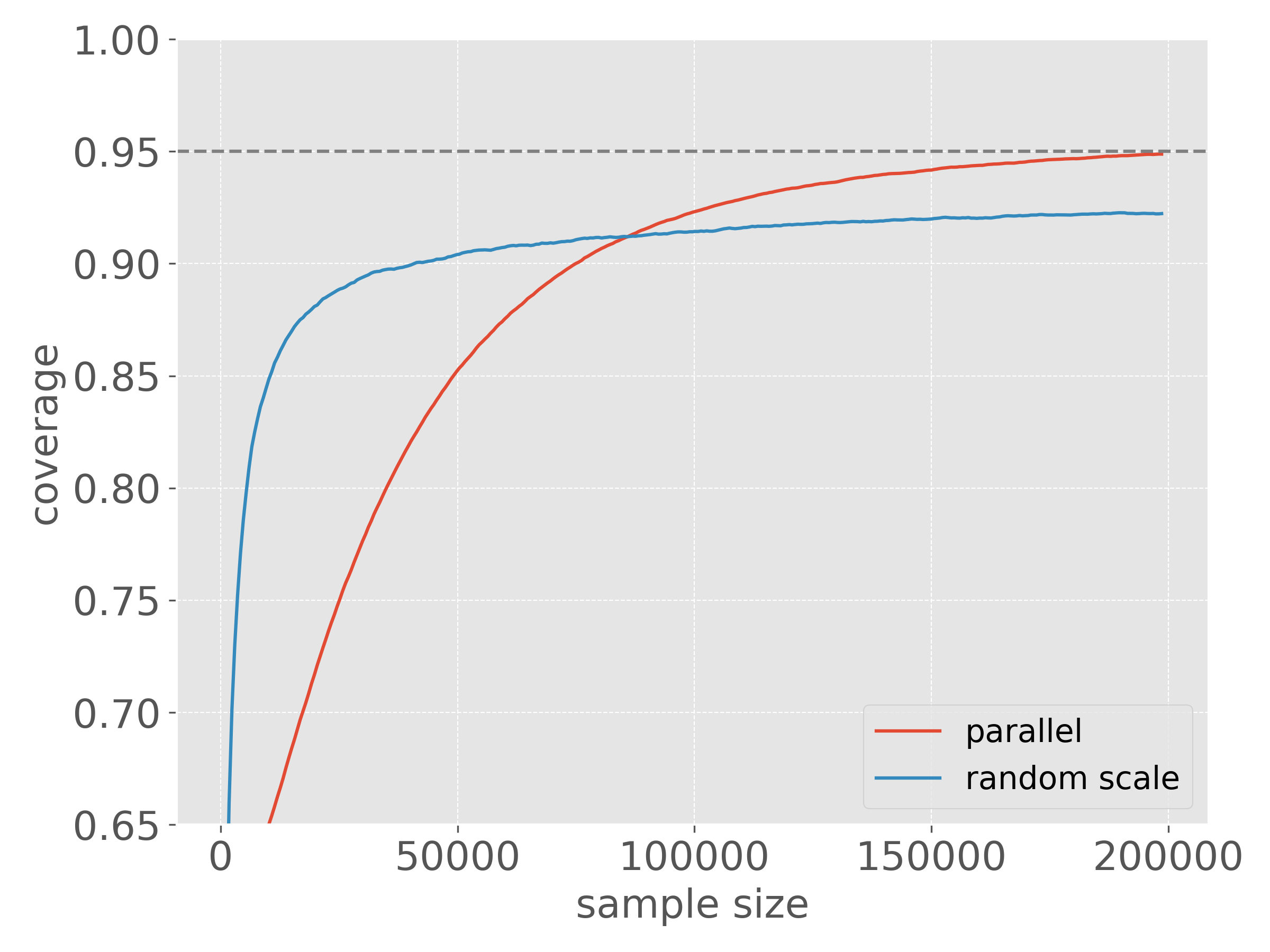} 
	\includegraphics[width=0.33\textwidth]{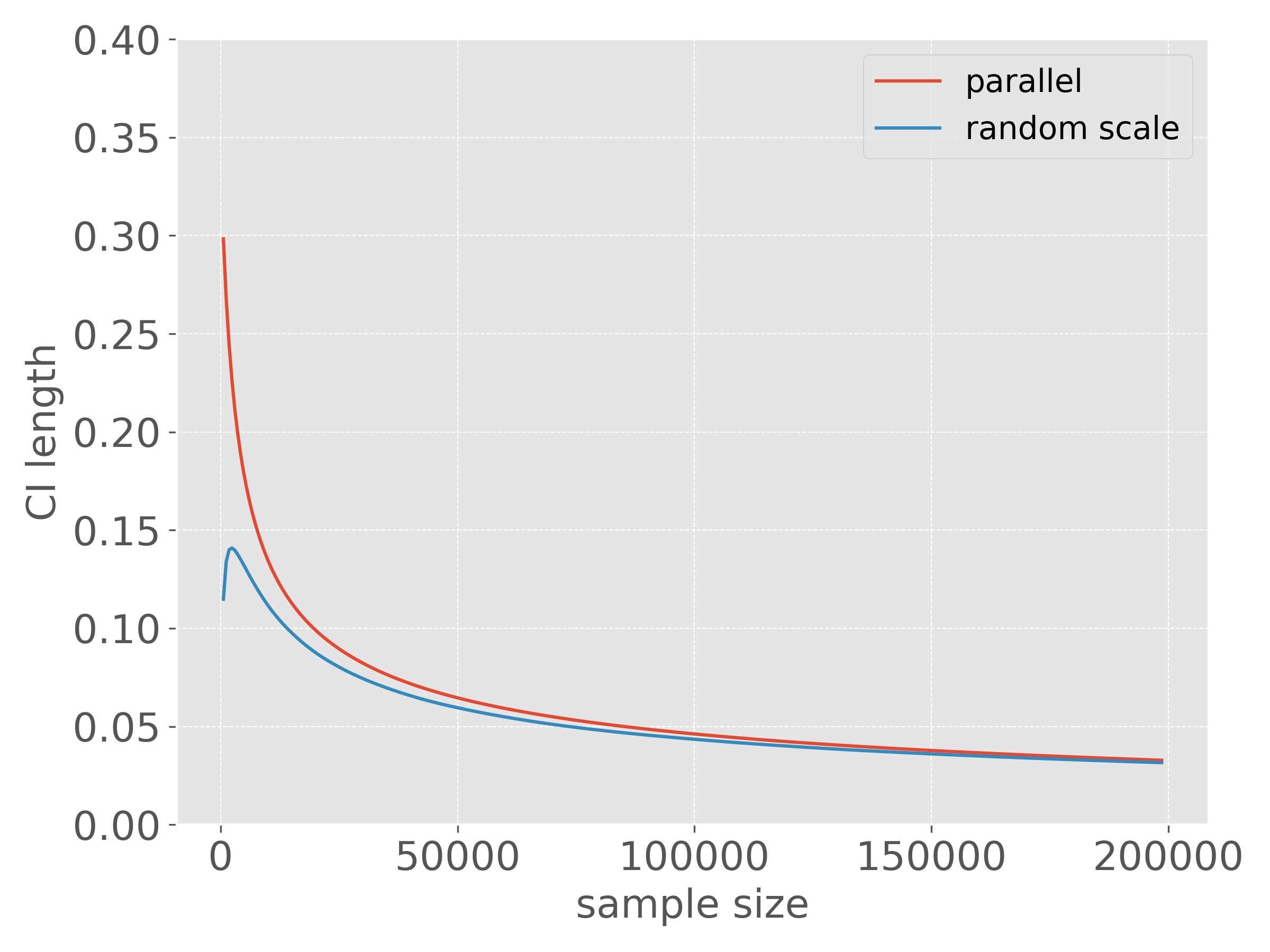}
}
\subfigure[$\alpha = 0.01$]{
	\includegraphics[width=0.33\textwidth]{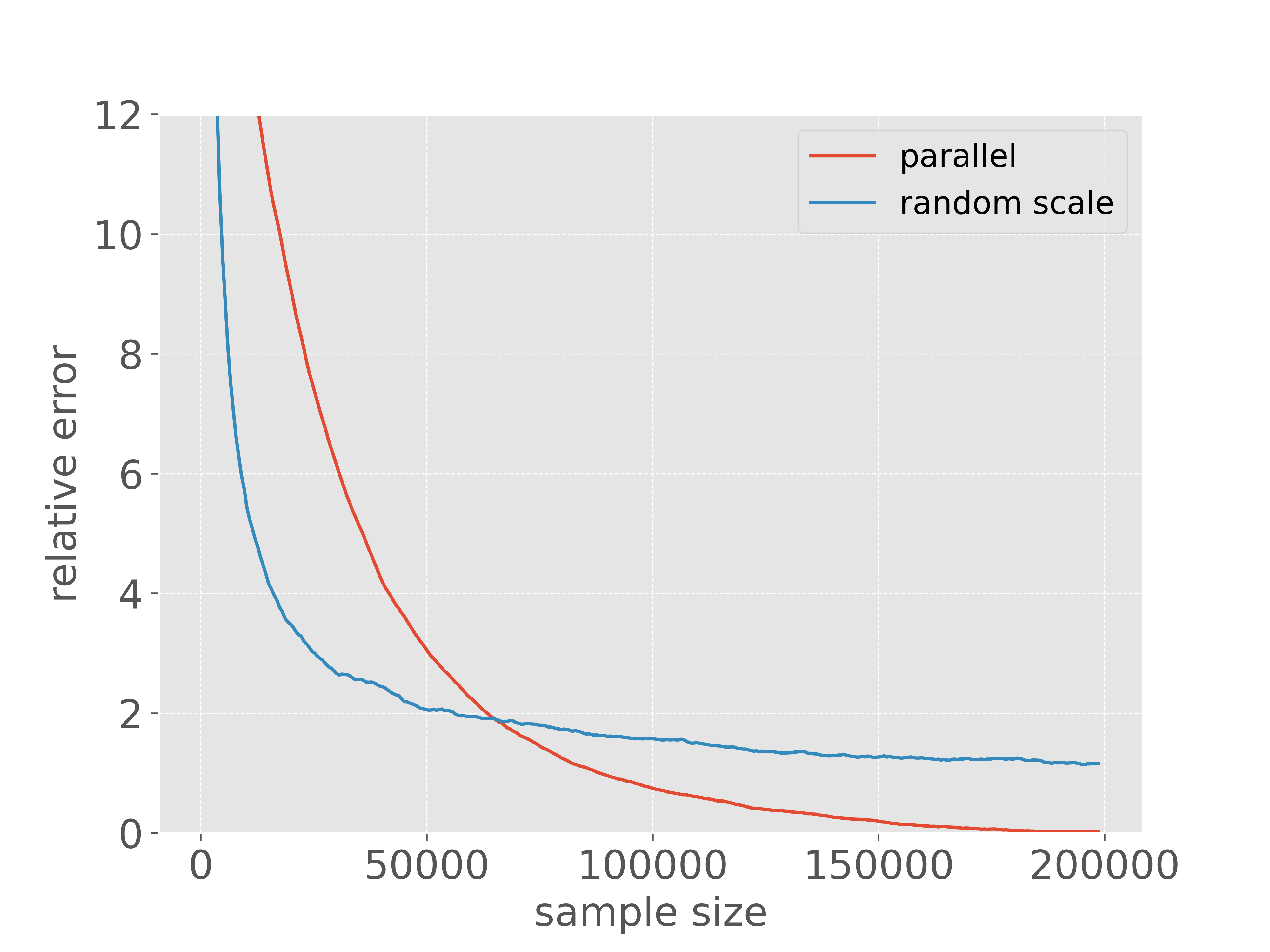} 
	\includegraphics[width=0.33\textwidth]{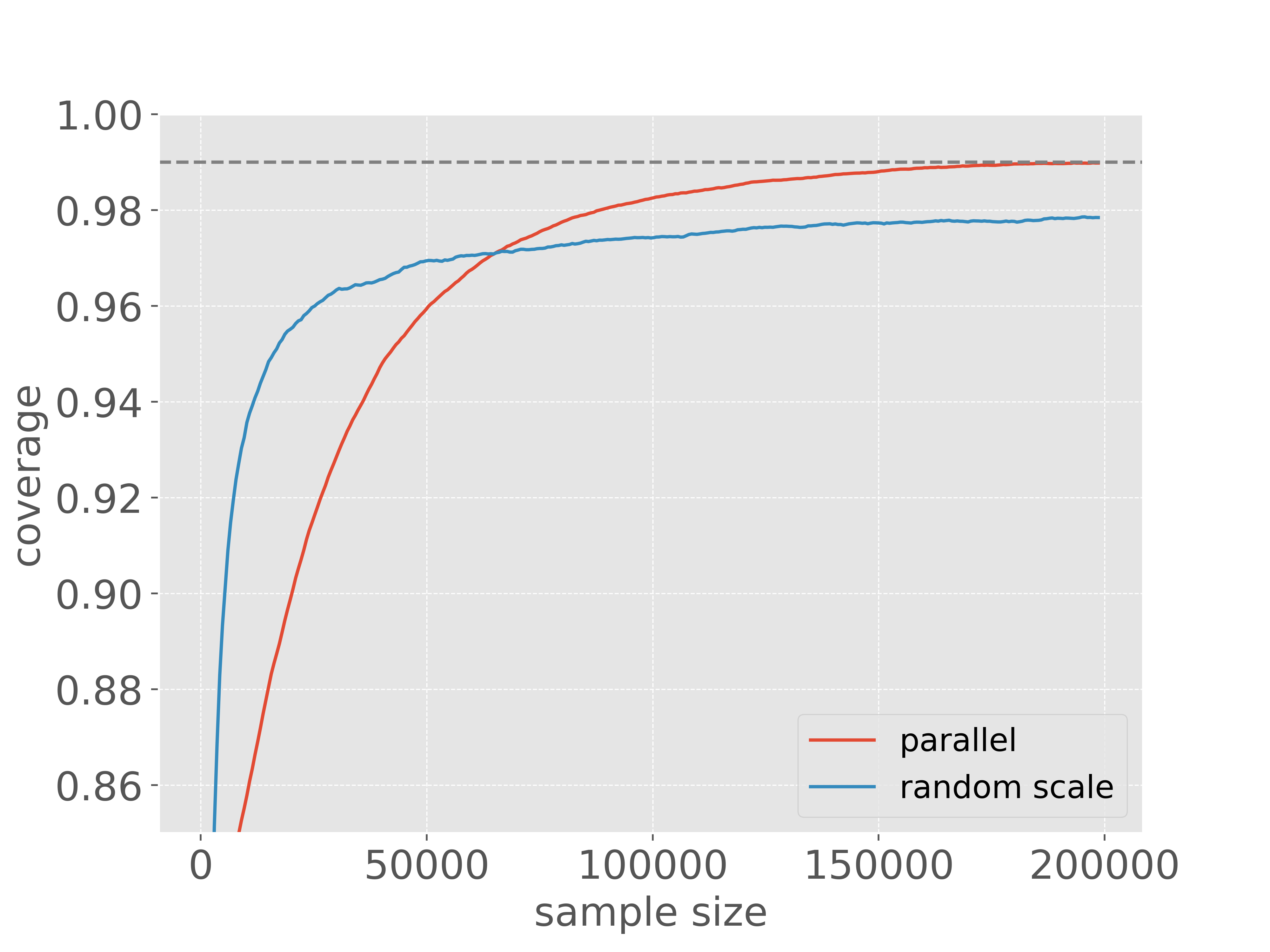} 
	\includegraphics[width=0.33\textwidth]{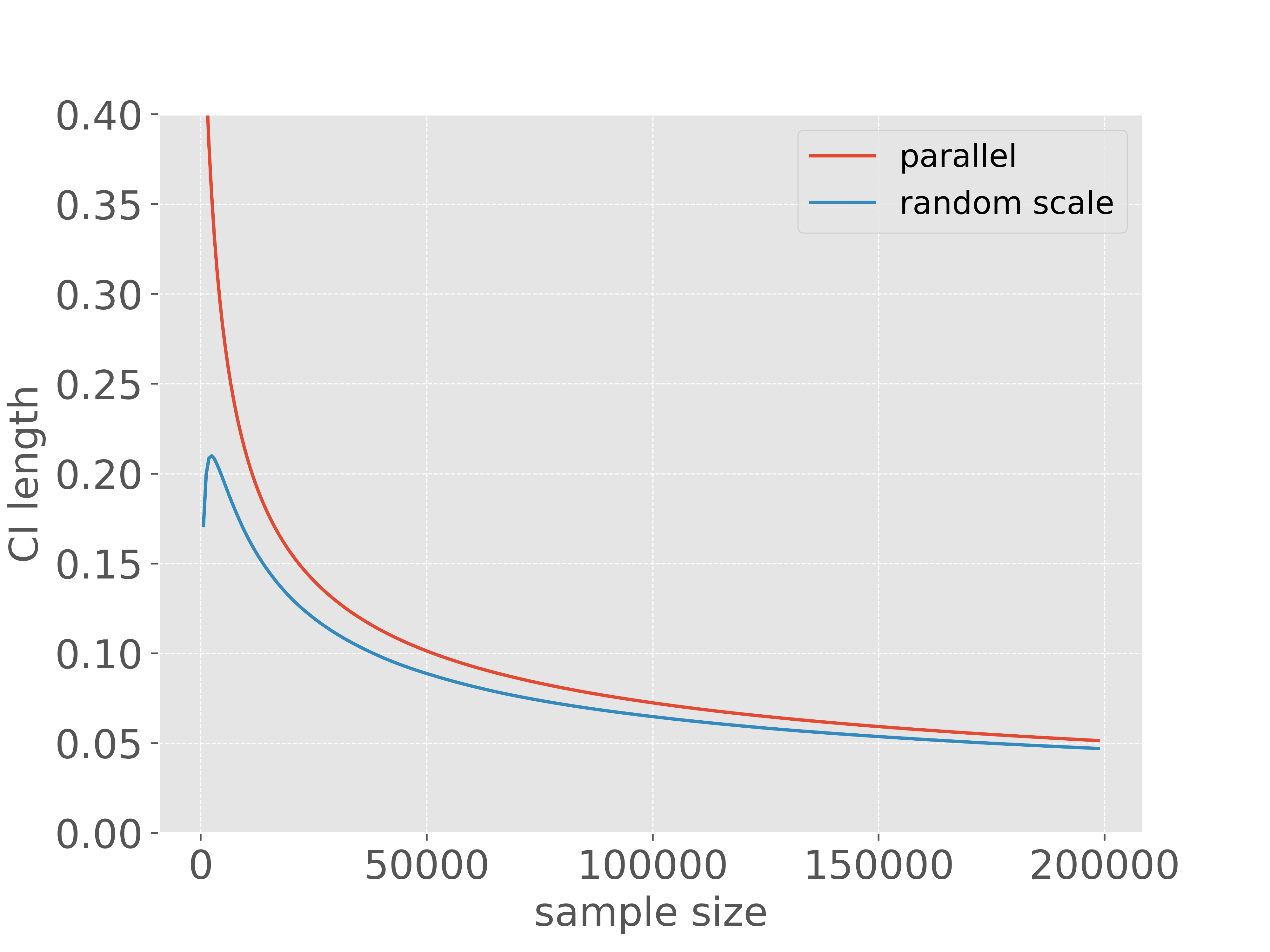}
}
\subfigure[$\alpha = 0.001$]{
	\includegraphics[width=0.33\textwidth]{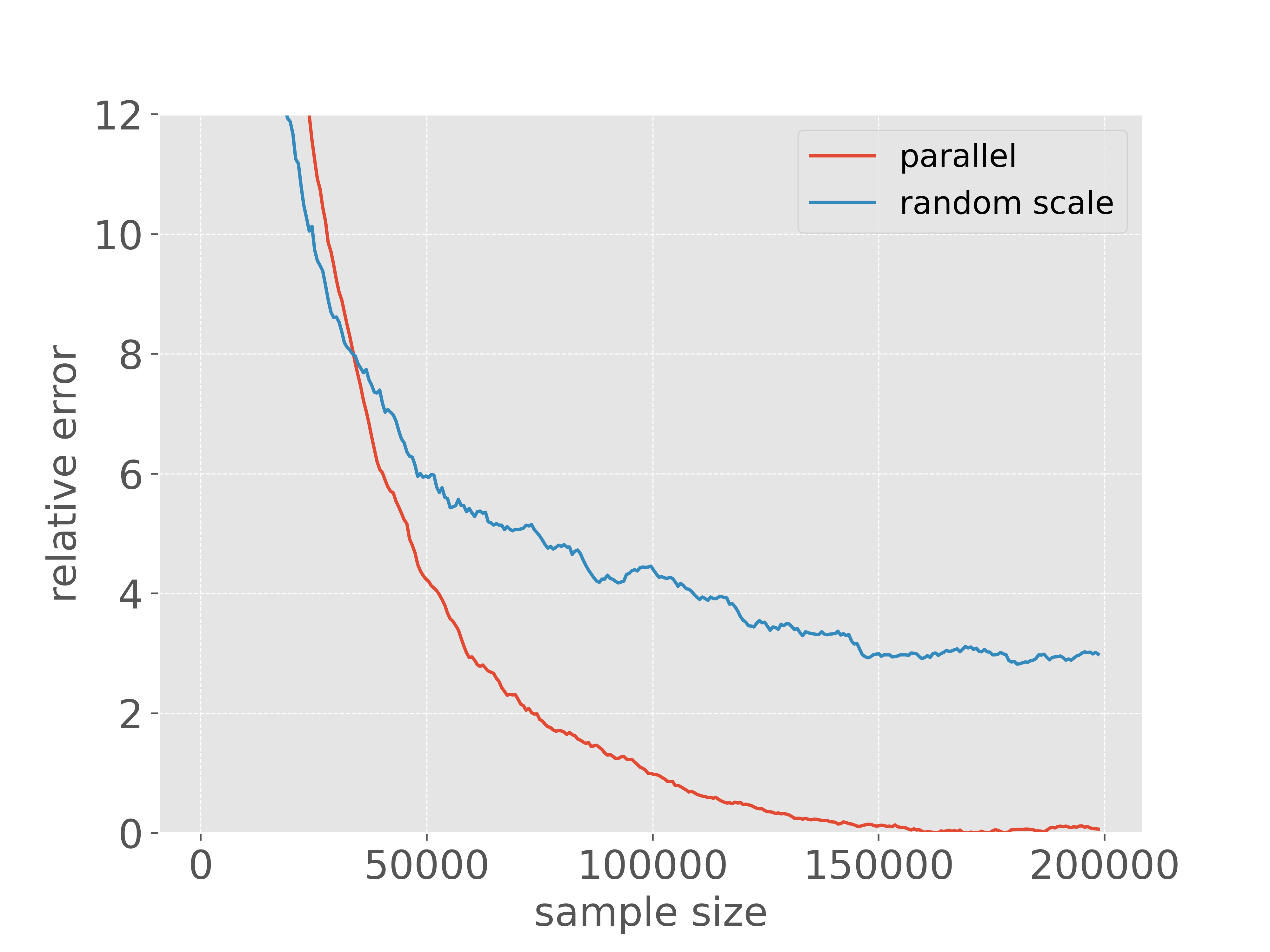} 
	\includegraphics[width=0.33\textwidth]{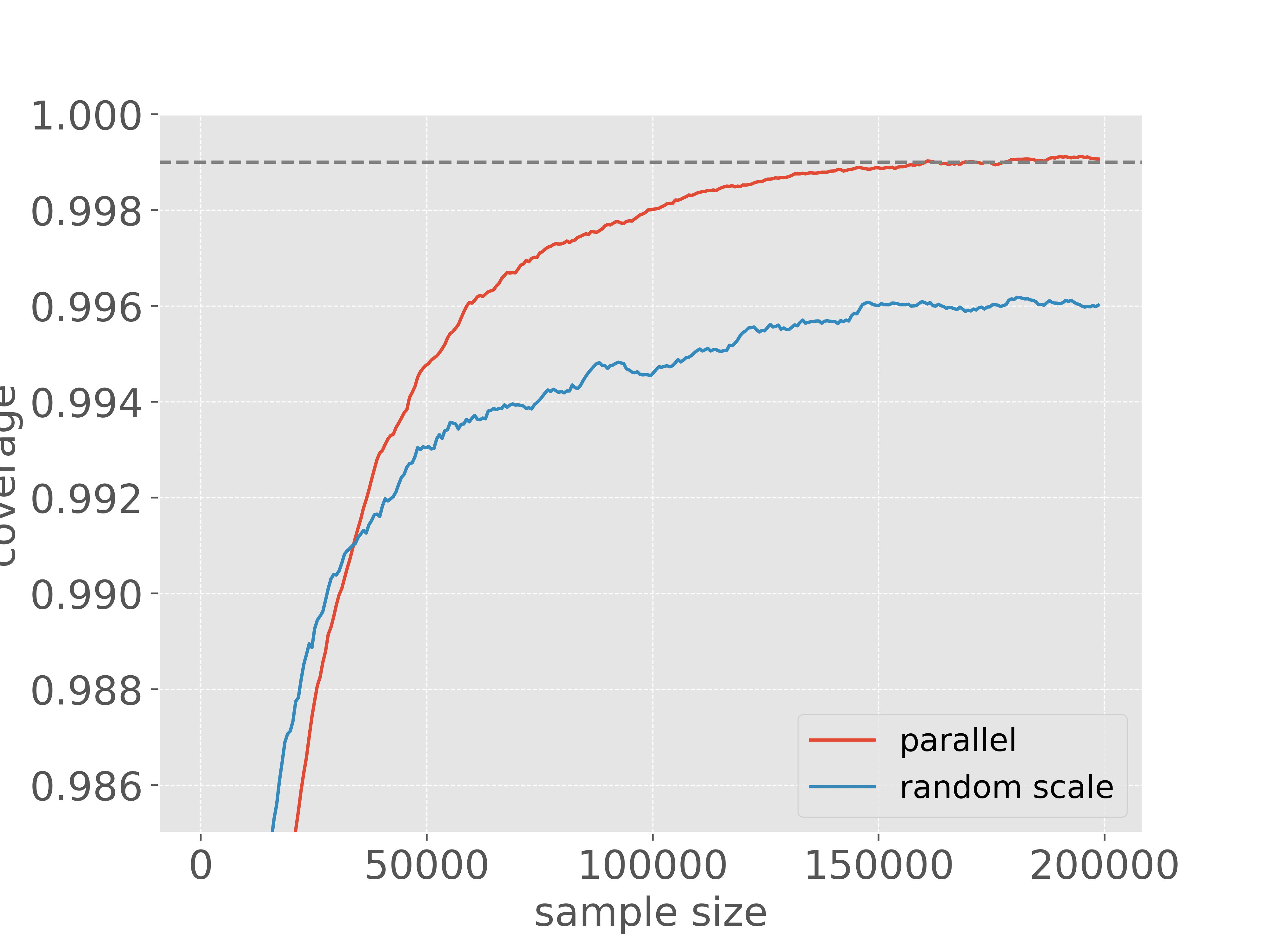} 
	\includegraphics[width=0.33\textwidth]{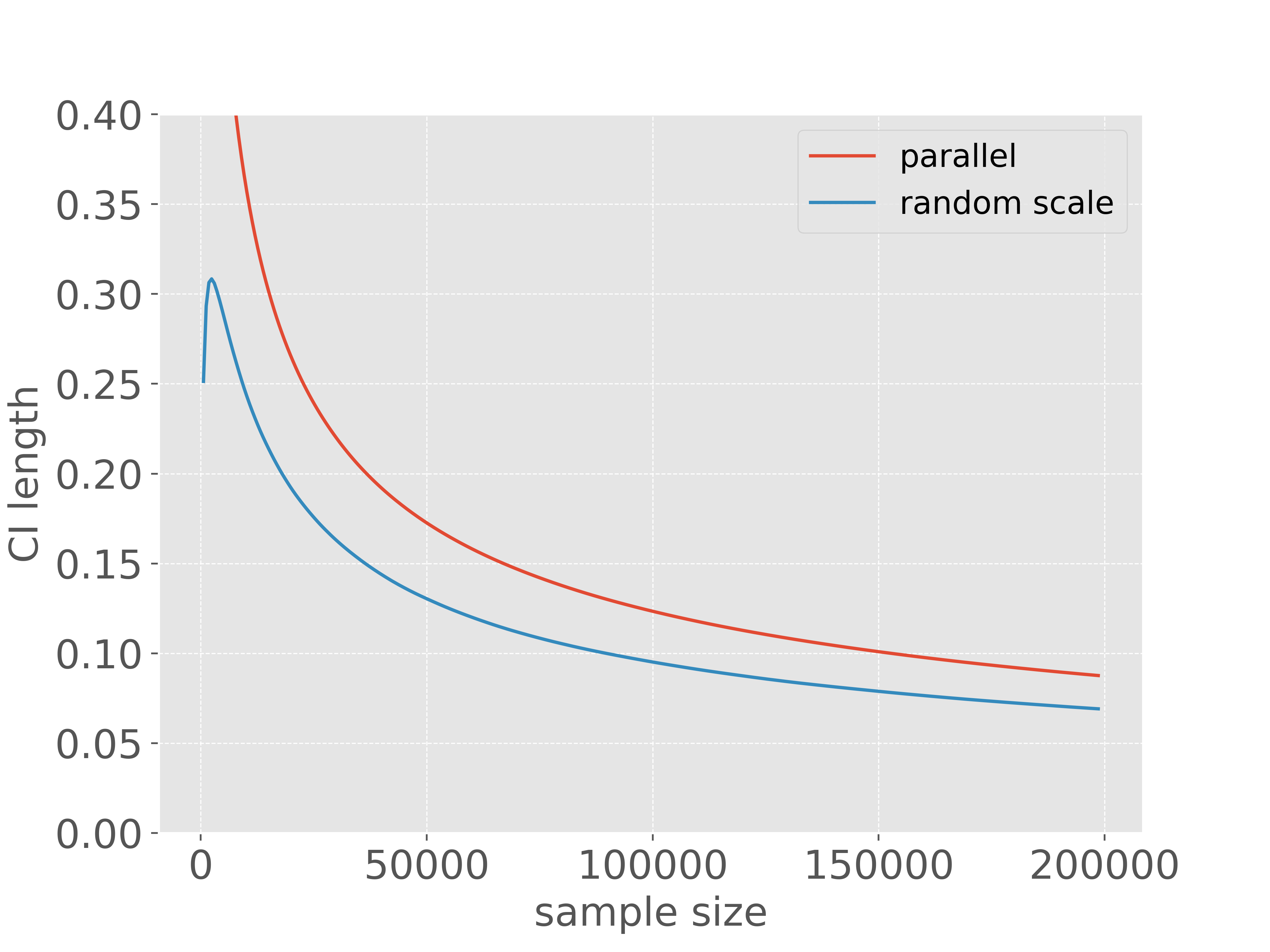}
}
\caption{Logistic Regression $d =20$. Left: relative error of coverage; Middle: empirical coverage; Right: length of confidence intervals.}
\label{fig:logistic_d20}
\end{figure}

\begin{figure}
\centering  
\subfigure[Linear regression]{
	\includegraphics[width=0.4\textwidth]{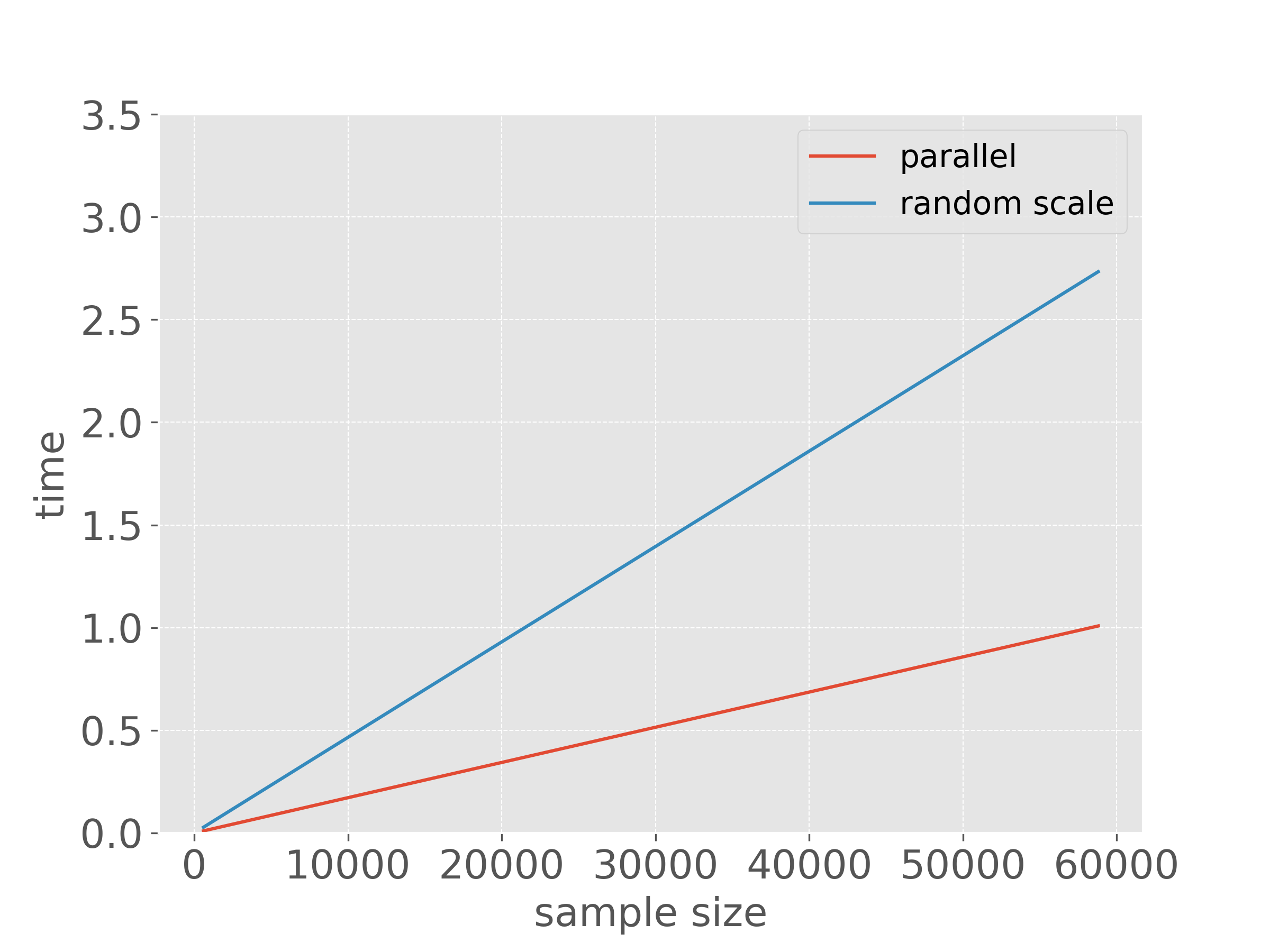}}  
\subfigure[Logistic regression]{
	\includegraphics[width=0.4\textwidth]{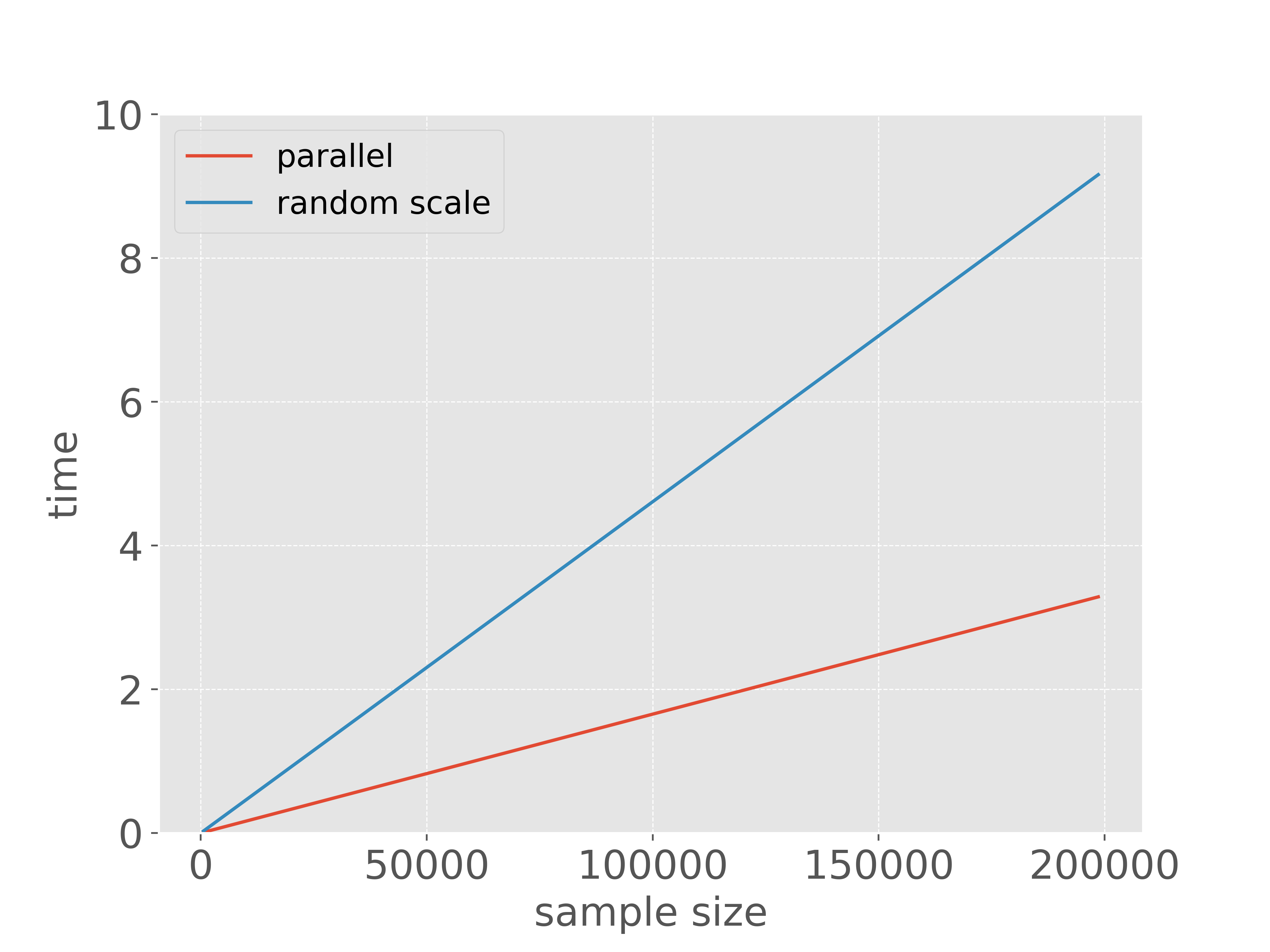}} 
\caption{Computation time ($d = 20$).}
\label{fig:time_d20}
\end{figure}

In Figures~\ref{fig:linear_d20} and \ref{fig:logistic_d20}, we present results for confidence intervals where the nominal coverage probability is set at $0.95$, $0.99$, and $0.999$, i.e., $\alpha = {0.05, 0.01, 0.001}$ for both linear regression and logistic regression  $d = 20$. We plot the relative error of coverage, the empirical coverage rate, and the length of the confidence intervals. We also compare the running time of a single trial in Figure~\ref{fig:time_d20}. More results are summarized in Appendix B in the supplementary material.
The relative error of our parallel method converges to zero faster than that of the random scaling method in all cases. The advantage becomes more obvious as the confidence levels increase. Note that in logistic regression with $d = 20$, both methods exhibit relatively large errors when the sample size is small. In this case, the parallel method converges more slowly, which can be attributed to the fact that data splitting in the parallel run exacerbates the issue of a small sample size. However, as the sample size increases, the convergence rate of the parallel method improves and eventually surpasses that of the random scaling method. 
The lengths of the confidence intervals are comparable between the two methods, with those derived from the parallel method being slightly larger. 
We also observe that our parallel method has a distinct advantage in terms of computing time, as it does not necessitate additional computations at each iteration, such as updating a $d$ by $d$ matrix, which is required by the random scaling method. Apart from the SGD update, the only additional computation needed for inference is calculating a sample covariance matrix or a sample variance (for the linear functional). This computation is minimal, making the inference process almost cost-free.  
The advantage in computing becomes even more significant when  
utilizing parallel computing across different cores.

\noindent{\bf Comparison with oracle.}
We also compare our method to the oracle approach, which constructs confidence intervals using the true limiting covariance matrix $\Sigma$ and the principles of asymptotic normality. The oracle method is given by: 
\[\hat{\textnormal{CI}}_{n, \textnormal{oracle}} = \left[\upsilon^{\top} \bar{x}_{n} -z_{1-\alpha/2}\sqrt{\frac{\upsilon^{\top}\Sigma\upsilon}{n}},\  \upsilon^{\top} \bar{x}_{n}+z_{1-\alpha/2}\sqrt{\frac{\upsilon^{\top}\Sigma\upsilon}{n}}\right].\]
In the linear regression model described above, the limiting covariance matrix $\Sigma = {\bf{I}}_{d}$. 
We focus on linear regression here since the true limiting covariance matrix is straightforward to compute. 
As illustrated in  
Figure~\ref{fig:linear_d20}
the coverage achieved by the parallel method surpasses that of the oracle method. This could be attributed to a discrepancy between the finite sample covariance matrix of ASGD and the limiting covariance $\Sigma$. Employing an asymptotic pivotal statistic helps to mitigate the impact of this difference. For further details and discussion on this topic, refer to Section~\ref{sec:thm_asgd}.

\subsection{Non-convex objectives}\label{sec:exp_non-convex}
\revise{
We next consider a non-convex optimization problem studied in \citep{Yunonconvex}, where SGD with a constant learning rate is known to exhibit asymptotic normality. This setting allows us to evaluate our method in a non-convex regime while maintaining a theoretical benchmark for comparison.

\cite{Yunonconvex} shows that SGD with a constant learning rate is asymptotically normally distributed around the expectation under a unique invariant distribution, provided that the  objective function satisfies a dissipativity condition.  Specifically, for a test function $\phi: \R^{d}\rightarrow R$, there exists a  unique stationary distribution $\pi$ such that
\[\sqrt{n}\left(\frac{1}{n}\sum_{i=1}^{n}\phi(x_i) - \pi(\phi)\right)\Rightarrow \mathcal{N}(0, \sigma^2_{\pi}(\phi)),\]
where $\{x_i\}_{i\ge 1}$ is the minibatch SGD sequence with a constant learning rate,  $\pi(\phi) = E_{\pi}(\phi(x))$  and $\sigma^2_{\pi}(\phi) = \lim_{n\rightarrow\infty}\E_{\pi}[n(\frac{1}{n}\sum_{i=1}^{n}\phi(x_i) - \pi(\phi))^2]$. 

We now apply our parallel inference framework. For machine $k\in\{1, \dots, K\}$,  we compute $\hat\phi_i^{(k)} = \frac{1}{i}\sum_{j=1}^{i}\phi(x_j^{(k)})$,   where $x_{i}^{(k)}$ denotes the $i$-th SGD iterate on the $k$-th machine. When inference is required at iteration $n$, the confidence interval is constructed as $$\left[\bar{\phi}_{K,n} -\frac{t_{1-\alpha/2, K-1}\hat\sigma}{\sqrt{K}}, \bar{\phi}_{K, n} +\frac{t_{1-\alpha/2, K-1}\hat\sigma}{\sqrt{K}}\right],$$ where $\bar{\phi}_{K, n} = \frac{1}{K}\sum_{k=1}^{K}\hat\phi_n^{(k)}$, and $\hat\sigma = \frac{1}{K-1}\sum_{k=1}^{K}(\hat\phi_n^{(k)} - \bar{\phi}_{K, n})^2$.
 
We follow the experimental setup described in \cite{Yunonconvex}. Specifically, let
$\xi_{i} = (a_i, b_i)$, where $a_i\in\mathbb{R}^{d}$ with $d = 10$, and  each coordinate generated from $\textnormal{Bernoulli}(\frac{1}{2},\{-\frac{1}{\sqrt{d}}, \frac{1}{\sqrt{d}}\})$. The response $b_i = a_{i}^{T}\theta + e_i$, where each coordinate of $\theta$ is drawn from  $\textnormal{Unif}(0, 1)$ and then held fixed throughout the experiment. The noise term $e_i$ follows a Student-$t$ distribution with $df = 10$. The non-convex objective function is
\[f(x, \xi_i) = \log(1+(b_i - a_i^{T}x)^2)+\frac{\lambda}{2}\|x\|^2,\]
where $\lambda$ is a small value.
Minibatch SGD with constant learning rate  is applied as:
$x_{i+1} = x_i - \frac{\eta}{b}\sum_{j=1}^{b}\nabla f(x_i, \xi^{(i)}_j)$,
with $\eta = 0.3$, minibatch size $b = 2$ and initialization $(1,\dots, 1)^{T}$. The test function is chosen as $\phi(x) = \|x\|$. 

We use Monte Carlo simulation to approximate the stationary mean $\pi(\phi)$ and construct confidence intervals at nominal levels $95\%, 99\%, 99.9\%$ with $K = 6$. For comparison, we also implement the subsampling quantile method of \cite{Yunonconvex}, which utilizes a single chain and corresponds to the first approach discussed in Section 1.1; we refer readers to their paper for further technical details.  As shown in Figure \ref{fig:nonconvex}, our parallel inference method achieves the nominal coverage from the very early stages of the algorithm, whereas the subsampling quantile method converges much more slowly. This difference is even more pronounced when examining the normalized coverage error. 
Thus, our parallel inference procedure demonstrates substantially faster stabilization of coverage in this non-convex setting.

\begin{figure}
\subfigure[$\alpha = 0.05$]{
\includegraphics[width=0.33\textwidth]{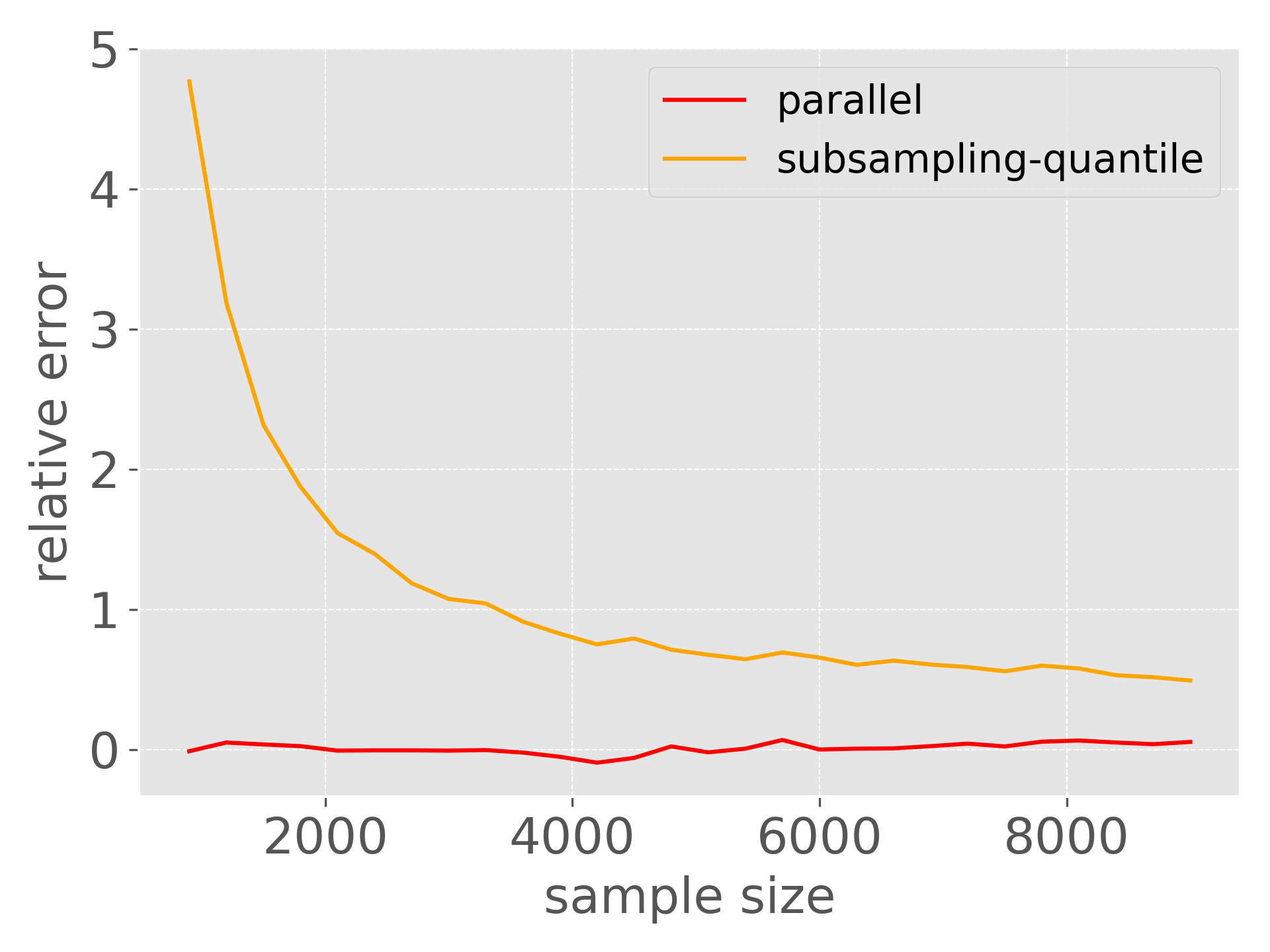}
	\includegraphics[width=0.33\textwidth]{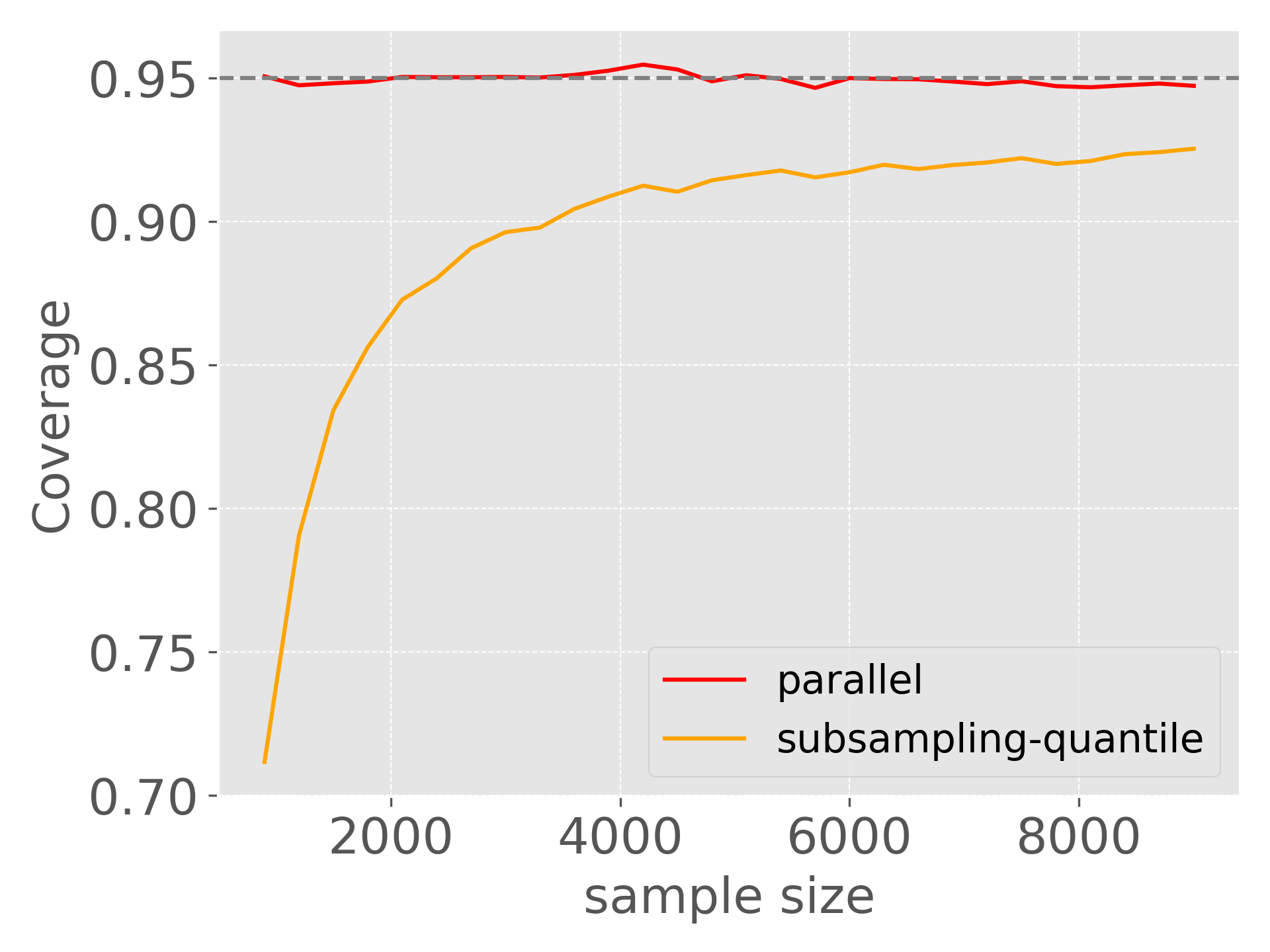} 
	\includegraphics[width=0.33\textwidth]{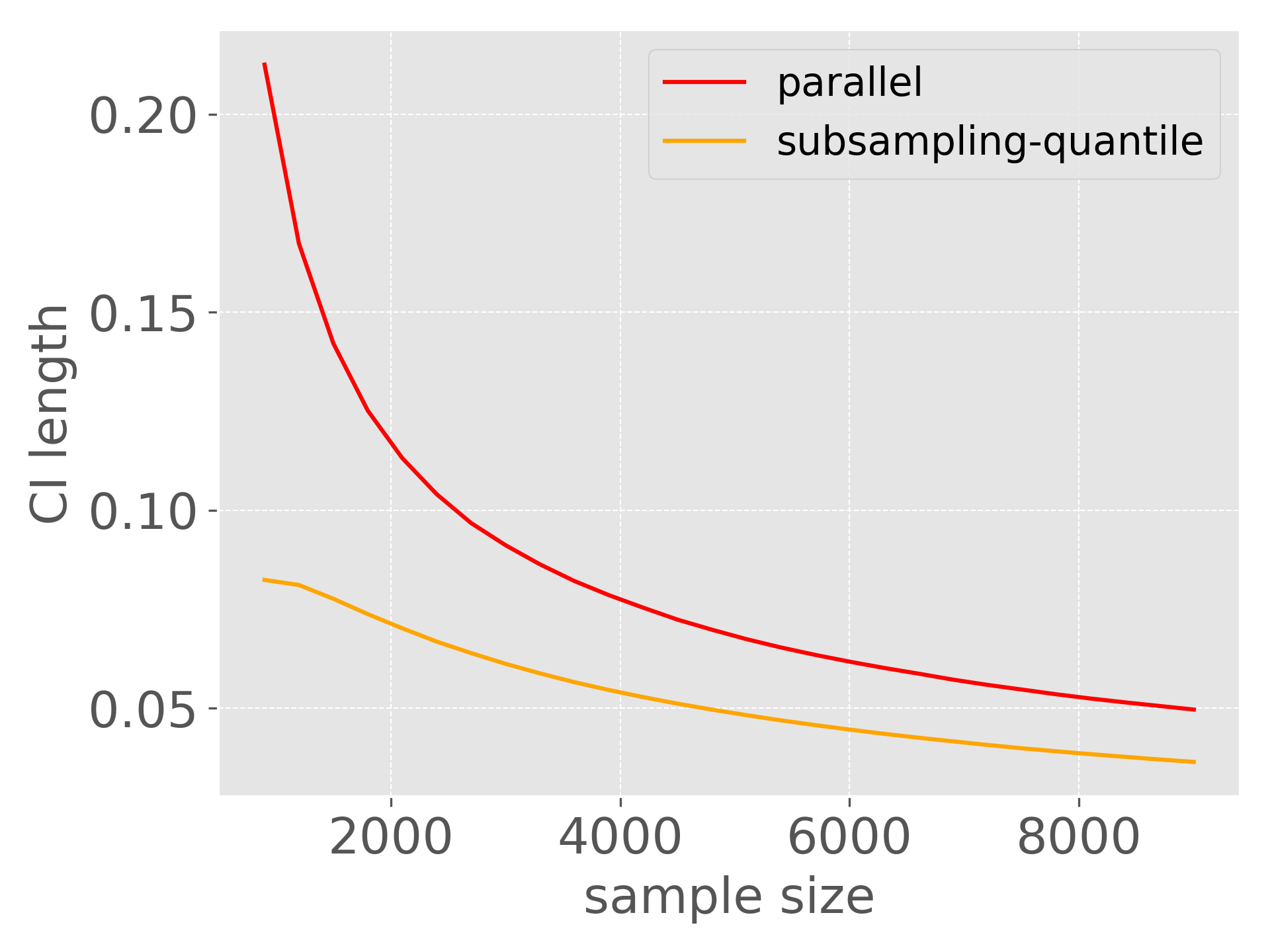}
}
\subfigure[$\alpha = 0.01$]{
	\includegraphics[width=0.33\textwidth]{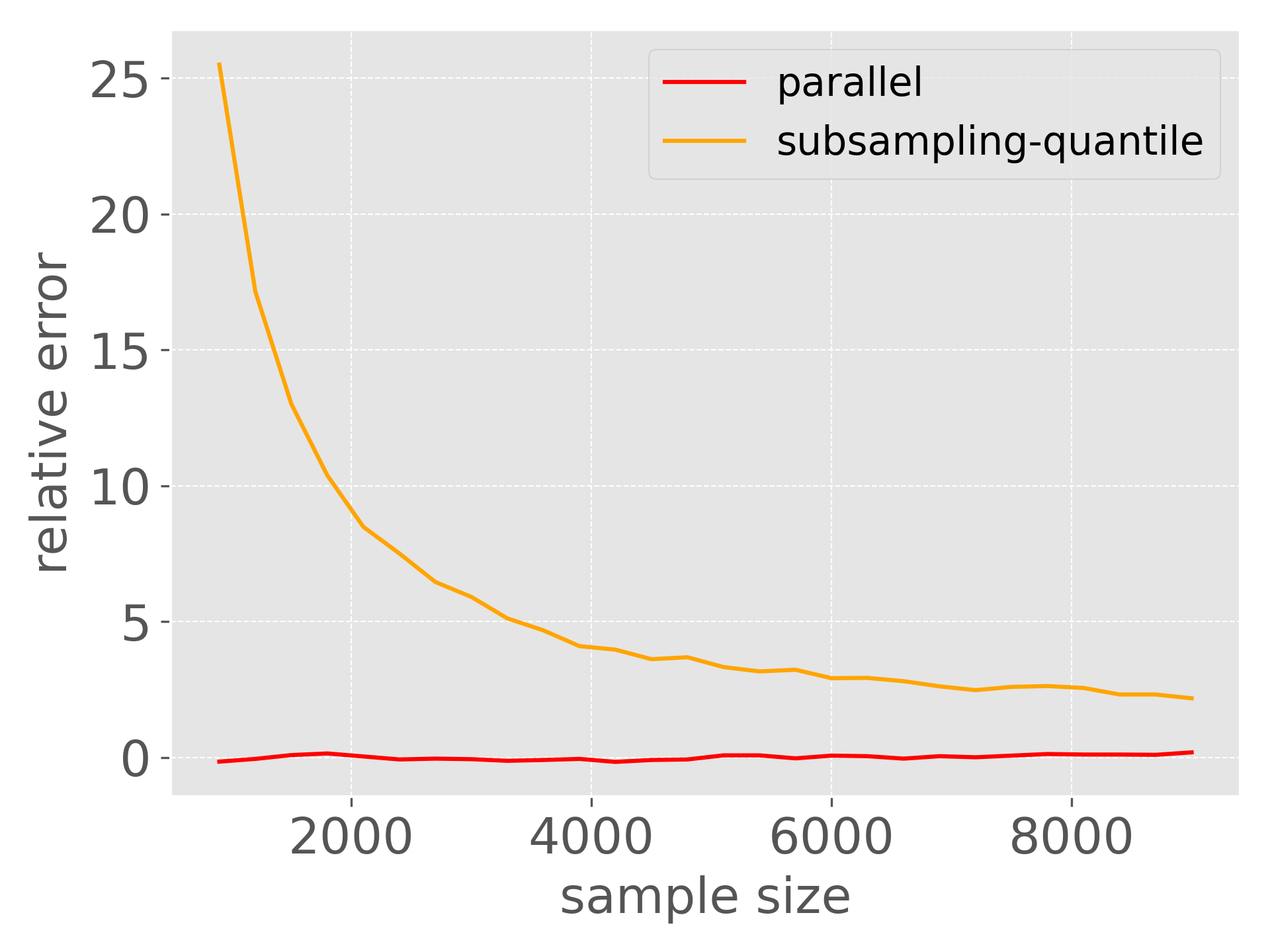} 
	\includegraphics[width=0.33\textwidth]{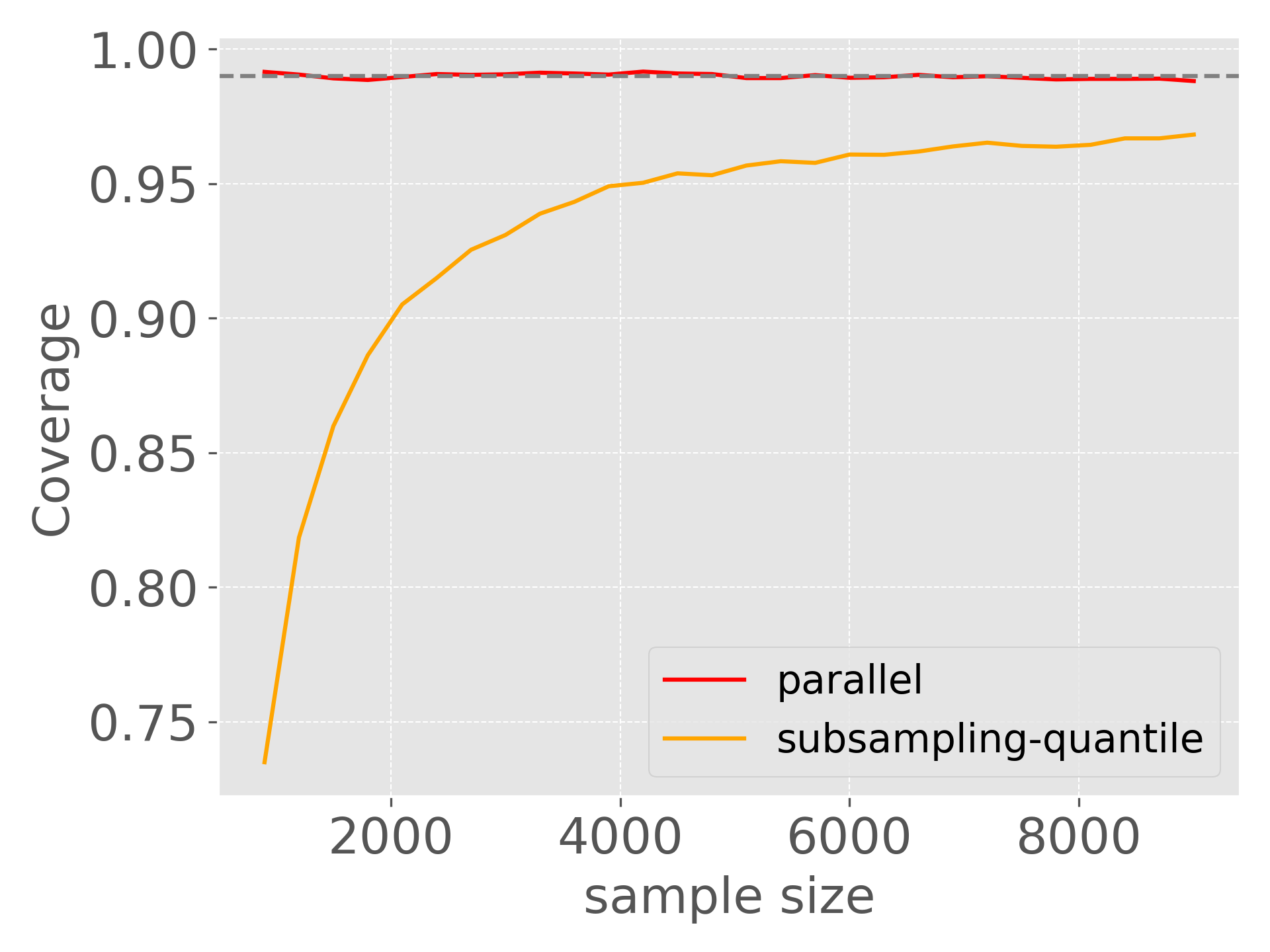} 
	\includegraphics[width=0.33\textwidth]{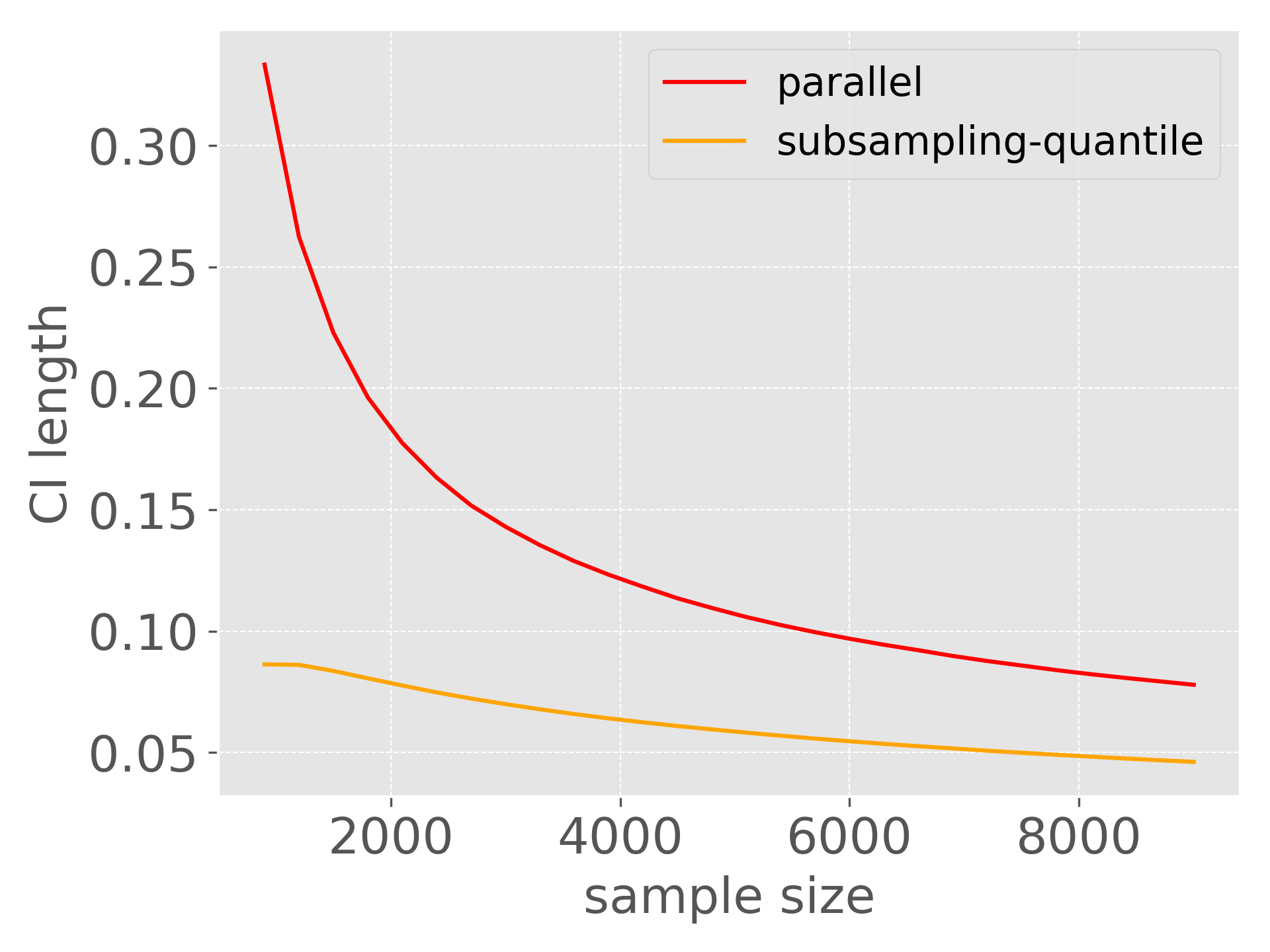}
}
\subfigure[$\alpha = 0.001$]{
\includegraphics[width=0.33\textwidth]{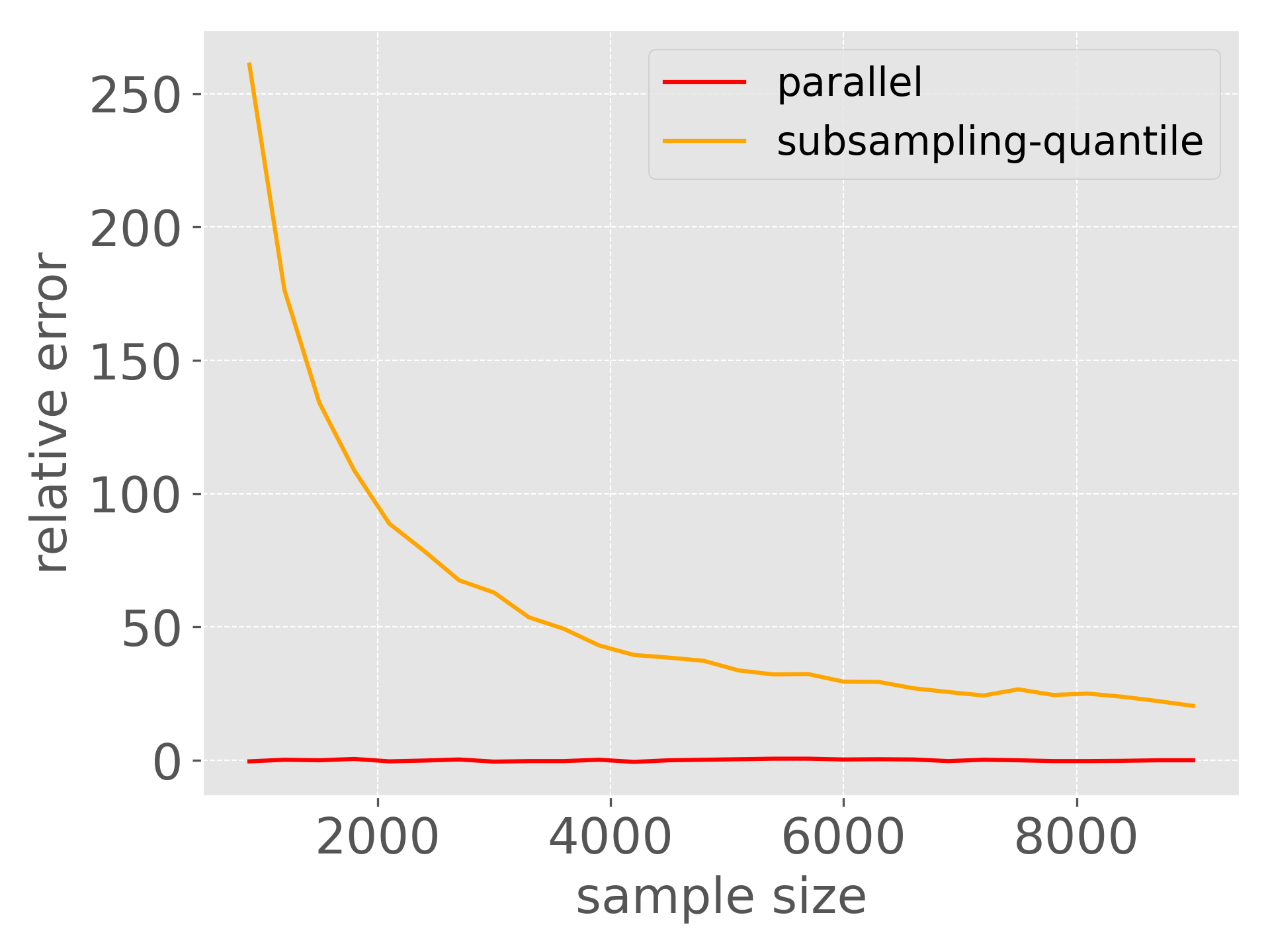} 
	\includegraphics[width=0.33\textwidth]{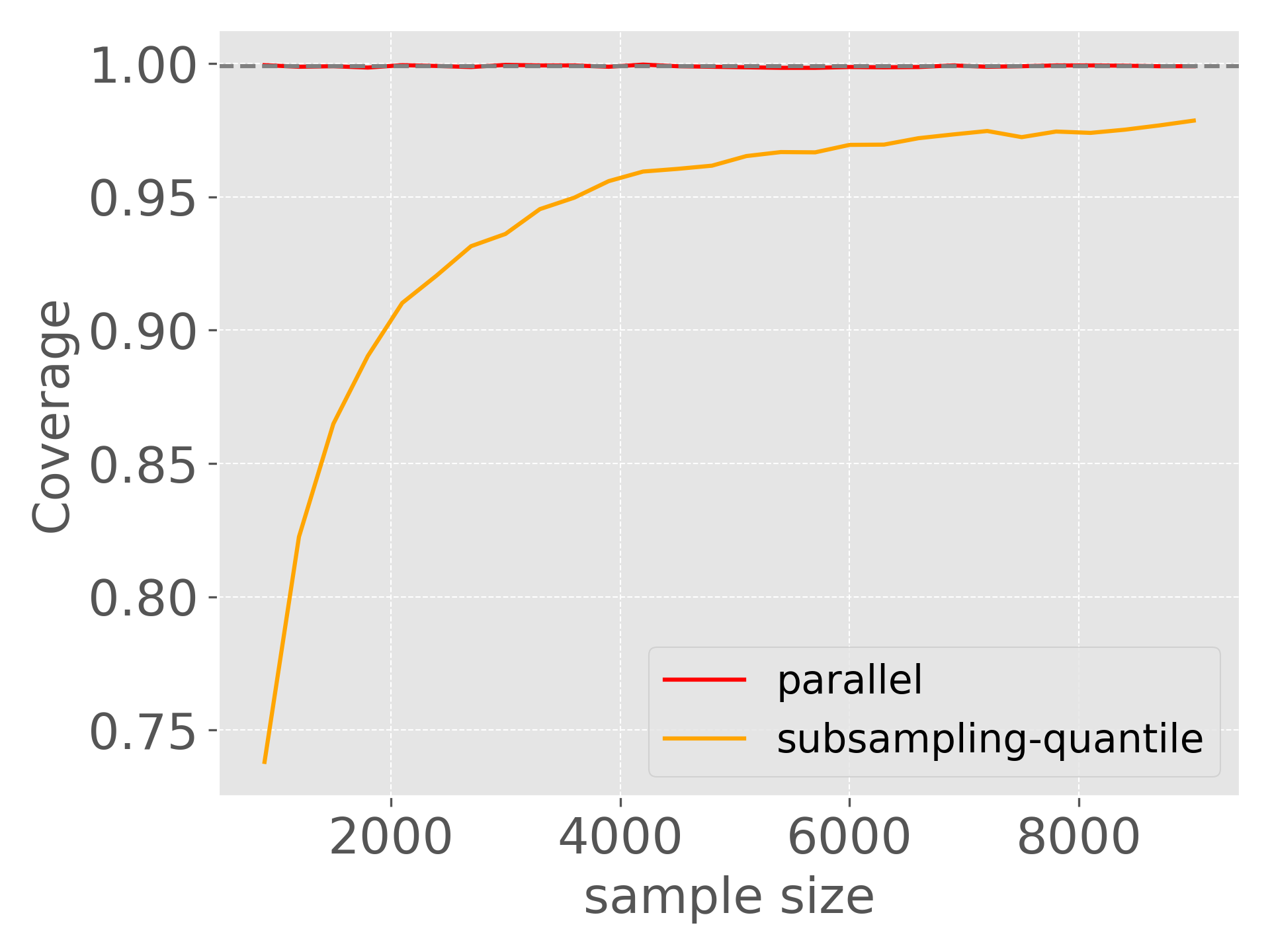}
	\includegraphics[width=0.33\textwidth]{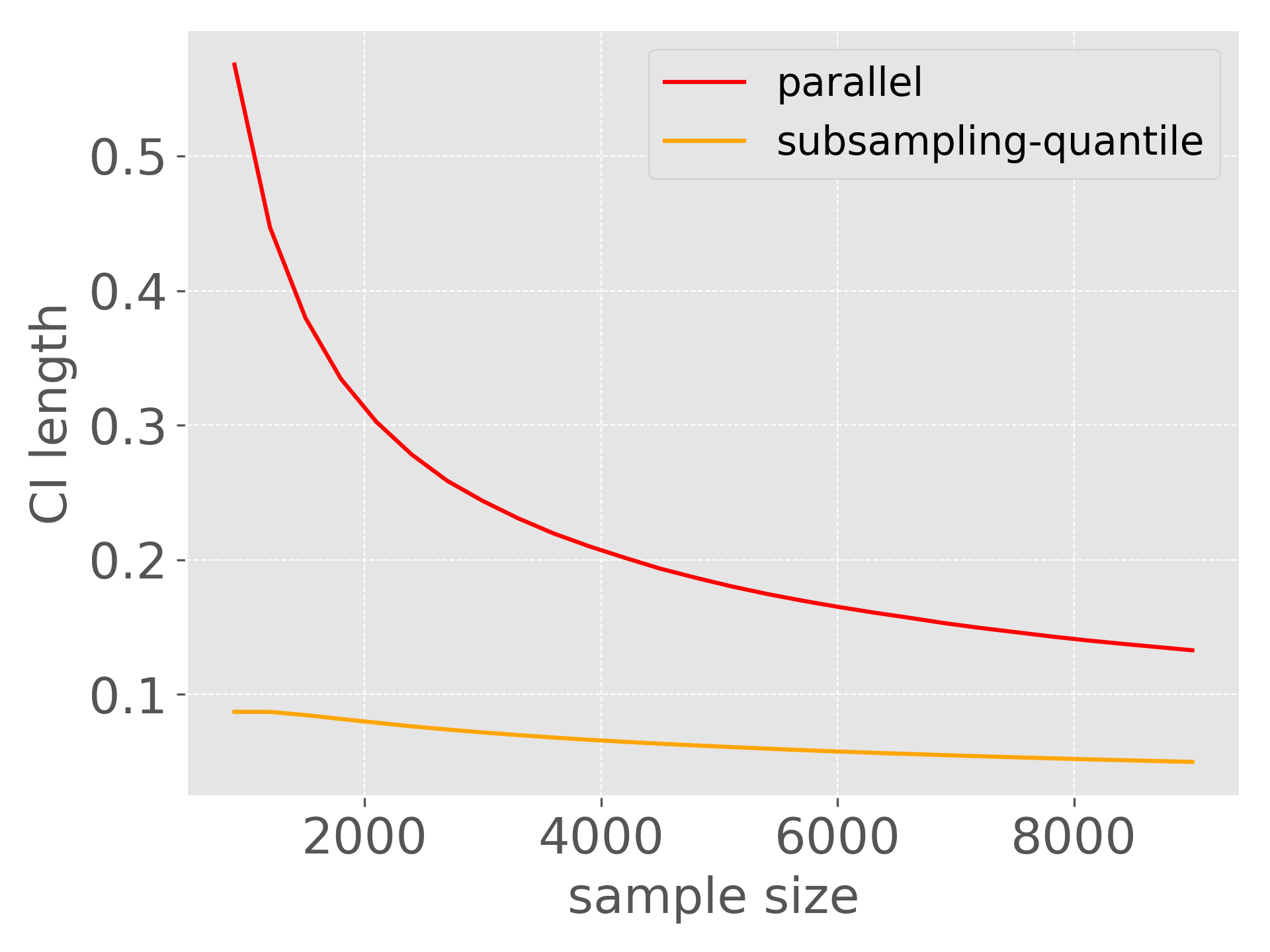}
}
\caption{Nonconvex setting. Left: relative error of coverage; Middle: empirical coverage; Right: length of confidence intervals.}
\label{fig:nonconvex}
\end{figure}

\subsection{Online source localization}
\label{sec:exp_loc}
\revise{
Finally, we consider an online source localization problem via pseudorange measurements \citep{bancroft2007algebraic}, which provides a more realistic application scenario. This problem involves a nonlinear and non-convex objective function and arises in practical settings such as signal processing and sensor networks.

 Source localization has gained significant prominence in signal processing due to its essential role in global positioning systems (GPS), the Internet of Things (IoT), and real-time asset tracking. This application is particularly well-suited for our parallel inference framework for several reasons. First, localization typically relies on continuous streams of measurements—such as pseudorange, time-of-arrival (TOA), or time-difference-of-arrival (TDOA) \citep{beck2008exact, liu2017local}—necessitating high-frequency, real-time updates of the source location. Second, the data is naturally distributed across multiple sensors or base stations, which aligns perfectly with our parallel architecture where each sensor stream can be processed as an independent stochastic sequence. By leveraging SGD and its variants \citep{rabbat2004decentralized, abanto2018tdoa} within our framework, we can efficiently handle these massive data streams while providing rigorous, real-time statistical uncertainty quantification for the estimated coordinates.

Consider a sequence of measurements $\{(P_i, D_i)\}_{i\ge 1}$ received over time. Here,  $P_i = (P_{i1},\ldots,P_{id})\in \mathbb{R}^d$ denotes the coordinates of the sensor providing the $i$-th piece of information, and $D_i\in \R$ represents the corresponding noisy pseudorange observation. Let $x^* = (\theta^*,b^*)$ denote the target parameter, where $\theta^*  \in \mathbb{R}^d$ is the source’s coordinate vector, and $b^* \in \mathbb{R}$ is the true clock offset or bias.   The objective function is given by
$$F(x) = F(\theta,b) = \frac{1}{2} \E\left( D_i -  \|P_i-\theta\| - b  \right)^2 = \frac{1}{2} \E\left( D_i -  \sqrt{\sum_{j=1}^d (P_{ij}-\theta_j)^2} - b  \right)^2$$
Due to the negative Euclidean distance in the loss function, this formulation results in a non-convex and non-smooth optimization problem. In practice, the spatial dimension is typically $d = 2$ or $3$, although existing literature also explores high-dimensional variants \cite{beck2008exact}.

For the experimental setup, we set the true parameter as $x^* = (\theta^{*\top}, b^*)^{\top} =(50, 50, 50, 5)^{\top}$, representing a 3D coordinate of $(50, 50, 50)$ in meters with a 5-meter equivalent clock bias. We generate a sequence of i.i.d.~random samples $\{(P_i, D_i)\}_{i\ge 1}$, where each coordinate of $P_i$ independently follows a uniform distribution over [0,100]. The distance observation $D_i = \|P_i-\theta^*\|+ b^*+\epsilon_i$ where $\epsilon_i$ follows $\mathcal{N}(0,\sigma^2)$ independently. We set $\sigma = 10$ to test the algorithm's performance under large noise conditions. We implement our inference framework using ASGD, following the setup in Section~\ref{sec:exp_convex}. 

 In Figure \ref{fig:trajectory}, we plot the trajectory of estimated location (left panel) and provide a detailed view along the $X$-axis (right panel), where we overlay the estimated confidence intervals with a nominal coverage $99\%$. The visual results demonstrate that our method successfully converges to the true location while explicitly quantifying the uncertainty.
Furthermore, in Figure \ref{fig:TOA results}, we report the empirical  coverage rates and average confidence interval lengths for each coordinate with a nominal coverage level of $99\%$. The results show that the parallel inference method achieves accurate coverage of the true location while maintaining sufficiently short interval lengths to remain informative.}

\begin{figure}
\centering 
\subfigure[]{\includegraphics[width=0.45\textwidth]{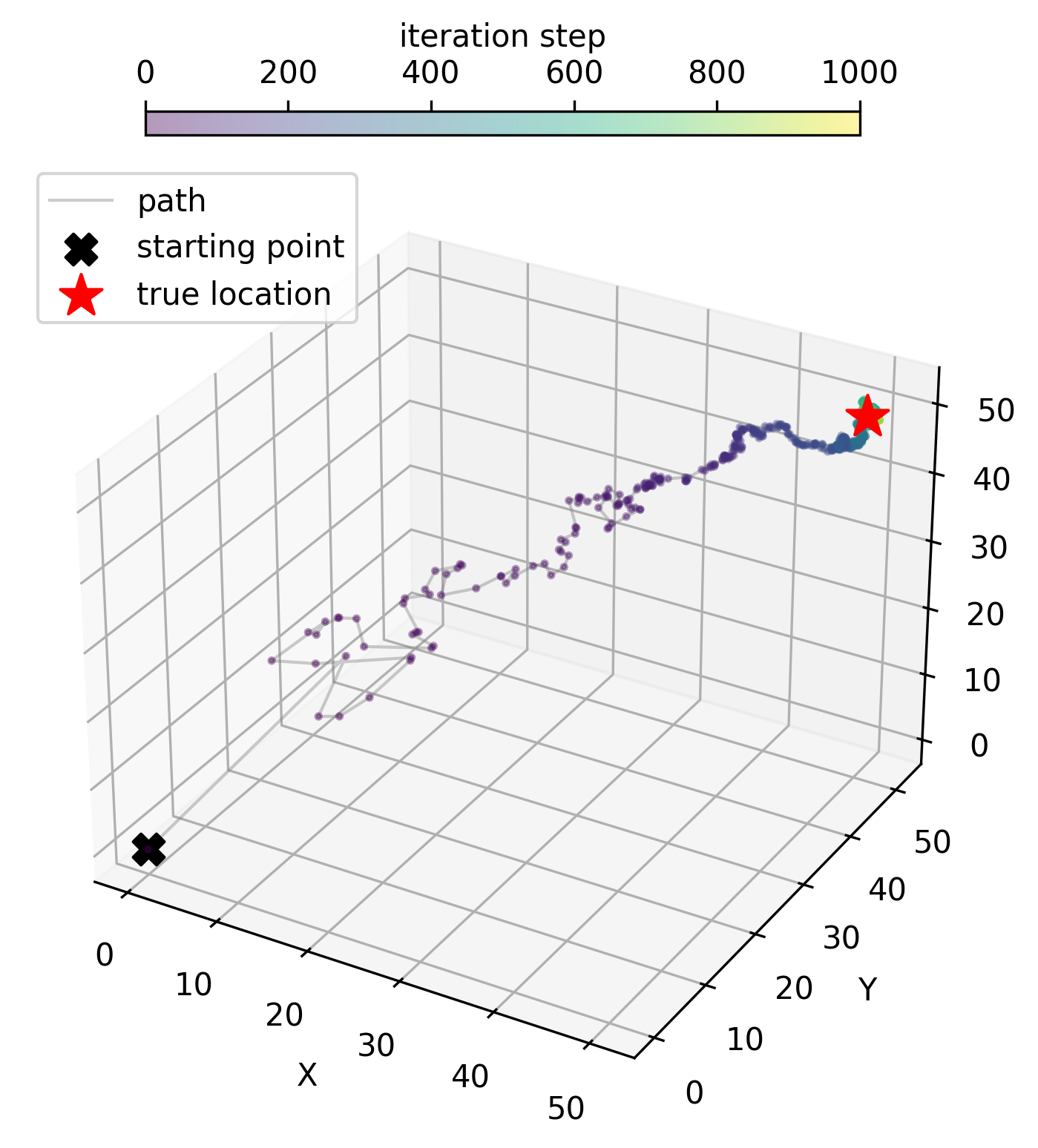}}
\subfigure[]{\includegraphics[width=0.45\textwidth]{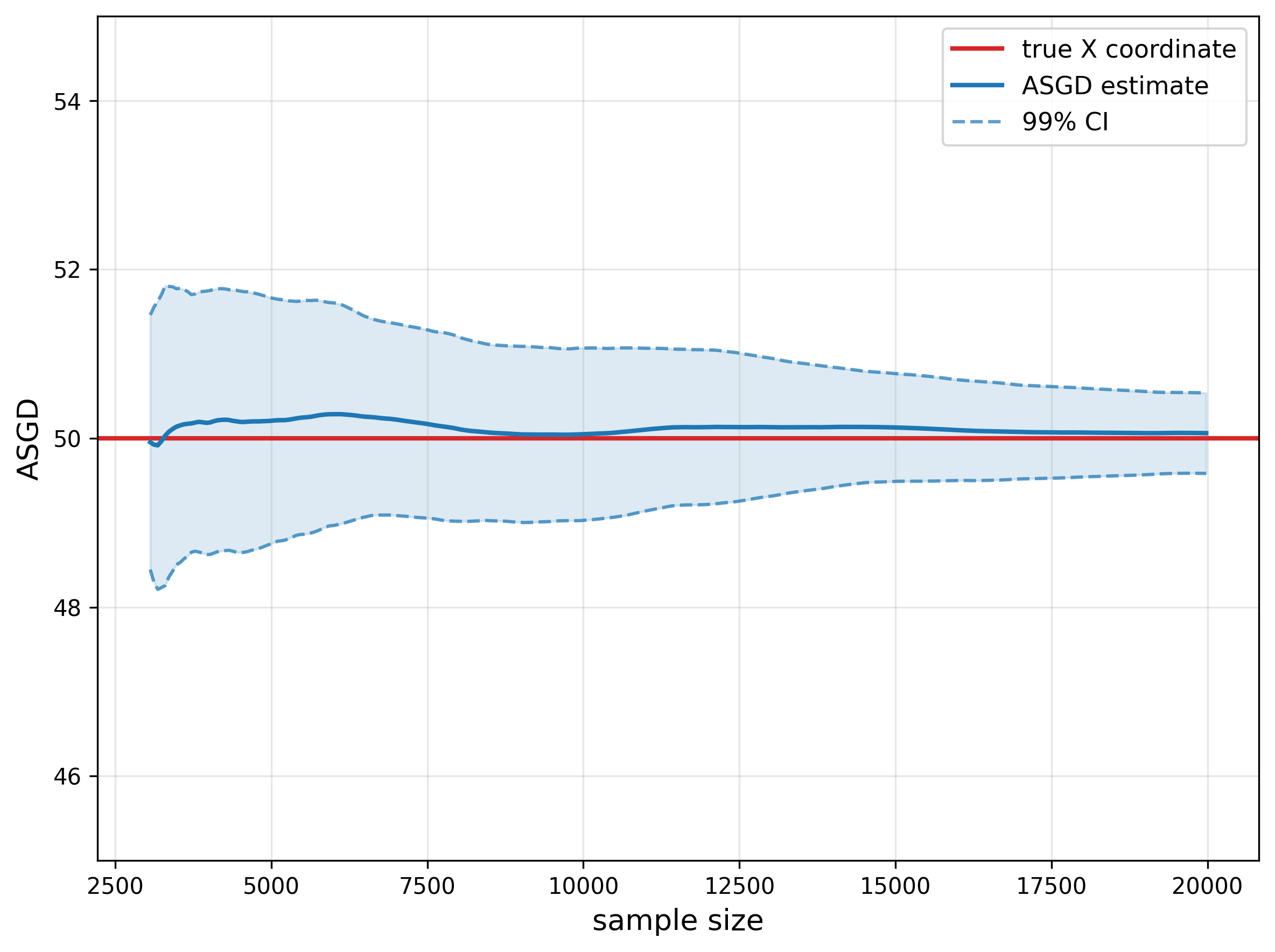}}
\caption{ (a): Trajectory of the 3D online source localization process; (b): ASGD estimate for the $X$-axis coordinate with its $99\%$ confidence interval.  }
\label{fig:trajectory} 
\end{figure}

\begin{figure}
\centering 
\subfigure[]{\includegraphics[width=0.32\textwidth]{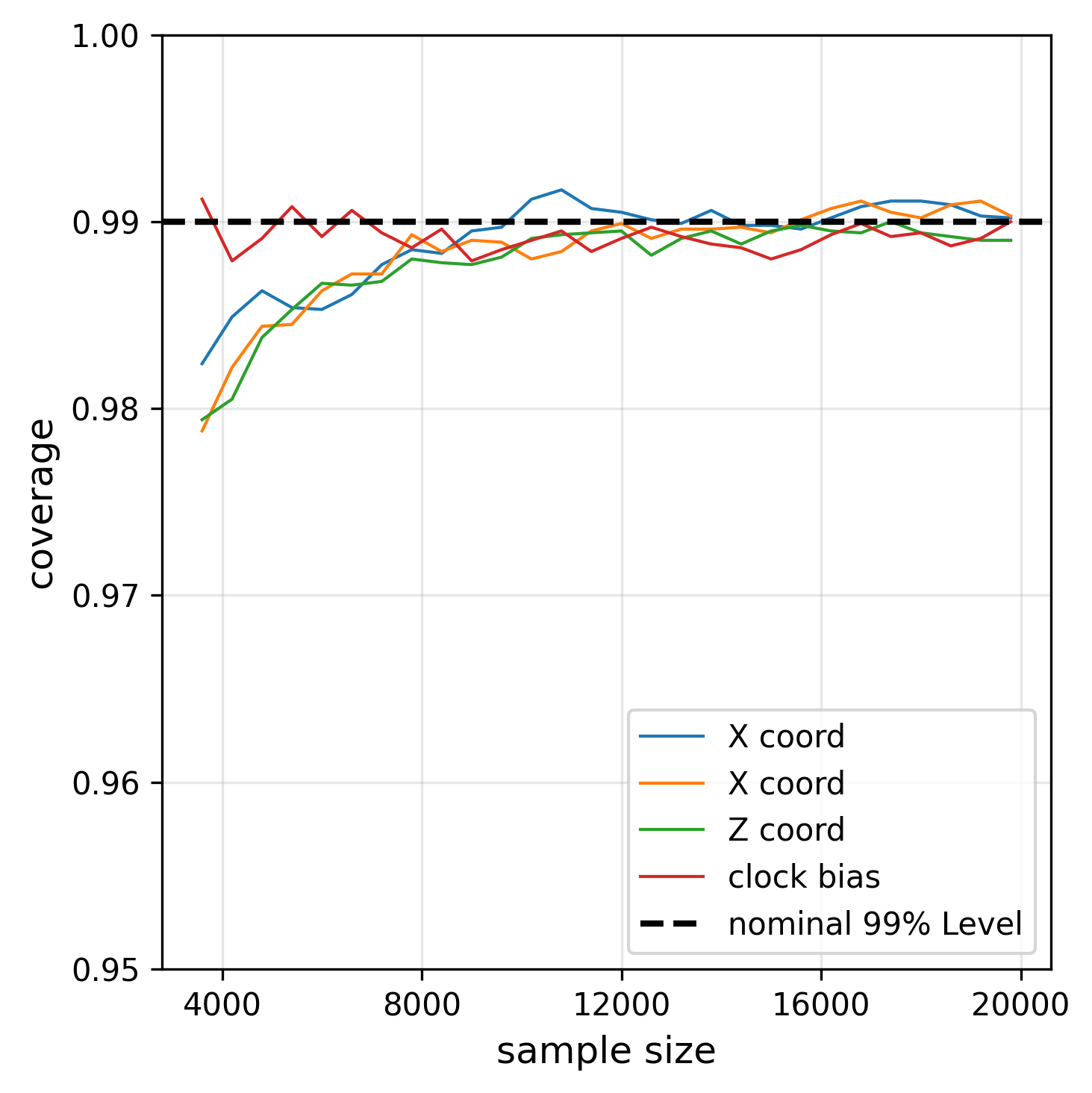}}
\subfigure[]{\includegraphics[width=0.32\textwidth]{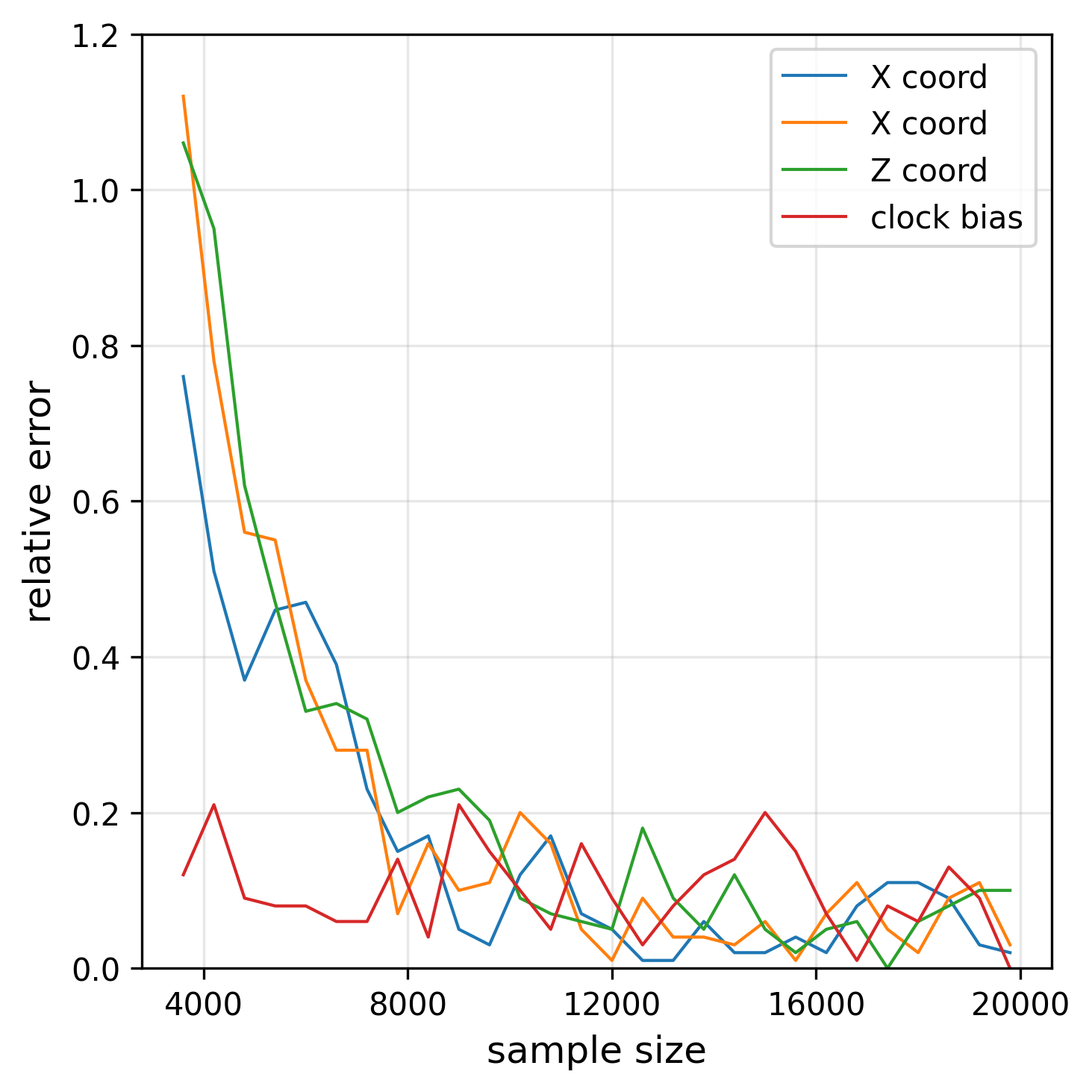}}
\subfigure[]{\includegraphics[width=0.32\textwidth]{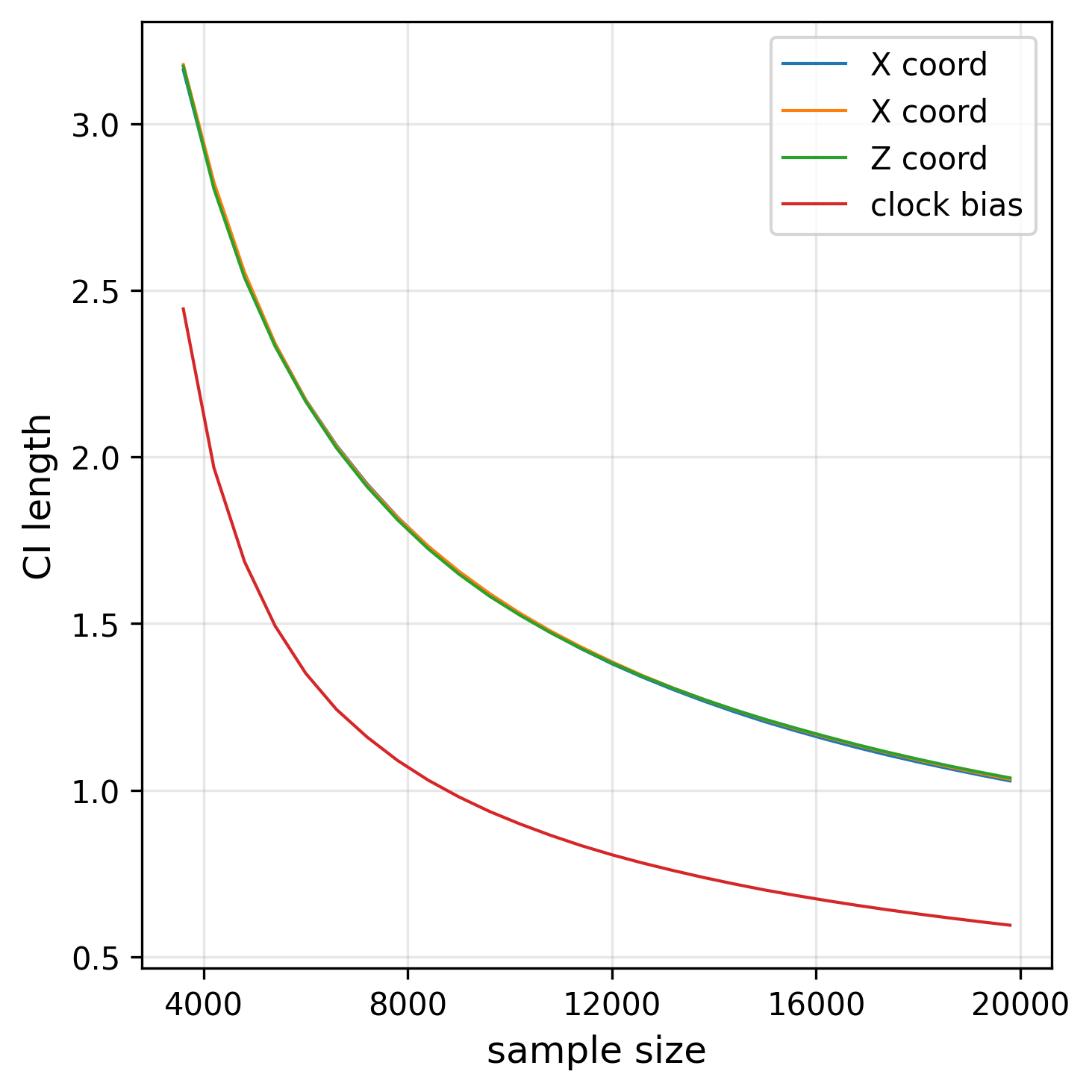}}
\caption{Online source localization ($99\%$ nominal level). (a): Empirical coverage;  (b): Relative error;  (c): The length of the confidence interval.}
\label{fig:TOA results} 
\end{figure}
}

\section{Discussion}
This paper presents a novel inference framework to construct confidence intervals for model parameters by employing stochastic algorithms in an online environment. The primary advantages of this method are its conceptual simplicity and ease of implementation, offering flexibility across various algorithms. The constructions of confidence intervals are computationally efficient and almost free (post-SGD update). Furthermore, we bolster our approach with rigorous theoretical guarantees,  demonstrating its capability to provide valid inference at a high confidence level.

\revise{
\section*{Acknowledgments}
The authors sincerely thank the editor, associate editor, and reviewers for their insightful and constructive comments that significantly improved the manuscript.
}

 \bibliography{reference} 

\appendix
\section{Proof}
\label{Section_Proof}
We first introduce extra notations for the proof. In the rest of the Appendix, for any vector $\nu = (\nu_{1}, \ldots, \nu_{m})^{\top} \in \mathbb{R}^{m}$, we use $|\nu| = \sqrt{\sum_{\ell = 1}^{m} \nu_{\ell}^{2}}$ to denote its Euclidean norm.
For any random vector $X \in \mathbb{R}^{m}$ and constant $q > 0$, we write $\|X\|_{q} = (\mathbb{E} |X|^{q})^{1/q}$ if $\mathbb{E} |X|^{q} < \infty$.

\subsection{Proof of Theorem~\ref{Theorem_strong_approximation}}
\begin{proof}
	Without loss of generality, we assume $x^{*} = 0$. Observe that
	\begin{align*}
		x_{i} &= x_{i - 1} - \eta_{i} \nabla F(x_{i - 1}) + \eta_{i} \{\nabla F(x_{i - 1}) - \nabla f(x_{i - 1}, \xi_{i})\} \cr 
		&= x_{i - 1} - \eta_{i} \nabla F(x_{i - 1}) + \eta_{i} \Delta (x_{i - 1}, \xi_{i}), 
	\end{align*}
	where $\{\Delta (x_{i - 1}, \xi_{i})\}_{i \in \mathbb{N}}$ is a sequence of martingale differences with respect to the filtration $\mathcal{F}_{i} = \sigma(\xi_{1}, \ldots, \xi_{i})$, $i \geq 1$. Then, by Assumptions~\ref{Assumption_convex_F} and~\ref{Assumption_Delta_Lip}, we have 
    \begin{align*}
        |x_{i - 1} - \eta_{i} \nabla F(x_{i - 1})|^{2} \leq (1 - 2 \tau \eta_{i} + \eta_{i}^{2} L^{2}) |x_{i - 1}|^{2}
    \end{align*}
    and 
	\begin{align*}
		\|x_{i}\|_{q}^{2} &\leq \|x_{i - 1} - \eta_{i} \nabla F (x_{i - 1})\|_{q}^{2} + (q - 1) \eta_{i}^{2} \|\Delta (x_{i - 1}, \xi_{i})\|_{q}^{2}\cr
		&\leq (1 - 2 \tau \eta_{i} + \eta_{i}^{2} L^{2}) \|x_{i - 1}\|_{q}^{2} + 2 (q - 1) \eta_{i}^{2} (\|\Delta (0, \xi_{i})\|_{q}^{2} + \gamma^{2} \|x_{i - 1}\|_{q}^{2})\cr
		&\leq (1 - 2 \tau \eta_{i} + \eta_{i}^{2} L^{2} + 2 (q - 1) \gamma^{2} \eta_{i}^{2}) \|x_{i - 1}\|_{q}^{2} + 2 (q - 1) \eta_{i}^{2} \|\Delta(0, \xi_{i})\|_{q}^{2}.  
	\end{align*}
    Consequently, by an argument analogous to the proof of Theorem $1$ in~\cite{Moulines_non-asymptotic_2011}, we obtain
    \begin{align}
    \label{eq_x_moment_bound}
        \|x_{i}\|_{q}^{2} \lesssim \eta_{i} + \exp(- c i^{1 - \beta}) |x_{0}|^{2}.
    \end{align}
	Let $y_{0} = 0$ and 
	\begin{align*}
		y_{i} = y_{i - 1} - \eta_{i} \nabla F(y_{i - 1}) + \eta_{i} \Delta(0, \xi_{i}), \quad i = 1, 2, \ldots.  
	\end{align*}
	Then, by Assumptions~\ref{Assumption_convex_F} and~\ref{Assumption_Delta_Lip}, we have      
	\begin{align*}
		\|x_{i} - y_{i}\|_{q}^{2} &\leq (1 - 2 \tau \eta_{i} + \eta_{i}^{2} L^{2}) \|x_{i - 1} - y_{i - 1}\|_{q}^{2} + (q - 1) \eta_{i}^{2} \|\Delta(x_{i - 1}, \xi_{i}) - \Delta(0, \xi_{i})\|_{q}^{2}\cr
		&\leq (1 - 2 \tau \eta_{i} + \eta_{i}^{2} L^{2}) \|x_{i - 1} - y_{i - 1}\|_{q}^{2} + (q - 1) \eta_{i}^{2} \gamma^{2} \|x_{i - 1}\|_{q}^{2}. 
	\end{align*}
	Similar to~\eqref{eq_x_moment_bound}, it is straightforward to derive that
    \begin{align*}
        \|x_{i} - y_{i}\|_{q}^{2} \lesssim \eta_{i}^{2} + \exp(- c i^{1 - \beta}) |x_{0}|^{2}.
    \end{align*}
    Consequently, by the triangle inequality, it follows that
	\begin{align}
		\label{eq_approximation_x_y}
		\|\bar{x}_{n} - \bar{y}_{n}\|_{q} \leq \frac{1}{n} \sum_{i = 1}^{n} \|x_{i} - y_{i}\|_{q} \lesssim  \max\left(\frac{|x_{0} - x^{*}|}{n}, \frac{1}{n^{\beta}}\right),
	\end{align}
	where $\bar{y}_{n} = n^{-1} \sum_{i = 1}^{n} y_{i}$.
    Let $w_{0} = 0$ and 
	\begin{align*}
		w_{i} &= w_{i - 1} - \eta_{i} \nabla^{2} F(0) w_{i - 1} + \eta_{i} \Delta(0, \xi_{i}) \cr 
		&=: A_{i} w_{i - 1} + \eta_{i} \Delta(0, \xi_{i}), \quad i = 1, 2, \ldots,
	\end{align*}
	where $A_{i} = I_{d} - \eta_{i} \nabla^{2} F(0)$. By Assumption~\ref{Assumption_Hessian_Lip}, we have
	\begin{align*}
		\|y_{i} - w_{i}\|_{q/2} &= \|A_{i}(y_{i - 1} - w_{i - 1}) + \eta_{i}\{\nabla^{2} F(0) y_{i - 1} - \nabla F(y_{i - 1})\}\|_{q/2} \cr 
		&\leq |A_{i}|_{\mathrm{op}} \|y_{i - 1} - w_{i - 1}\|_{q/2} + \eta_{i} \|\nabla^{2} F(0) y_{i - 1} - \nabla F(y_{i - 1})\|_{q/2} \cr 
		&\leq |A_{i}|_{\mathrm{op}} \|y_{i - 1} - w_{i - 1}\|_{q/2} + \mathcal{L} \eta_{i} \|y_{i - 1}\|_{q}^{2}, 
	\end{align*}
    where $|A_{i}|_{\mathrm{op}}$ stands for the operator norm. 
    Let $\lambda_{1} \geq \cdots \geq \lambda_{d} > 0$ denote the eigenvalue of the Hessian matrix $\nabla^{2} F(0) \in \mathbb{R}^{d \times d}$ and let $\mathcal{I} = \inf\{i \geq 1: 1 - \eta_{i} \lambda_{1} \geq 0\}$. Then, $|A_{i}|_{\mathrm{op}} \leq \eta_{i} \lambda_{1} - 1$ for $i < \mathcal{I}$, and $|A_{i}|_{\mathrm{op}} \leq 1 - \eta_{i} \lambda_{d}$ for $i \geq \mathcal{I}$.  Let $\bar{w}_{n} = n^{-1} \sum_{i = 1}^{n} w_{i}$. Similar to~\eqref{eq_approximation_x_y}, it is straightforward to derive that
	\begin{align*}
		\|\bar{y}_{n} - \bar{w}_{n}\|_{q/2} \leq \frac{1}{n} \sum_{i = 1}^{n} \|y_{i} - \omega_{i}\|_{q/2} \lesssim \max\left(\frac{|x_{0} - x^{*}|}{n}, \frac{1}{n^{\beta}}\right),
	\end{align*}
which implies
       \begin{align*}
        \|\sqrt{n} (\bar{x}_{n} - x^{*}) - M_{n}\|_{q/2}^{2} \lesssim \max\left(n^{1 - 2 \beta}, \frac{|x_{0} - x^{*}|^{2}}{n}\right).
    \end{align*}
Thus (\ref{eq:lineaapprox}) holds since $q / 2 \ge 2$.
	Now it remains to derive the strong Gaussian approximation for $\bar{w}_{n}$. Notice that 
	\begin{align*}
		\sum_{i = 1}^{n} w_{i} &= \sum_{i = 1}^{n} \sum_{k = 1}^{i} \prod_{l = k + 1}^{i} A_{l} \eta_{k} \Delta(0, \xi_{k}) \cr
		&= \sum_{k = 1}^{n} \sum_{i = k}^{n} \prod_{l = k + 1}^{i} A_{l} \eta_{k} \Delta(0, \xi_{k}) \cr
		&=: \sum_{k = 1}^{n} D_{k}, 
	\end{align*} 
	where $D_{1}, \ldots, D_{n}$ are independent random vectors. 
    For each $k \geq \mathcal{I}$, it follows from the triangle inequality that
	\begin{align*}
		\|D_{k}\|_{q} &\leq \eta_{k} \sum_{i = k}^{n} \prod_{l = k + 1}^{i} (1 - \eta_{l} \lambda_{d}) \|\Delta(0, \xi_{k})\|_{q}\cr
		&\leq \eta_{k} \sum_{i = k}^{n} \exp\left(-\lambda_{d} \sum_{l = k + 1}^{i} \eta_{l}\right) \|\Delta(0, \xi)\|_{q} \cr
        &\leq \eta_{k} \sum_{i = k}^{n} \exp\left(- c \lambda_{d} i^{1 - \beta} + c \lambda_{d} k^{1 - \beta}\right) \|\Delta(0, \xi)\|_{q} \cr
        &\lesssim \eta_{k} k^{-\beta} \|\Delta(0, \xi)\|_{q}.
	\end{align*}
    For each $k < \mathcal{I}$, we have $\|D_{k}\|_{q} \leq C_{q}$ and hence $\sum_{k = 1}^{n} \|D_{k}\|_{q}^{2} \lesssim n$.
    Consequently, by Theorem 2.1 in~\citet{mies2023sequential}, on a sufficiently rich probability space, there exist independent random vectors $W_{n} \overset{\mathcal{D}}{=} \sqrt{n} \bar{w}_{n}$ and $Z_{n} \sim N(0, \Gamma_{n})$ such that 
	\begin{align*}
        \mathbb{E} |W_{n} - Z_{n}|^{2} \lesssim \frac{\log n}{n^{1 - 2/q}}.
	\end{align*}
	Putting all these pieces together, we obtain~\eqref{eq_strong_approximation_W_Z}.
\end{proof}

\subsection{Proof of Theorem~\ref{thm:sup}}
\begin{proof}
	Let $n = N/K$.	Under Assumption~\ref{ass:normal_convergence_rate}, we have  $Z_{n, k}, k = 1, \ldots, K,$ which are \emph{i.i.d} Gaussian $\mathcal{N}(0, \Sigma_{n})$ such that 
	\[(\E|\sqrt{n}(\hat{x}_{n}^{(k)} - x^{*}) - Z_{n, k}|^2)^{1/2} = O(\delta(n)).
	\]
	For notation simplicity we use $Z_{k}$ to denote $Z_{n, k}$ in the rest of the proof.
	Define 
	\[S = \frac{1}{\sqrt{K}}\sum_{k=1}^{K}\upsilon^{\top}Z_{k},\quad \textnormal{and}\quad R= \sqrt{\frac{1}{K-1} \sum_{k = 1}^{K} (\upsilon^{\top}(Z_{k}  - K^{-1}\sum_{k=1}^{K}Z_k))^{2}}.\]
	It can be shown that 
	\[\frac{S}{R} \sim t_{K-1}.\]
	Further define
	\[\hat{S}_{n} = \frac{1}{\sqrt{K}}\sum_{k=1}^{K}\upsilon^{\top}\sqrt{n}(\hat{x}_{n}^{(k)}-x^{*}) ,\quad \textnormal{and}\quad \hat{R}_{n} = \sqrt{\frac{1}{K-1} \sum_{k = 1}^{K} (\upsilon^{\top}\sqrt{n}(\hat{x}_{n}^{(k)} - \bar{x}_{K, n}))^{2}}.\]
	Then  $\hat{t}_{\upsilon}$  can be rewritten as  $\hat{t}_{\upsilon} = \hat{S}_{n}/\hat{R}_{n}$. Now  it is suffice to show 
	\[\sup_{t}\left| \P\left(\frac{\hat{S}_{n}}{\hat{R}_{n}} \ge t\right) - \P\left(\frac{S}{R}\ge t\right)\right|  \lesssim c_{1}(n)^{1/4}.\] 
	
	\bigskip
	\noindent{\bf Step 1: Bound $\E(\hat{S}_{n}  - S)^2$ and $\E(\hat{R}_{n}  - R)^2$.} We first show that $\hat{S}_{n}  - S$ and $\hat{R}_{n}  - R$ have the same convergence rate in Assumption~\ref{ass:normal_convergence_rate}.
	
	Using Cauchy–Schwarz inequality and Assumption~\ref{ass:normal_convergence_rate}, we have 
	 \[\E(\hat{S}_{n}  - S)^2 \le \sum_{k=1}^{K}|\upsilon|^2\E|\sqrt{n}(\hat{x}_{n}^{(k)} - x^{*}) - Z_{k})|^2 = K |\upsilon|^2 \delta^{2},\]
	where $\delta = O(\delta(n))$, which converges to $0$.  
	Similarly, applying triangle inequality  
	\begin{equation*}
		\begin{split}
			(\hat{R}_{n}  - R)^2&\le \frac{1}{K-1}\sum_{k=1}^{K}\left[\upsilon^{\top}(\sqrt{n}(\hat{x}_{n}^{(k)} - x^{*}) - Z_{k}) - \frac{1}{\sqrt{K}}(\hat{S}_{n} - S)\right]^2\\
			&\le \frac{2}{K-1}\sum_{k=1}^{K}\left[ | \upsilon|^2 |\sqrt{n}(\hat{x}_{n}^{(k)} - x^{*}) - Z_{k}|^2 + \frac{1}{K}(\hat{S}_{n} - S)^2\right].
		\end{split}
	\end{equation*} Note that $K$ is a fixed number and $\upsilon$  a fixed projection vector, we obtain:
    \[\mathbb{E}(\hat{R}_{n} - R)^2 \le C(K, \upsilon) \cdot  \delta^2,\]
where $C(K, \upsilon)$ is a constant independent of $n$.
	So  for $\hat{R}_{n}$ we   have
	\[\mathbb{E}(\hat{R}_{n} - R)^2 \lesssim  \delta^{2}.\]
	
	\noindent{\bf Step 2: Bound $ \P(| \hat{S}_{n}/\hat{R}_{n} -  S/{R}|\ge \epsilon)$.} Our next step is to bound the tail of the difference between $\hat{S}_{n}/\hat{R}_{n}$ and $S/R$. $\P(| \hat{S}_{n}/{\hat{R}_{n}} -  S/{R}|\ge \epsilon)$ can be decompose as
	\[	\P\left(\left|\frac{\hat{S}_{n}}{\hat{R}_{n}} - \frac{S}{R}\right|\ge \epsilon\right) \le \P\left(\left|\frac{\hat{S}_{n}}{\hat{R}_{n}} - \frac{\hat{S}_{n}}{R}\right|\ge \epsilon\right) + \P\left(\left|\frac{\hat{S}_{n}}{R} - \frac{S}{R}\right|\ge \epsilon\right)\]
	
	To deal with the first term in the above inequality, we first look at $\left|\frac{1}{\hat{R}_{n}} - \frac{1}{R}\right|$. For any $a > 0, y> 0, z>0$,
	\begin{equation*}
		\begin{split}
			\P\left( \left|\frac{1}{\hat{R}_{n}} - \frac{1}{R}\right|\ge a \right) \le &\P\left( \left|\frac{1}{\hat{R}_{n}} - \frac{1}{R}\right|\ge a, |R|\ge y,  \left|\hat{R}_{n} - R\right|\le z\right)\\
			& +  \P(|R| < y) + \P( |\hat{R}_{n} - R| >  z)\\
			\le &\P(|\hat{R}_{n} - R|\ge ay(y-z) ) +   \P(|R| < y) + \P( |\hat{R}_{n} - R| >  z)\\
			\lesssim& \frac{\delta^2}{(ay(y-z))^2}+ y + \frac{\delta^2}{z^{2}}
		\end{split}
	\end{equation*}
	The last line is derived from Markov's inequality and the probability density function (pdf) of the chi-square distribution. By choosing $y = (\delta/a)^{2/5}, z = y/2$, when $a\le \delta^{-2/3}$ we have
	\[\frac{\delta^2}{(ay(y-z))^2}+ y + \frac{\delta^2}{z^{2}}  = \frac{4\delta^2}{a^2y^4}  +y + \frac{4\delta^2}{y^2}  \lesssim \left(\frac{\delta}{a}\right)^{2/5}.\]
	Then we have 
	\begin{equation*}
		\begin{split}
			\P\left(\left|\frac{\hat{S}_{n}}{\hat{R}_{n}} - \frac{\hat{S}_{n}}{R}\right|\ge \epsilon\right) \le &\P\left(|\hat{S}_{n}|\left|\frac{1}{\hat{R}_{n}} - \frac{1}{R}\right|\ge\epsilon,  \left|\frac{1}{\hat{R}_{n}} - \frac{1}{R}\right|\ge a \right) \\  &+ \P\left(|\hat{S}_{n}|\left|\frac{1}{\hat{R}_{n}} - \frac{1}{R}\right|\ge\epsilon,  \left|\frac{1}{\hat{R}_{n}} - \frac{1}{R}\right|\le a \right)\\
			\le & \P\left( \left|\frac{1}{\hat{R}_{n}} - \frac{1}{R}\right|\ge a \right)  +  \P\left(|\hat{S}_{n}|\ge \frac{\epsilon}{a} \right)\\
			\lesssim& \left(\frac{\delta}{a}\right)^{2/5}  + \left(\frac{a}{\epsilon}\right)^{2} 
		\end{split}
	\end{equation*}
	Similarly, for any $b > 0$,
	\begin{equation*}
		\begin{split}
			\P\left(\left|\frac{\hat{S}_{n}}{R} - \frac{S}{R}\right|\ge \epsilon\right)\le 	&\P\left(\frac{|\hat{S}_{n} - S|}{R} \ge \epsilon, R\ge b\right) + \P\left(\frac{|\hat{S}_{n} - S|}{R} \ge \epsilon, R\le b\right)\\
			\le & \P(|\hat{S}_{n} - S| \ge b\epsilon)+ \P(R\le b)\\
			\lesssim & \frac{\delta^{2}}{b^2\epsilon^2} + b.
		\end{split}
	\end{equation*}
	The last step is derived from Markov's inequality and the probability density function (pdf) of the chi-square distribution. Then combine everything we have 
	\begin{equation*}
		\begin{split}
			\P\left(\left|\frac{\hat{S}_{n}}{\hat{R}_{n}} - \frac{S}{R}\right|\ge \epsilon\right)& \le \P\left(\left|\frac{\hat{S}_{n}}{\hat{R}_{n}} - \frac{\hat{S}_{n}}{R}\right|\ge \epsilon\right) + \P\left(\left|\frac{\hat{S}_{n}}{R} - \frac{S}{R}\right|\ge \epsilon\right)\\
			& \lesssim \left(\frac{\delta}{a}\right)^{2/5}  + \left(\frac{a}{\epsilon}\right)^{2} +  \frac{\delta^{2}}{b^2\epsilon^2} + b. 
		\end{split}
	\end{equation*}

	\noindent{\bf Step 3: Bound  $ | \P ( (\hat{S}_{n}/{\hat{R}_{n}}) \ge t ) - \P ( (S/{R})\ge t ) |$.}
	Let $f_{t_{K-1}}$ denote the pdf of $t_{K-1}$, then for any $t$ and $ 0<\epsilon < t$,
	\[\P(t_{K-1}\ge t - \epsilon) - \P(t_{K-1}\ge t)\le  \epsilon f_{t_{K-1}}(t-\epsilon)\lesssim \epsilon.\]
	Then we can bound $ | \P ( (\hat{S}_{n}/{\hat{R}_{n}}) \ge t ) - \P ( (S/{R})\ge t ) |$ as following 
	\begin{equation*}
		\begin{split}
			\left| \P\left(\frac{\hat{S}_{n}}{\hat{R}_{n}} \ge t \right) - \P\left(\frac{S}{R}\ge t\right) \right|& \le \P(t_{K-1}\ge t - \epsilon) - \P(t_{K-1}\ge t) + \P\left(\left|\frac{\hat{S}_{n}}{\hat{R}_{n}} - \frac{S}{R}\right|\ge \epsilon\right) \\
			&\lesssim  \epsilon + \left(\frac{\delta}{a}\right)^{2/5}  + \left(\frac{a}{\epsilon}\right)^{2} +  \frac{\delta^{2}}{b^2\epsilon^2} + b
		\end{split}
	\end{equation*}
	Choose   $a = (\delta\epsilon^5)^{1/6}$, $b = (\delta/\epsilon)^{2/3}$, $\epsilon \asymp \delta^{1/4}$,   we  obtain
	\[  \epsilon + \left(\frac{\delta}{a}\right)^{2/5}  + \left(\frac{a}{\epsilon}\right)^{2} +  \frac{\delta^{2}}{b^2\epsilon^2} + b\lesssim \left(\frac{\delta}{\epsilon}\right)^{1/3}+  \left(\frac{\delta}{\epsilon}\right)^{2/3}  + \epsilon\lesssim \delta^{1/4}.\]
	We therefore have  
	\[\sup_{t}	\left| \P\left(\frac{\hat{S}_{n}}{\hat{R}_{n}} \ge t \right) - \P\left(\frac{S}{R}\ge t\right) \right|  \lesssim \delta(n)^{1/4}.\]
	
\end{proof}

\revise{
\subsection{Ellipsoid-shaped Confidence Set}\label{sec:app_set}
When the number of parallel runs satisfies $K>d$, our framework can also be used to construct an ellipsoidal set.
Using same notation as in the paper that $\bar x_{1,n},\dots,\bar x_{K,n}\in\mathbb{R}^d$,
  each computed from a subsample of size $n=N/K$.
  Define the sample mean and covariance
  \[ 
  \bar x_{K,n} = \frac{1}{K}\sum_{k=1}^K \bar x_{k,n},
  \qquad
  \hat\Sigma = \frac{1}{K-1}\sum_{k=1}^K (\bar x_{k,n}-\bar x_{K,n}) (\bar x_{k,n}-\bar x_{K,n})^\top .
  \]
  A Hotelling $T^2$ statistic for the target $x^\star$ is
  \begin{equation}\label{eq:t2}
     \hat{T}^2 = K\,(\bar x_{K,n} - x^\star)^\top  \hat\Sigma^{-1}  (\bar x_{K,n} - x^\star). 
  \end{equation}
  Following the classical Hotelling construction, a 
  $(1-\alpha)\times 100\%$ confidence set for $x^\star$
  is
  \[
  \widehat{\mathcal C}_{1-\alpha}
  =
  \left\{
  x\in\mathbb{R}^d:\;
  K\,(\bar x_{K,n} - x)^\top 
  \hat\Sigma^{-1}
  (\bar x_{K,n} - x)
  \le 
  c_{K,d,\alpha}
  \right\},
  \]
  where
  \[
  c_{K,d,\alpha}
  =
  \frac{d(K-1)}{K-d}\,
  F_{d,K-d;\,1-\alpha},
  \]
  and $F_{d,K-d;\,1-\alpha}$ denotes the $(1-\alpha)$ quantile of the
  $F$-distribution with $(d,K-d)$ degrees of freedom.  
 The following Theorem, analogous to Theorem \ref{thm:sup}, establishes the asymptotic convergence of the corresponding $T^2$ statistic. This result implies the asymptotic validity of the above confidence region.

\begin{theorem} 
\label{thm:multi}
Suppose we run the parallel algorithm with $K$ trials, and a multivariate version of the approximation holds: 
$\left(\mathbb{E}|\sqrt{n}(\widehat{x}_{n}^{(k)} - x^{*}) - Z_{k}|^2\right)^{1/2} = \mathcal{O}(\delta(n))$, 
where $Z_{k} \sim \mathcal{N}(0, \Sigma_{n})$ with $c I_d \preceq \Sigma_n \preceq C I_d$. 
Assume the degrees of freedom satisfy $K - d \ge 1$. For $\hat{T}$ defined in \eqref{eq:t2}, we have 
\[\sup_{t} \left|\P\left(\hat{T}^2\ge t\right) - \P\left(T^2\ge t\right)\right|\lesssim \delta(n)^{1/4},\]
where $\frac{K-d}{d(K-1)}T^2\sim F_{d, K-d}$.
\end{theorem}

\begin{proof}
Let $\widehat{Z}_{k} = \sqrt{n}(\widehat{x}_{n}^{(k)}-x^{*})$, and $S$, $R$, $\widehat{S}_n$, and $\widehat{R}_n$ be defined as 
\begin{align*}
	\widehat{S}_n &= \frac{1}{\sqrt{K}}\sum_{k=1}^{K} \widehat{Z}_{k}, \quad \textnormal{and}\quad \widehat{R}_n = \frac{1}{\sqrt{K-1}} \big[ \widehat{Z}_{1} - \frac{1}{\sqrt{K}}\widehat{S}_n, \dots, \widehat{Z}_{K} - \frac{1}{\sqrt{K}}\widehat{S}_n \big], \\
	S &= \frac{1}{\sqrt{K}}\sum_{k=1}^{K} Z_{k},\quad \textnormal{and}\quad R = \frac{1}{\sqrt{K-1}} \big[ Z_{1} - \frac{1}{\sqrt{K}}S, \dots, Z_{K} - \frac{1}{\sqrt{K}}S \big].
\end{align*}
It is sufficient to show  
\[ \sup_{t} \left|\P\left(\widehat{S}_n^\top (\widehat{R}_n \widehat{R}_n^\top)^{-1} \widehat{S}_n \ge t\right) - \P\left( S^\top (R R^\top)^{-1} S \ge t\right)\right| \lesssim \delta(n)^{1/4}. \]
	\noindent{\bf Step 1: Bound $\mathbb{E}|\widehat{S}_n - S|^2$ and $\mathbb{E}|\widehat{R}_n - R|_F^2$.} 
	By the assumption, we have $\mathbb{E}|\widehat{Z}_k - Z_k|^2 = \mathcal{O}(\delta(n)^2)$. Using the Cauchy–Schwarz inequality, 
	\[\mathbb{E}|\widehat{S}_n - S|^2 \le \frac{1}{K} \cdot K\sum_{k=1}^{K}\mathbb{E}|\widehat{Z}_k - Z_{k}|^2 = \delta^2,\]
	where $\delta = \mathcal{O}(\delta(n))$.  
	
	By the definition of the Frobenius norm, $|\widehat{R}_{n} - R|_{F}^2$ is the sum of the squared Euclidean norms of its column vectors. Using the definitions of $\widehat{R}_{n}$ and $R$, we can write the squared Frobenius norm of their difference as:
	\begin{equation*}
	\begin{split}
	|\widehat{R}_{n} - R |_{F}^2 = \frac{1}{K-1} \sum_{k=1}^{K} \left| \left( \widehat{Z}_{k} - Z_{k} \right) - \frac{1}{\sqrt{K}} (\widehat{S}_{n} - S) \right|^2.
	\end{split}
	\end{equation*}

	Applying $|a - b|^2 \le 2|a|^2 + 2|b|^2$, we can bound the summation:
	\begin{equation*}
	\begin{split}
	|\widehat{R}_{n} - R|_{F}^2 &\le \frac{1}{K-1} \sum_{k=1}^{K} \left( 2\left| \widehat{Z}_{k} - Z_{k} \right|^2 + 2\left| \frac{1}{\sqrt{K}} (\widehat{S}_{n} - S) \right|^2 \right) \\
	&= \frac{2}{K-1} \sum_{k=1}^{K} \left[ \left| \widehat{Z}_{k} - Z_{k} \right|^2 + \frac{1}{K} \left| \widehat{S}_{n} - S \right|^2 \right].
	\end{split}
	\end{equation*}

	Taking the expectation on both sides:
	\begin{equation*}
	\mathbb{E}|\widehat{R}_{n} - R|_{F}^2 \le \frac{2}{K-1} \sum_{k=1}^{K} \left[ \mathbb{E}\left| \widehat{Z}_{k} - Z_{k} \right|^2 + \frac{1}{K} \mathbb{E}\left| \widehat{S}_{n} - S \right|^2 \right] \lesssim \delta^2.
	\end{equation*}

	\bigskip
	\noindent{\bf Step 2: Full Rank Properties and Wishart Tail Bounds.}
	First, we establish that $R$ almost surely has full row rank and characterize the tail probability of its minimum singular value $\sigma_{\min}(R)$. 
	Since $R$ is constructed from $K$ independent Gaussian vectors centered by their sample mean, the scaled matrix $(K-1)RR^\top$ follows a Wishart distribution $\mathcal{W}_d(K-1, \Sigma_n)$. The degree of freedom condition $K - d \ge 1$ implies that $(K-1)RR^\top$ is strictly positive definite almost surely. Consequently, $R$ has full row rank $d$, and its right Moore-Penrose pseudo-inverse is uniquely and explicitly given by $R^\dagger = R^\top (RR^\top)^{-1}$.

	By the properties of the Wishart distribution, $(K-1)RR^\top \stackrel{d}{=} \Sigma_n^{1/2} \widetilde{V} \Sigma_n^{1/2}$, where $\widetilde{V} \sim \mathcal{W}_d(K-1, I_d)$ is a standard Wishart matrix. The minimum eigenvalue satisfies $\lambda_{\min}(RR^\top) \ge (K-1)^{-1}\lambda_{\min}(\Sigma_n) \lambda_{\min}(\widetilde{V})$. Assuming $\Sigma_n$ is non-singular, we can bound the tail probability by analyzing $\widetilde{V}$. 
	
	Let $\lambda_1, \dots, \lambda_d$ be the eigenvalues of $\widetilde{V}$. The symmetric joint probability density function is:
	\[ f(\lambda_1, \dots, \lambda_d) = C_{K,d} \prod_{i=1}^d \lambda_i^{\frac{K-d-2}{2}} e^{-\frac{1}{2}\lambda_i} \prod_{1 \le i < j \le d} |\lambda_i - \lambda_j|, \]
	where $C_{K,d}>0$ is a normalizing constant. Let $U = \min(\lambda_1, \dots, \lambda_d)$. By symmetry, the marginal density $f_U (u)$ for $u > 0$ is:
	\[ f_U(u) = d \int_{u}^{\infty} \dots \int_{u}^{\infty} f(u, \lambda_2, \dots, \lambda_d) \, d\lambda_2 \dots d\lambda_d. \]
	Substituting $f$ and isolating $u$ yields:
	\[ f_U(u) = d \cdot C_{K,d} \, u^{\frac{K-d-2}{2}} e^{-\frac{1}{2}u} \int_{u}^{\infty} \dots \int_{u}^{\infty} \prod_{i=2}^d \lambda_i^{\frac{K-d-2}{2}} e^{-\frac{1}{2}\lambda_i} \prod_{i=2}^d (\lambda_i - u) \prod_{2 \le i < j \le d} |\lambda_i - \lambda_j| \, d\lambda_2 \dots d\lambda_d. \]
	As $u \to 0$, the multiple integral term is monotonically bounded above by the same integral evaluated from $0$ to $\infty$, which is strictly finite. Thus, the behavior near the origin is dominated by the pre-factor: $f_U(u) \lesssim u^{\frac{K-d-2}{2}}$. As a result, for any sufficiently small $y > 0$,
    $$\P(U < y^2) = \int_{0}^{y^2} f_U(u) du \lesssim \int_{0}^{y^2} u^{\frac{K-d-2}{2}} du.$$
	Since $K - d \ge 1$, the exponent $(K-d-2)/2 \ge -1/2$, which guarantees that $$\P(U < y^2) \lesssim \int_{0}^{y^2} u^{-1/2} du = 2y.$$
	This implies $\P(\lambda_{\min}(RR^\top) < y^2) \lesssim y$. Because $\sigma_{\min}(R) = \sqrt{\lambda_{\min}(RR^\top)}$, we obtain $\P(\sigma_{\min}(R) < y) = \P(\lambda_{\min}(RR^\top) < y^2) \lesssim y$. 

     Similar to the proof of Theorem~\ref{thm:sup}, where the boundedness of the chi-square density is used to control the tail behavior of the univariate $|R|$, we achieve the analogous result here via the eigenvalue density of Wishart matrices.

	Regarding the empirical counterpart $\widehat{R}_n$, it does not need to almost surely have full rank. In the subsequent analysis, we condition on a high-probability event $\Omega = \{\sigma_{\min}(R) \ge y\} \cap \{|\widehat{R}_n - R|_F \le z\}$ with $y > z > 0$. By Weyl's inequality, on the event $\Omega$, we have $\sigma_{\min}(\widehat{R}_n) \ge \sigma_{\min}(R) - |\widehat{R}_n - R|_F \ge y - z > 0$. Therefore, on $\Omega$, $\widehat{R}_n$ inherently has full row rank, ensuring that its pseudo-inverse $\widehat{R}_n^\dagger = \widehat{R}_n^\top (\widehat{R}_n \widehat{R}_n^\top)^{-1}$ is rigorously well-defined for the matrix perturbation bounds.

	\bigskip
	\noindent{\bf Step 3: Bound $\P( | \widehat{R}_n^\dagger \widehat{S}_n - R^\dagger S | \ge \epsilon )$.}
	Let $|\cdot|$ denote the Euclidean norm for vectors and the spectral norm for matrices. Note that $|\cdot| \le |\cdot|_F$. 
	By the triangle inequality and the sub-multiplicativity of operator norms:
	\[ | \widehat{R}_n^\dagger \widehat{S}_n - R^\dagger S | \le | (\widehat{R}_n^\dagger - R^\dagger) \widehat{S}_n | + | R^\dagger (\widehat{S}_n - S) |. \]
	Therefore, we decompose the tail probability as:
	\[ \P\left( | \widehat{R}_n^\dagger \widehat{S}_n - R^\dagger S | \ge \epsilon \right) \le \P\left( | \widehat{R}_n^\dagger - R^\dagger | | \widehat{S}_n | \ge \frac{\epsilon}{2} \right) + \P\left( | R^\dagger | | \widehat{S}_n - S | \ge \frac{\epsilon}{2} \right). \]

	For the first term, restricting to the event $\Omega$ from Step 2, the bound on the pseudo-inverse difference satisfies 
    $$| \widehat{R}_n^\dagger - R^\dagger | \lesssim \frac{|\widehat{R}_n - R|_F}{\sigma_{\min}(R)\sigma_{\min}(\widehat{R}_n)} \le \frac{|\widehat{R}_n - R|_F}{y(y-z)}.$$
	Thus, we have:
	\begin{equation*}
		\begin{split}
			\P\left( | \widehat{R}_n^\dagger - R^\dagger | \ge a \right) \le & \P\left( |\widehat{R}_n - R|_F \ge a y(y-z) \right) + \P\left(\sigma_{\min}(R) < y\right) + \P\left(|\widehat{R}_n - R|_F > z\right) \\
			\lesssim & \frac{\delta^2}{(a y(y-z))^2} + y + \frac{\delta^2}{z^2}.
		\end{split}
	\end{equation*}
	Choosing $y = (\delta/a)^{2/5}$ and $z = y/2$, we obtain $\P( | \widehat{R}_n^\dagger - R^\dagger | \ge a ) \lesssim (\delta/a)^{2/5}$.
	Then,
	\begin{equation*}
		\begin{split}
			\P\left( | \widehat{R}_n^\dagger - R^\dagger | | \widehat{S}_n | \ge \frac{\epsilon}{2} \right) &\le \P\left( | \widehat{R}_n^\dagger - R^\dagger | \ge a \right) + \P\left( | \widehat{S}_n | \ge \frac{\epsilon}{2a} \right) \\
			&\lesssim \left(\frac{\delta}{a}\right)^{2/5} + \left(\frac{a}{\epsilon}\right)^{2}.
		\end{split}
	\end{equation*}

	For the second term, noting that $|R^\dagger| = 1/\sigma_{\min}(R)$, we introduce a threshold $b > 0$:
	\begin{equation*}
		\begin{split}
			\P\left( \frac{| \widehat{S}_n - S |}{\sigma_{\min}(R)} \ge \frac{\epsilon}{2} \right) &\le \P\left( | \widehat{S}_n - S | \ge \frac{b\epsilon}{2} \right) + \P\left(\sigma_{\min}(R) < b\right) \\
			&\lesssim \frac{\delta^2}{b^2 \epsilon^2} + b.
		\end{split}
	\end{equation*}

	Combining these bounds gives:
	\begin{equation} \label{eq:vector_bound}
		\P\left( | \widehat{R}_n^\dagger \widehat{S}_n - R^\dagger S | \ge \epsilon \right) \lesssim \left(\frac{\delta}{a}\right)^{2/5} + \left(\frac{a}{\epsilon}\right)^{2} + \frac{\delta^2}{b^2 \epsilon^2} + b.
	\end{equation}

	\bigskip
	\noindent{\bf Step 4: Bound $$\sup_{t} \left|\P\left(\widehat{S}_n^\top (\widehat{R}_n \widehat{R}_n^\top)^{-1} \widehat{S}_n \ge t\right) - \P\left( S^\top (R R^\top)^{-1} S \ge t\right)\right|.$$}
Observe that 
$$\widehat{S}_n^\top (\widehat{R}_n \widehat{R}_n^\top)^{-1} \widehat{S}_n = \widehat{S}_n^\top (\widehat{R}_n^\dagger)^\top \widehat{R}_n^\dagger \widehat{S}_n = |\widehat{R}_n^\dagger \widehat{S}_n|^2.$$ 
Similarly, $S^\top (R R^\top)^{-1} S = |R^\dagger S|^2$. Let $\widehat{T} = \widehat{R}_n^\dagger \widehat{S}_n$ and $T = R^\dagger S$. Our goal is to bound 
$$\sup_{t} \left|\P(|\widehat{T}|^2 \ge t) - \P(|T|^2 \ge t)\right|.$$

	For $t < 0$, the absolute difference is exactly 0 since the squared norms are non-negative.
	For $t \ge 0$, let $u = \sqrt{t} \ge 0$. The events $\{|\widehat{T}|^2 \ge t\}$ and $\{|\widehat{T}| \ge u\}$ are equivalent. 
	By the reverse triangle inequality, $\big| |\widehat{T}| - |T| \big| \le |\widehat{T} - T|$. Thus, the event $\{|\widehat{T}| \ge u\}$ is subset to $\{|T| \ge u - \epsilon\} \cup \{|\widehat{T} - T| \ge \epsilon\}$. Using this two-sided event inclusion, we bound the difference:
	\begin{equation*}
		\left| \P(|\widehat{T}| \ge u) - \P(|T| \ge u) \right| \le \max_{s \in \{u, u+\epsilon\}} \P(s - \epsilon < |T| \le s) + \P(|\widehat{T} - T| \ge \epsilon).
	\end{equation*}

	Notice that $|T|^2 = S^\top (R R^\top)^{-1} S$ follows a scaled Hotelling's $T^2$-distribution. Since $K - d \ge 1$, its radial distribution (i.e., the distribution of $|T|$) has a globally bounded probability density function $f_{|T|}(x)$. Therefore, 
	\[ \P(s - \epsilon < |T| \le s) \le \epsilon \cdot \sup_x f_{|T|}(x) \lesssim \epsilon. \]

	Substituting the bound from \eqref{eq:vector_bound} into the difference equation yields:
	\begin{equation*}
		\begin{split}
			\sup_{t} \left|\P(|\widehat{T}|^2 \ge t) - \P(|T|^2 \ge t)\right| &\lesssim \epsilon + \P(|\widehat{T} - T| \ge \epsilon) \\
			&\lesssim \epsilon + \left(\frac{\delta}{a}\right)^{2/5} + \left(\frac{a}{\epsilon}\right)^{2} + \frac{\delta^2}{b^2 \epsilon^2} + b.
		\end{split}
	\end{equation*}

	Finally, we balance the parameters by choosing $a = (\delta\epsilon^5)^{1/6}$, $b = (\delta/\epsilon)^{2/3}$, and $\epsilon \asymp \delta^{1/4}$. Substituting these in, the entire right-hand side is bounded by:
	\[ \mathcal{O}(\epsilon) + \mathcal{O}\left( \left(\frac{\delta}{\epsilon}\right)^{1/3} \right) + \mathcal{O}\left( \left(\frac{\delta}{\epsilon}\right)^{2/3} \right) \lesssim \delta(n)^{1/4}. \]
	This completes the proof. 
\end{proof}
}

\section{Additional numerical results}
\begin{figure}
	\centering  
	\subfigure[$\alpha = 0.05$]{
		\includegraphics[width=0.33\textwidth]{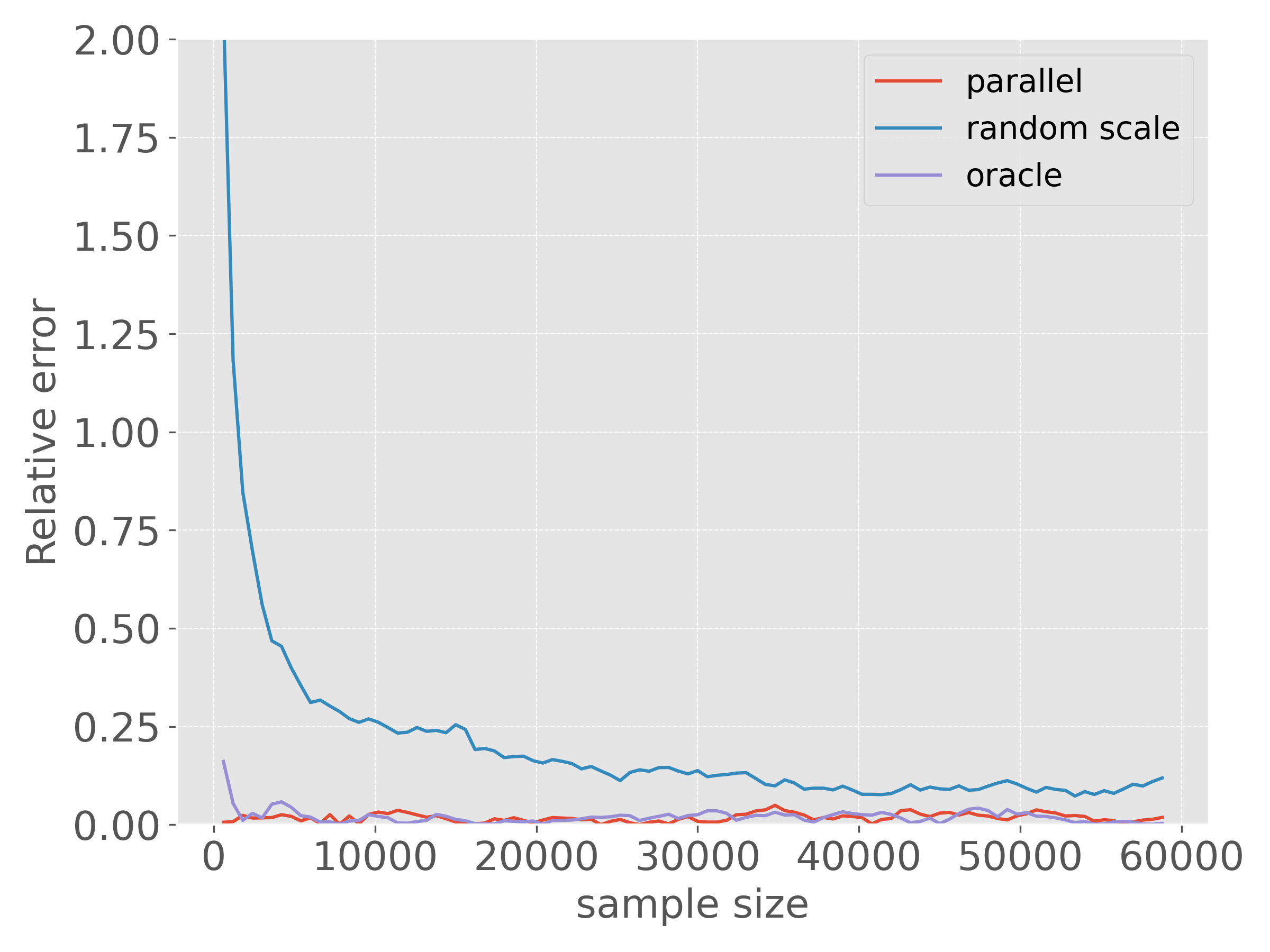} 
		\includegraphics[width=0.33\textwidth]{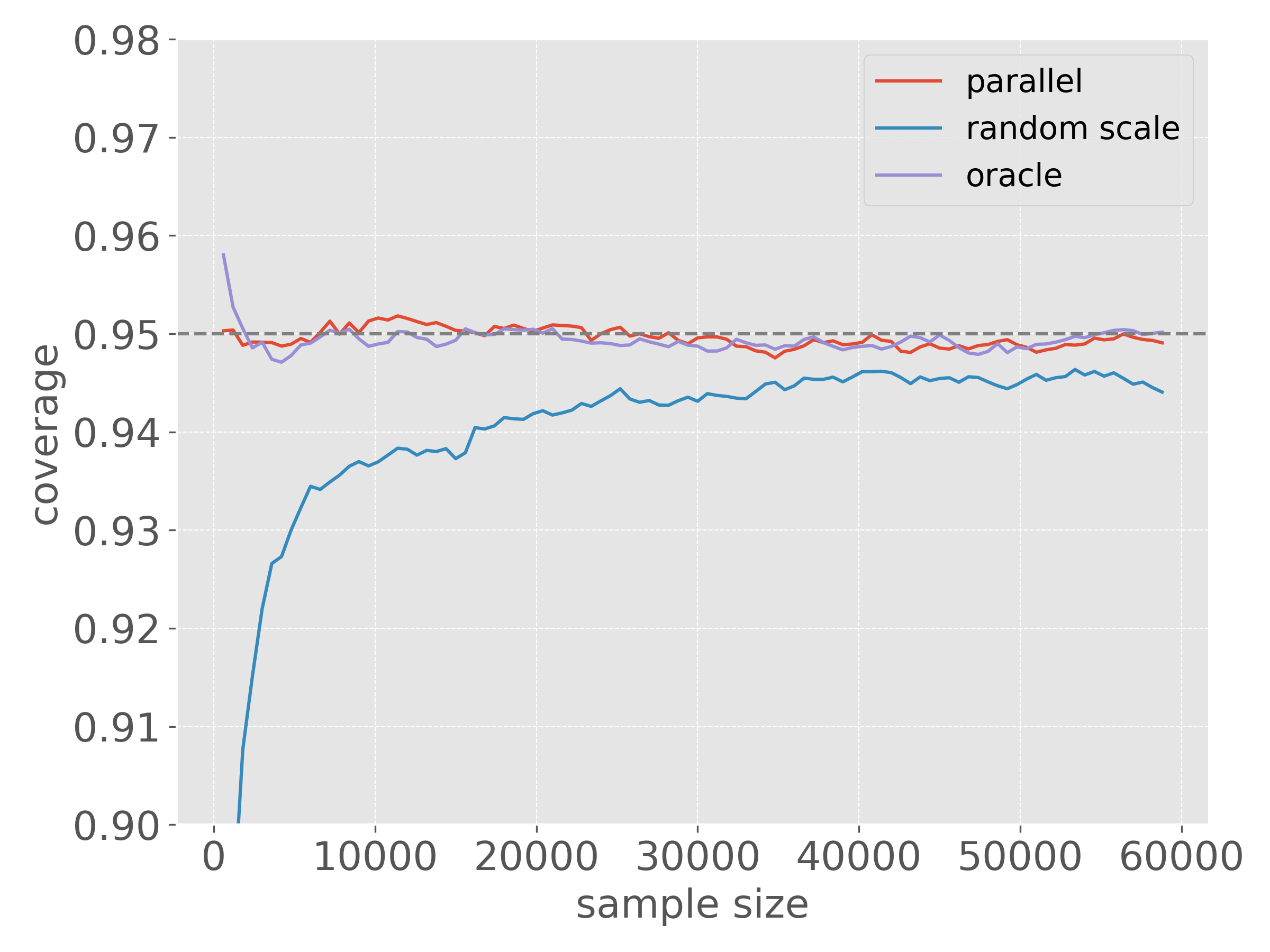} 
		\includegraphics[width=0.33\textwidth]{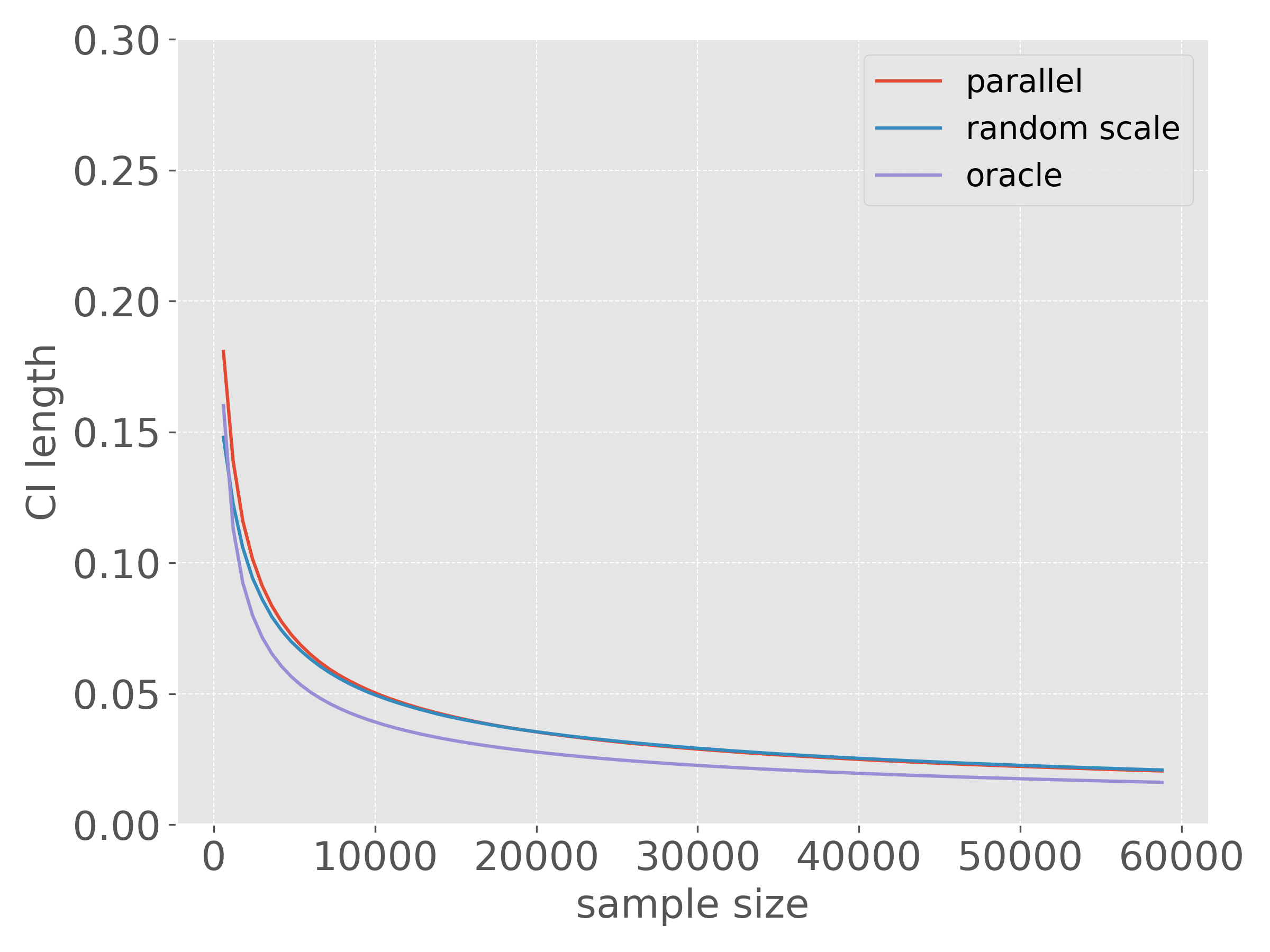}
	}
	\subfigure[$\alpha = 0.01$]{
		\includegraphics[width=0.33\textwidth]{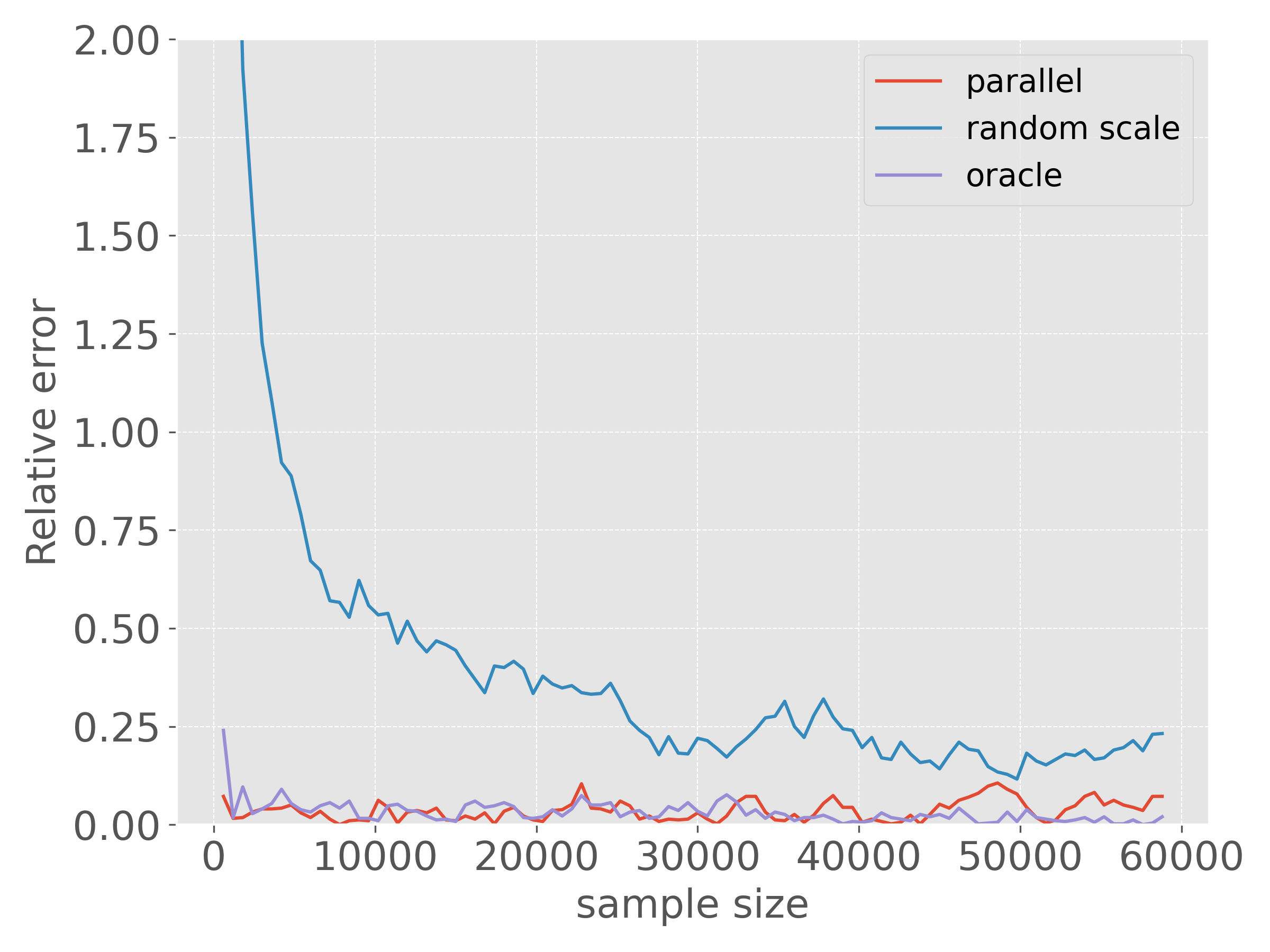} 
		\includegraphics[width=0.33\textwidth]{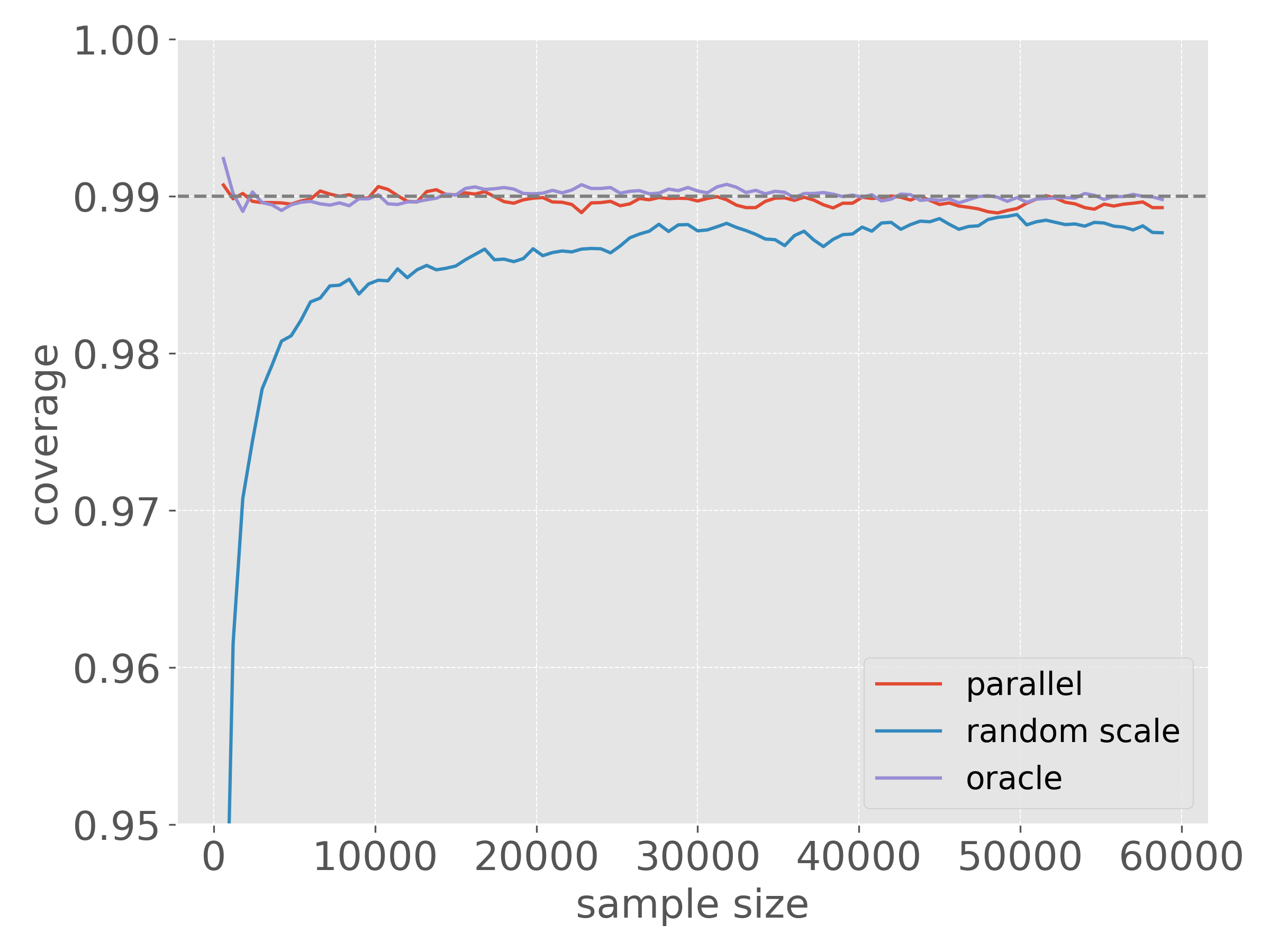} 
		\includegraphics[width=0.33\textwidth]{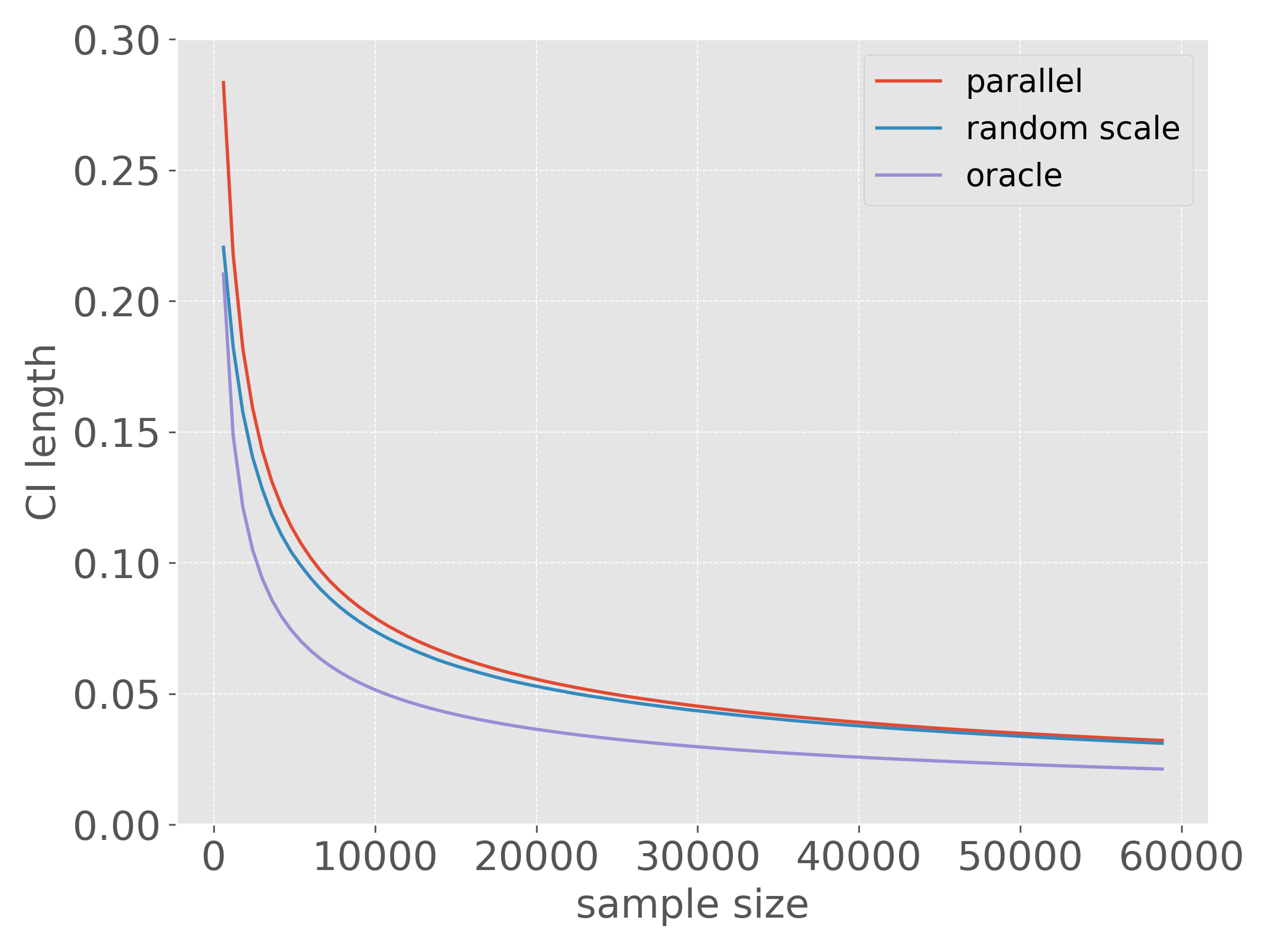}
	}
	\subfigure[$\alpha = 0.001$]{
		\includegraphics[width=0.33\textwidth]{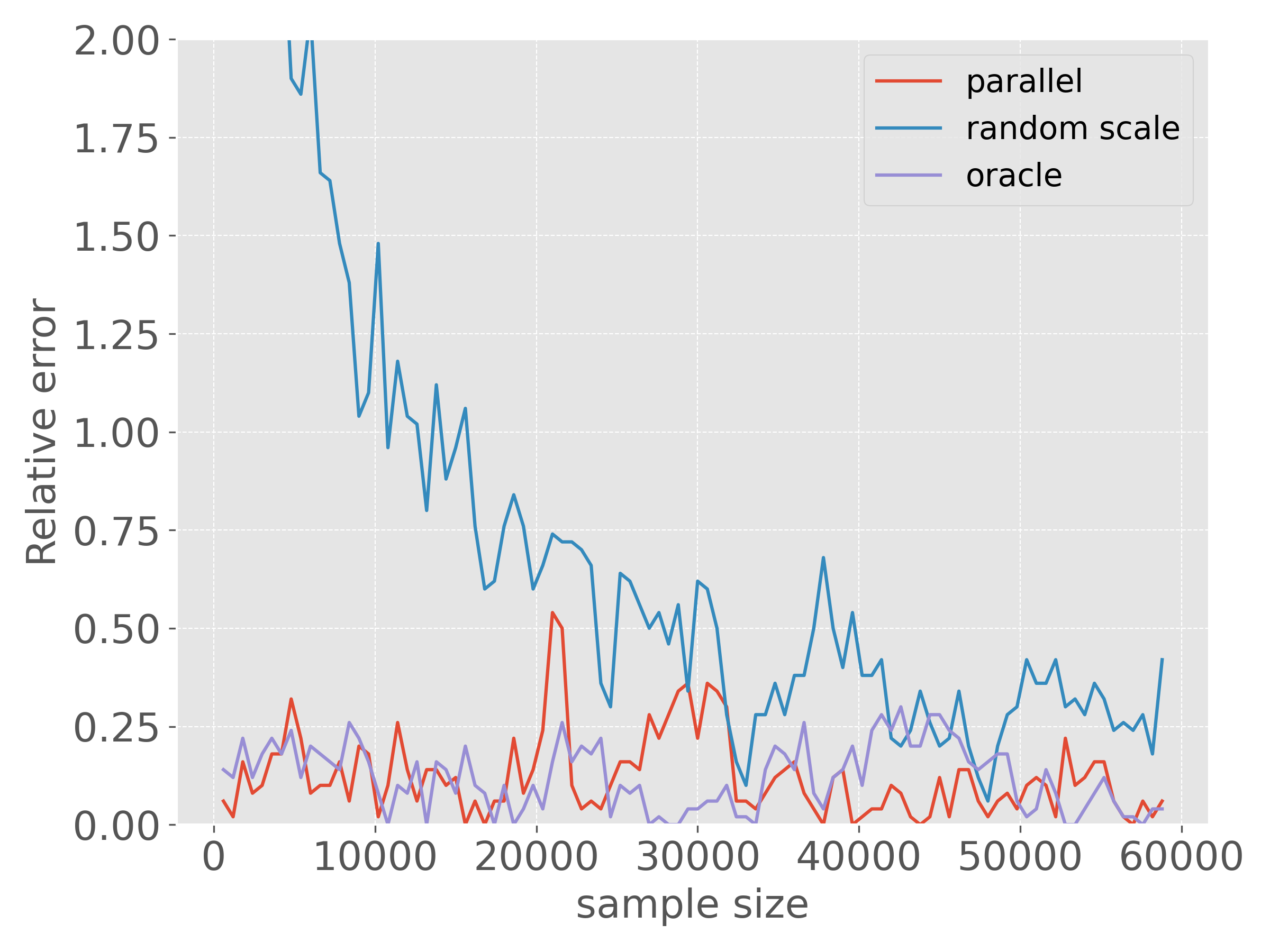} 
		\includegraphics[width=0.33\textwidth]{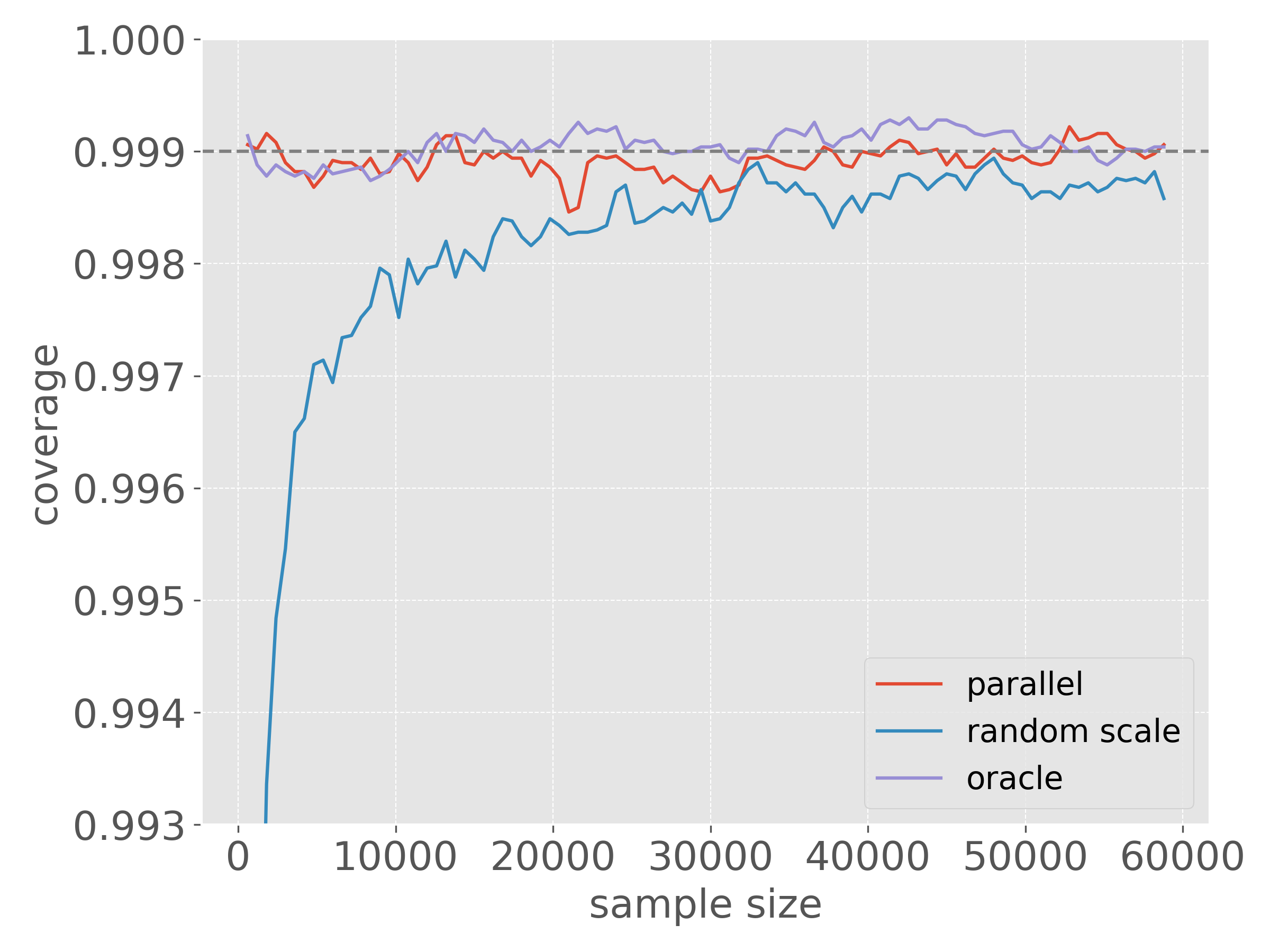} 
		\includegraphics[width=0.33\textwidth]{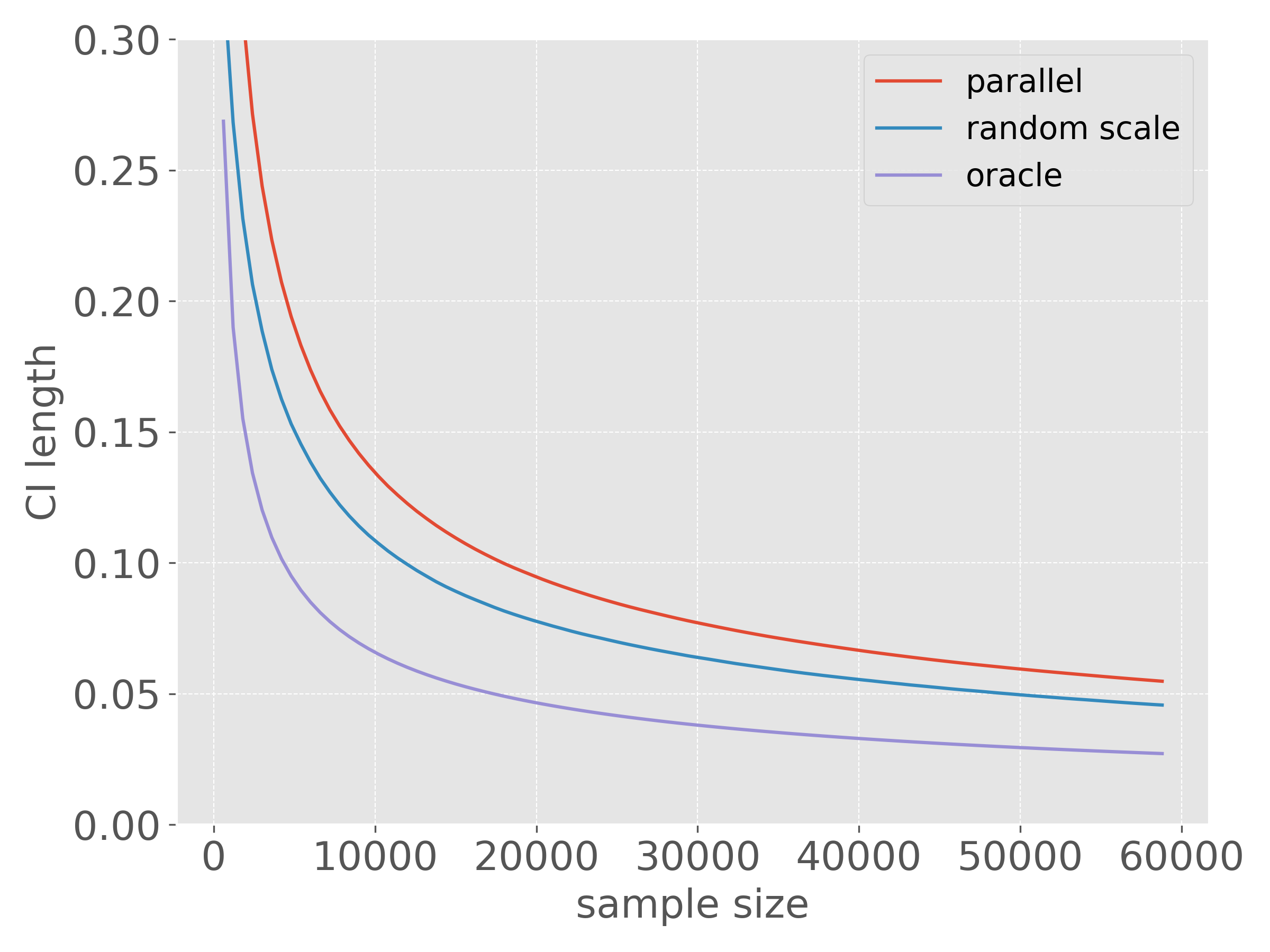}
	}
	\caption{Linear Regression $d = 5$: Left: relative error of coverage; Middle: empirical coverage; Right: length of confidence intervals.}
	\label{fig:linear_d5}
\end{figure}

\begin{figure}
	\centering  
	\subfigure[$\alpha = 0.05$]{
		\includegraphics[width=0.33\textwidth]{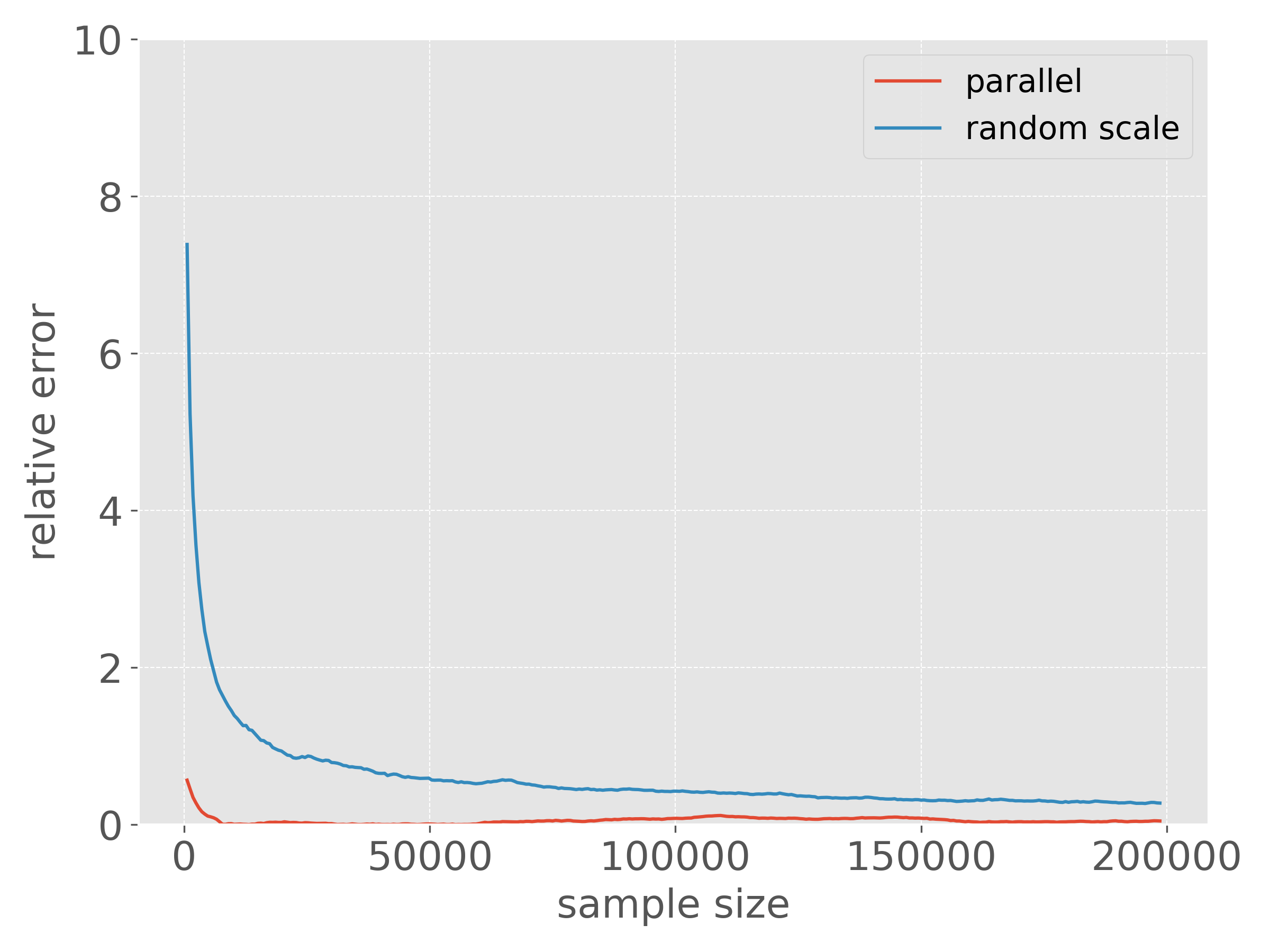} 
		\includegraphics[width=0.33\textwidth]{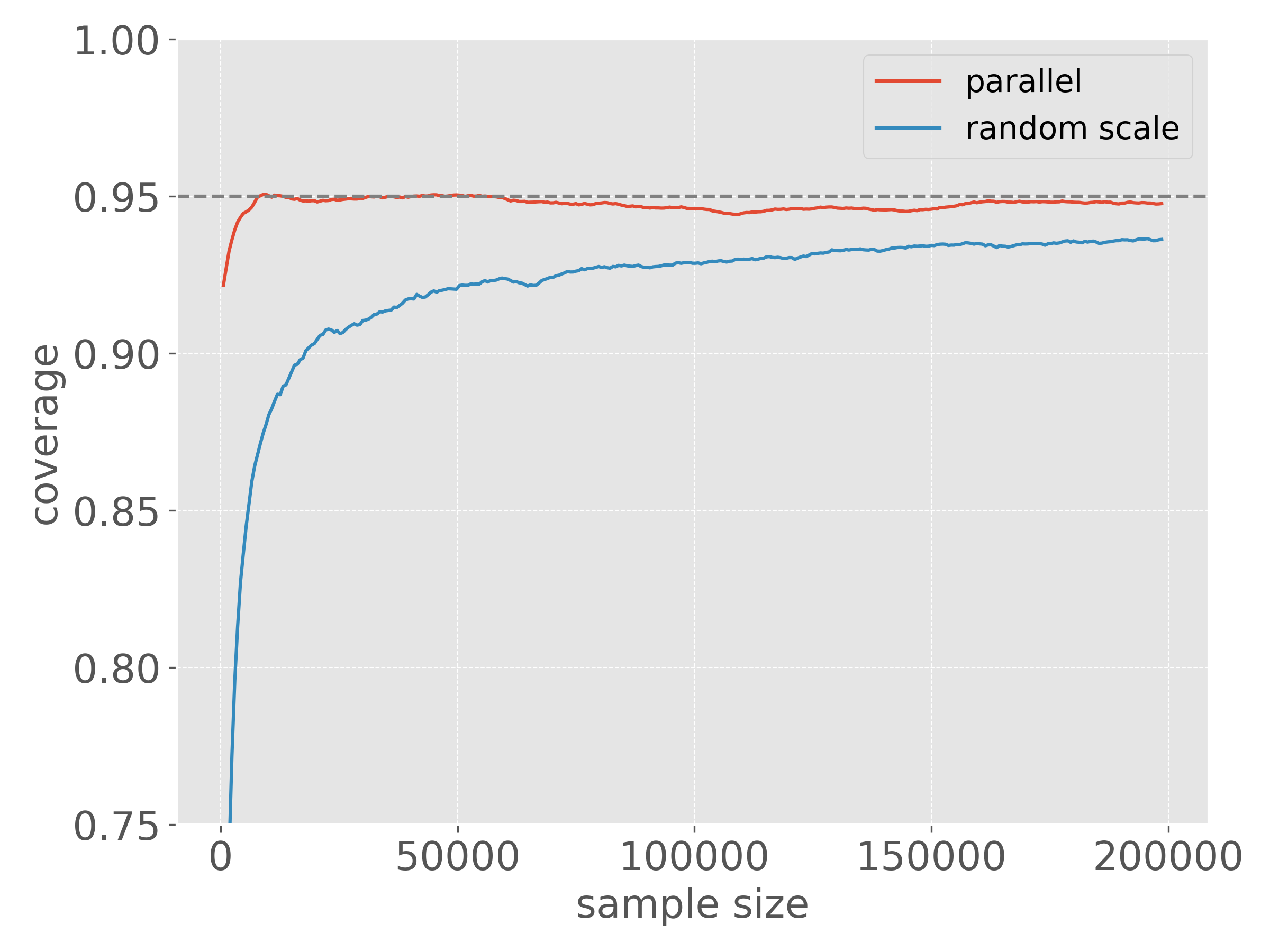} 
		\includegraphics[width=0.33\textwidth]{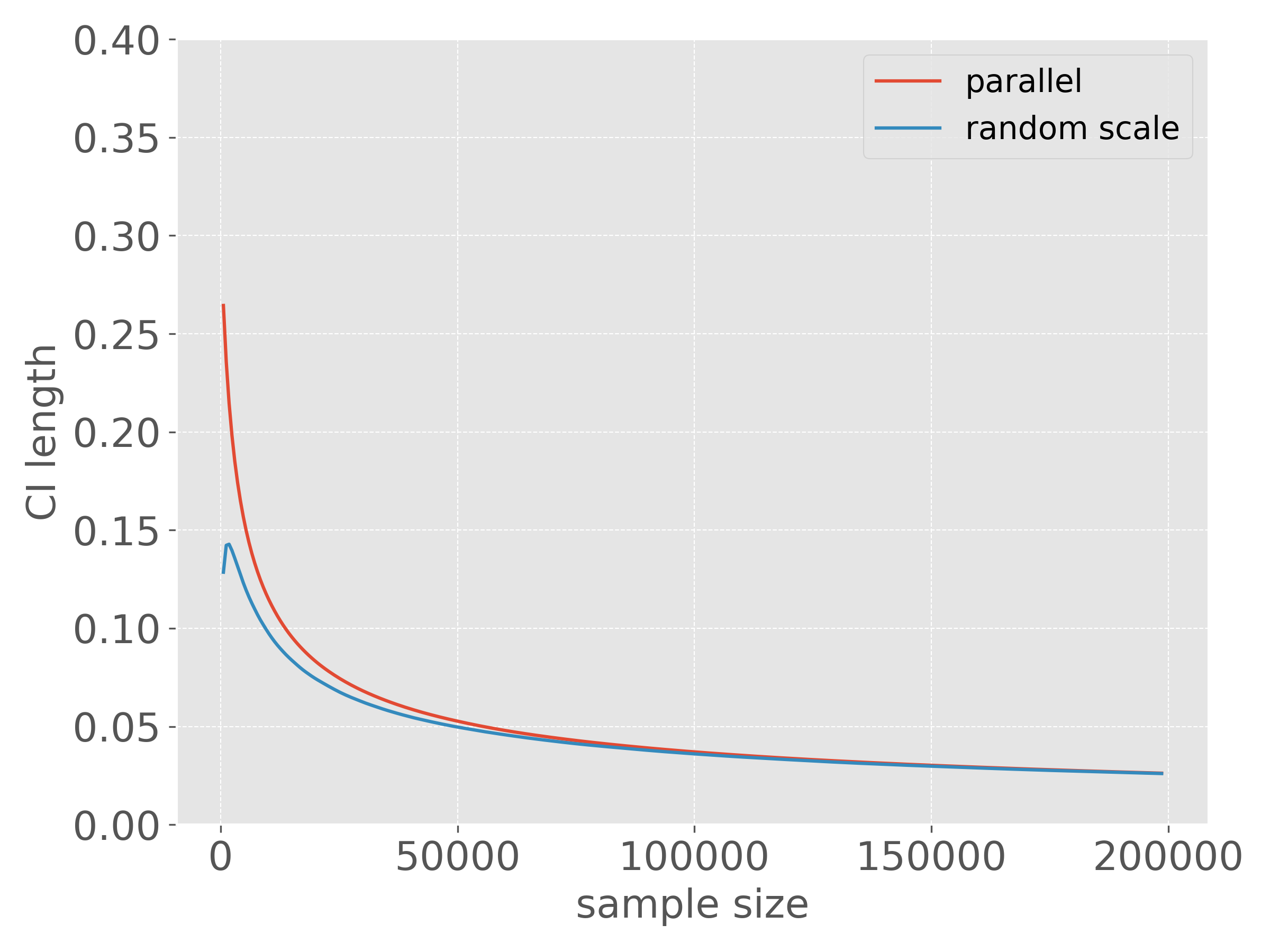}
	}
	\subfigure[$\alpha = 0.01$]{
		\includegraphics[width=0.33\textwidth]{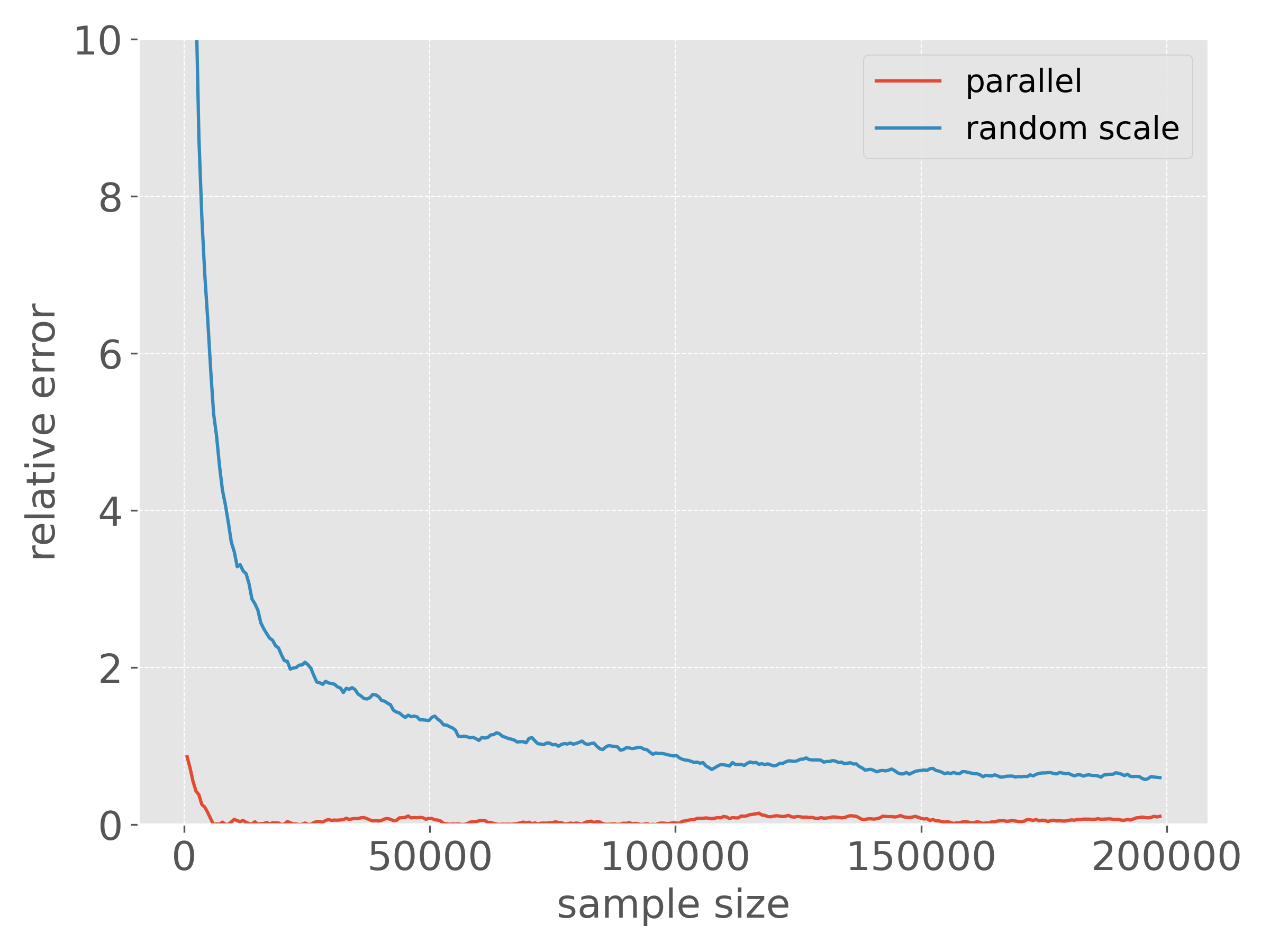} 
		\includegraphics[width=0.33\textwidth]{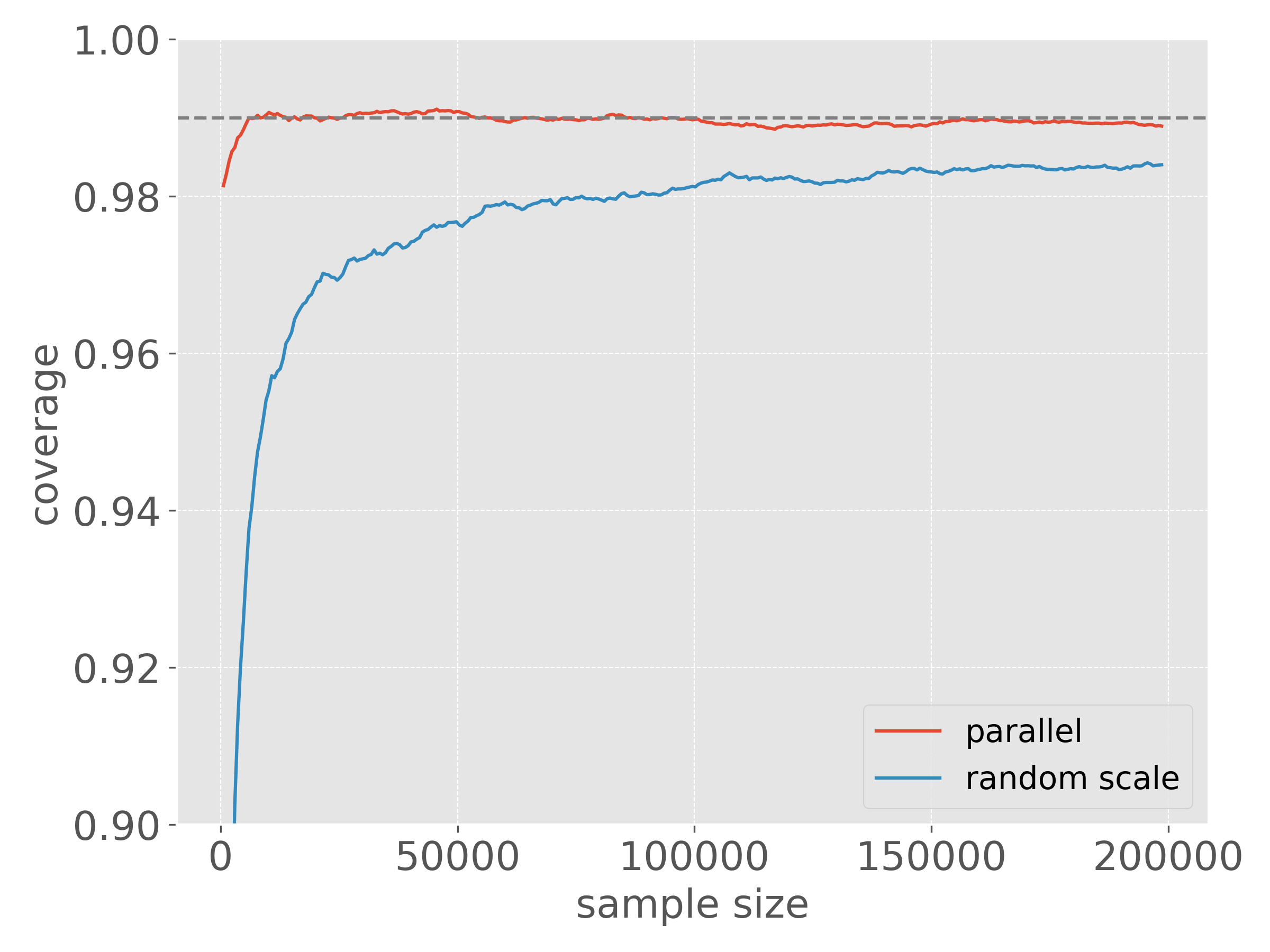} 
		\includegraphics[width=0.33\textwidth]{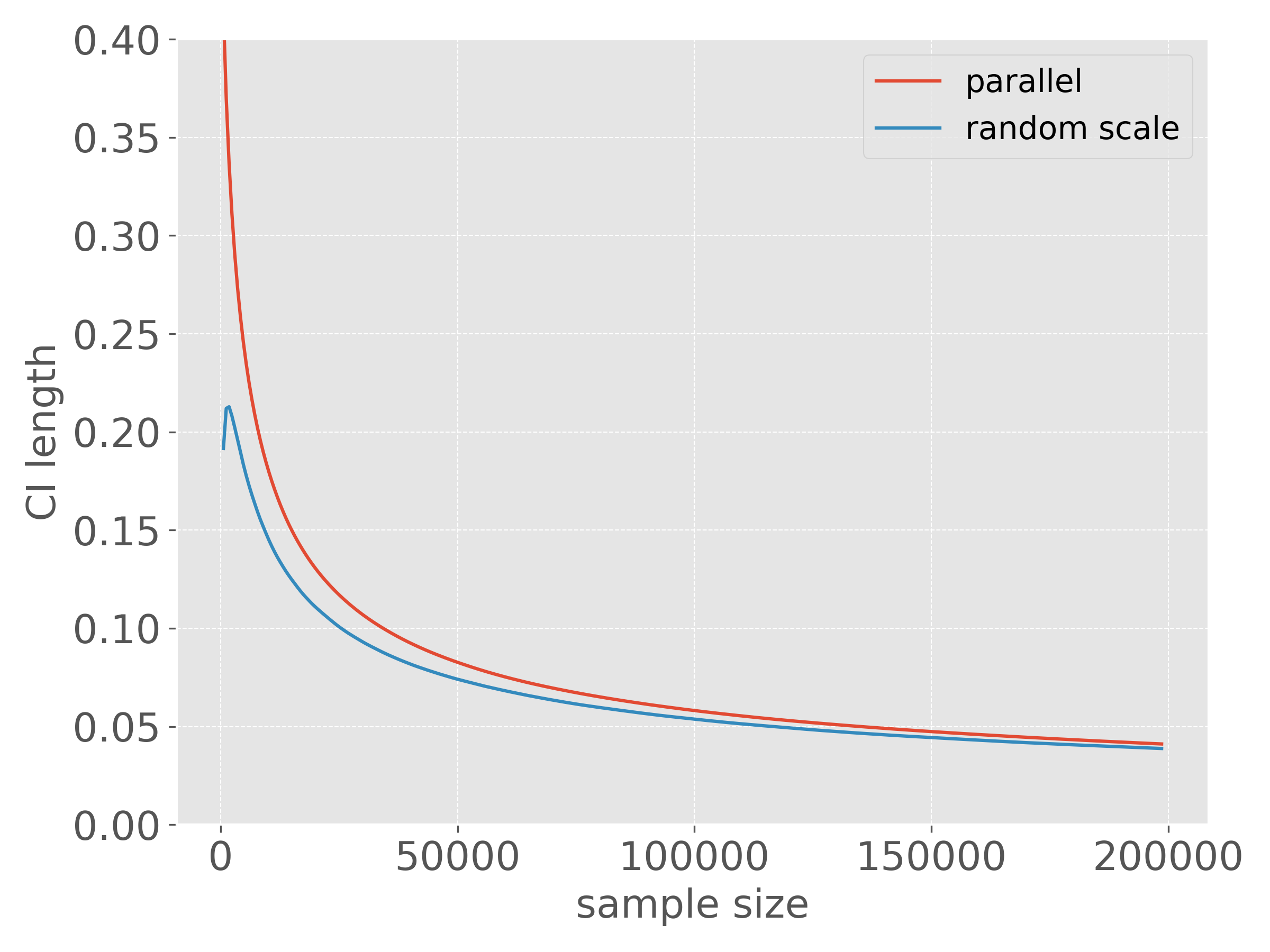}
	}
	\subfigure[$\alpha = 0.001$]{
		\includegraphics[width=0.33\textwidth]{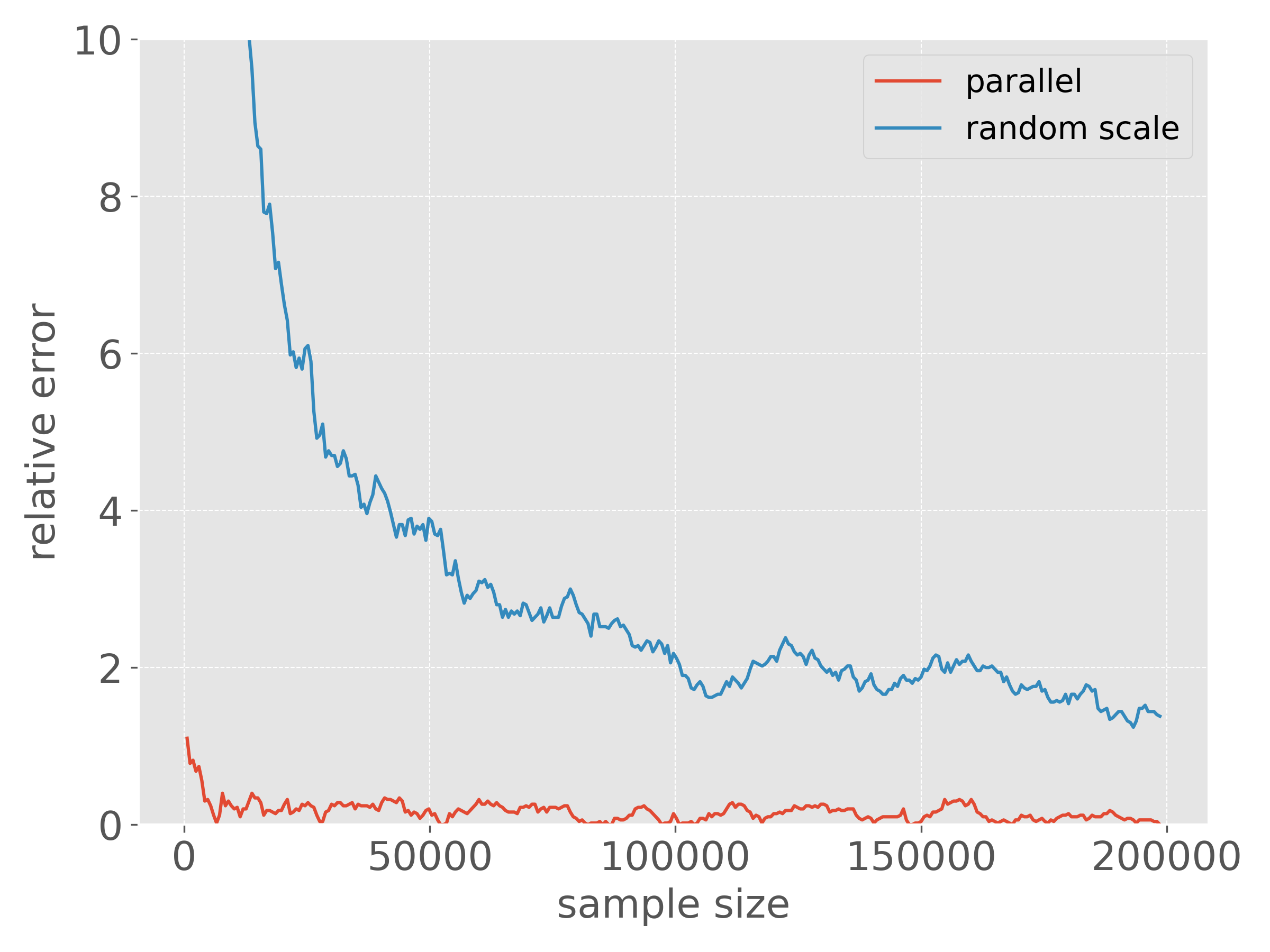} 
		\includegraphics[width=0.33\textwidth]{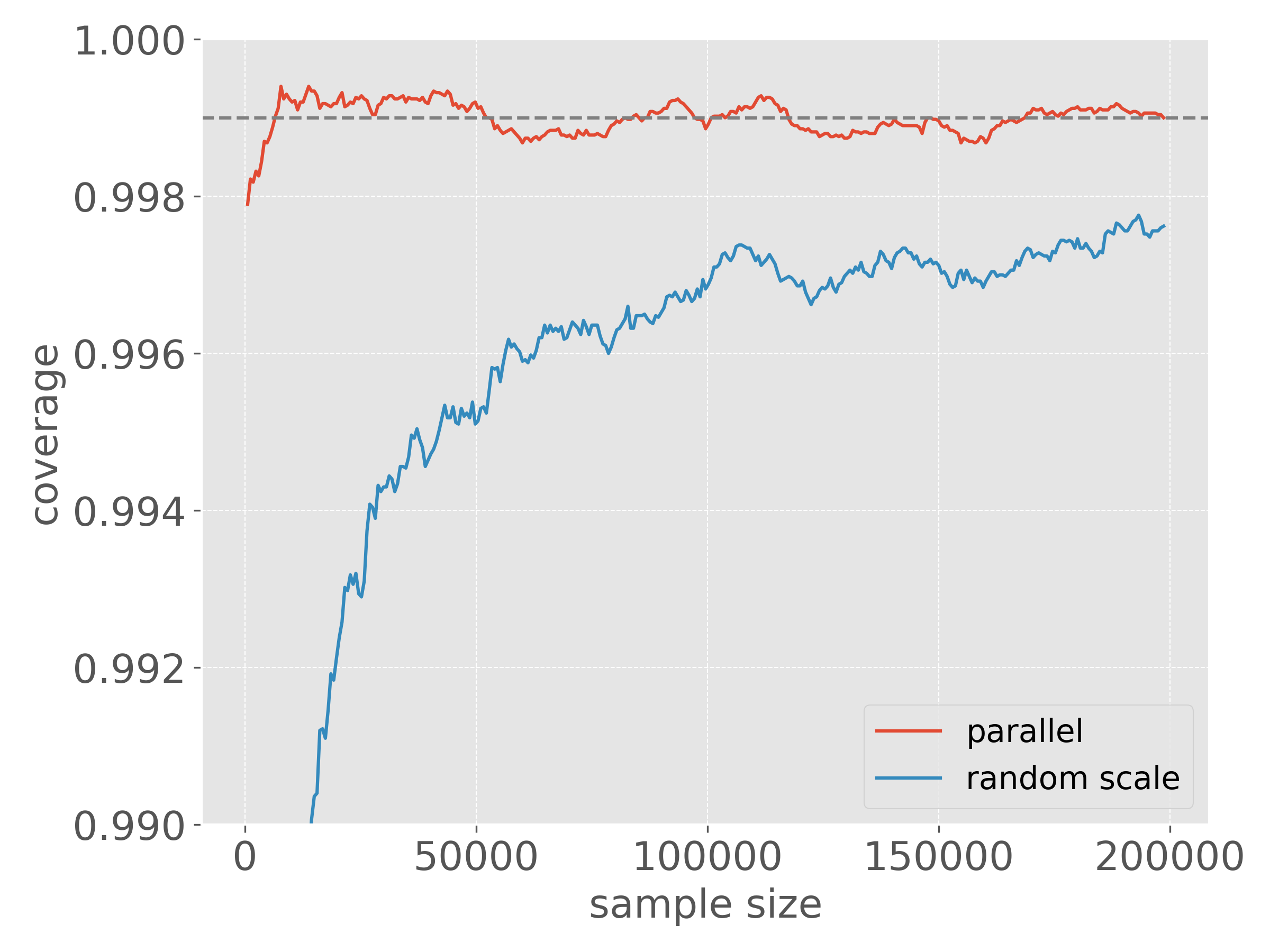} 
		\includegraphics[width=0.33\textwidth]{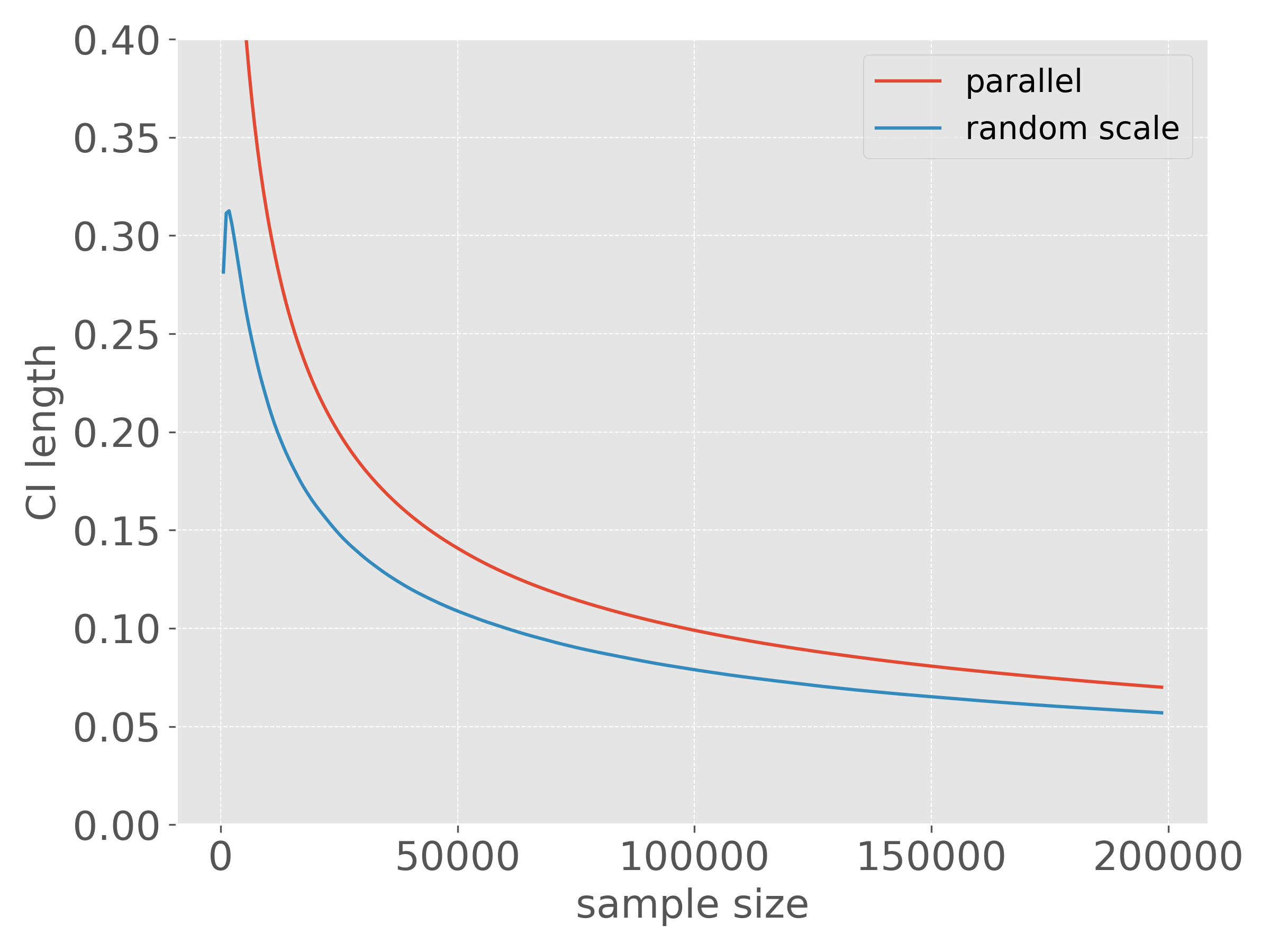}
	}
	\caption{Logistic Regression $d = 5$: Left: relative error of coverage; Middle: empirical coverage; Right: length of confidence intervals.}
	\label{fig:logistic_d5}
\end{figure}

\begin{figure}
	\subfigure[Linear, $d = 5$]{
		\includegraphics[width=0.45\textwidth]{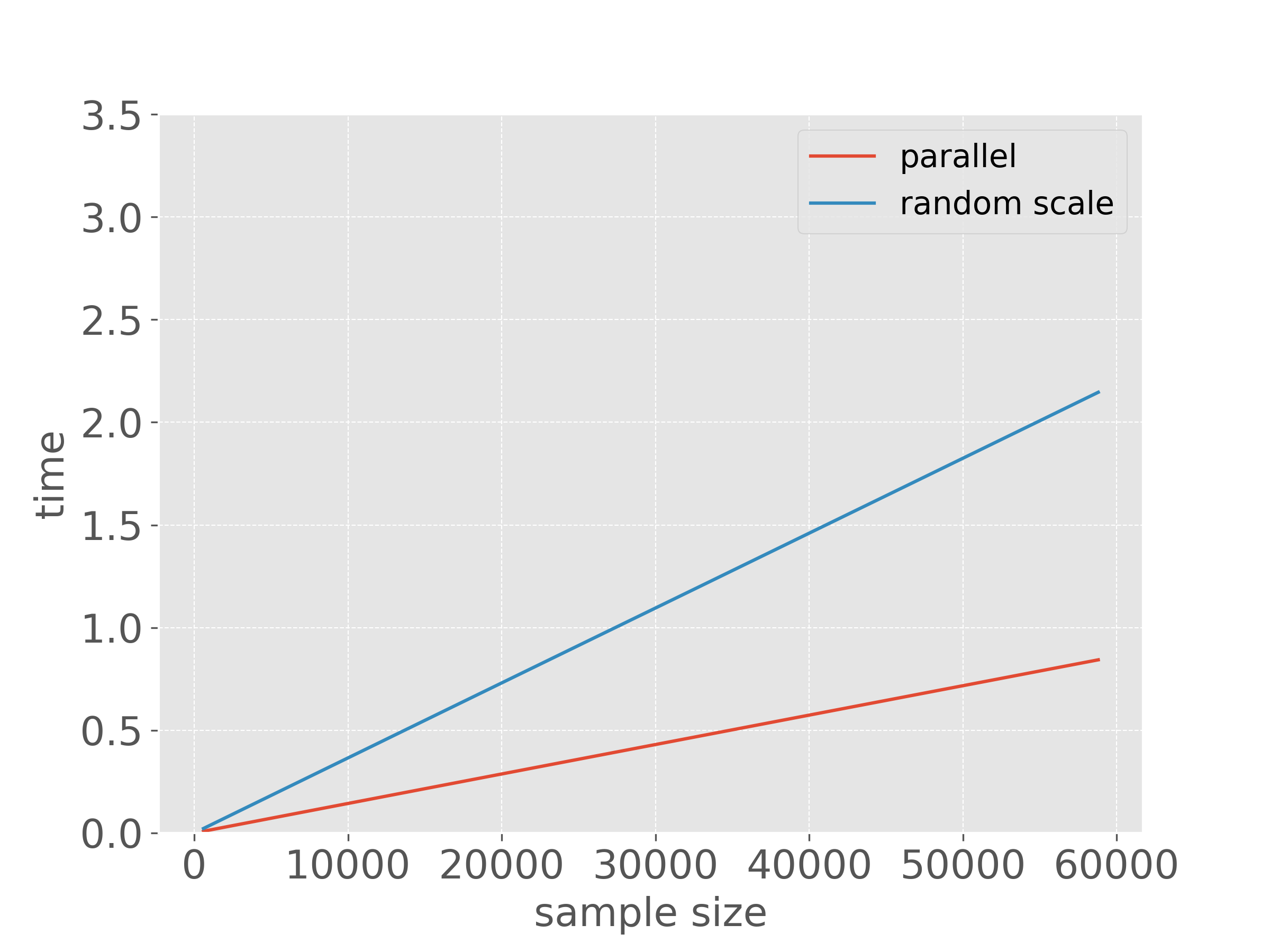}} 
	\subfigure[Logistic, $d = 5$]{
		\includegraphics[width=0.45\textwidth]{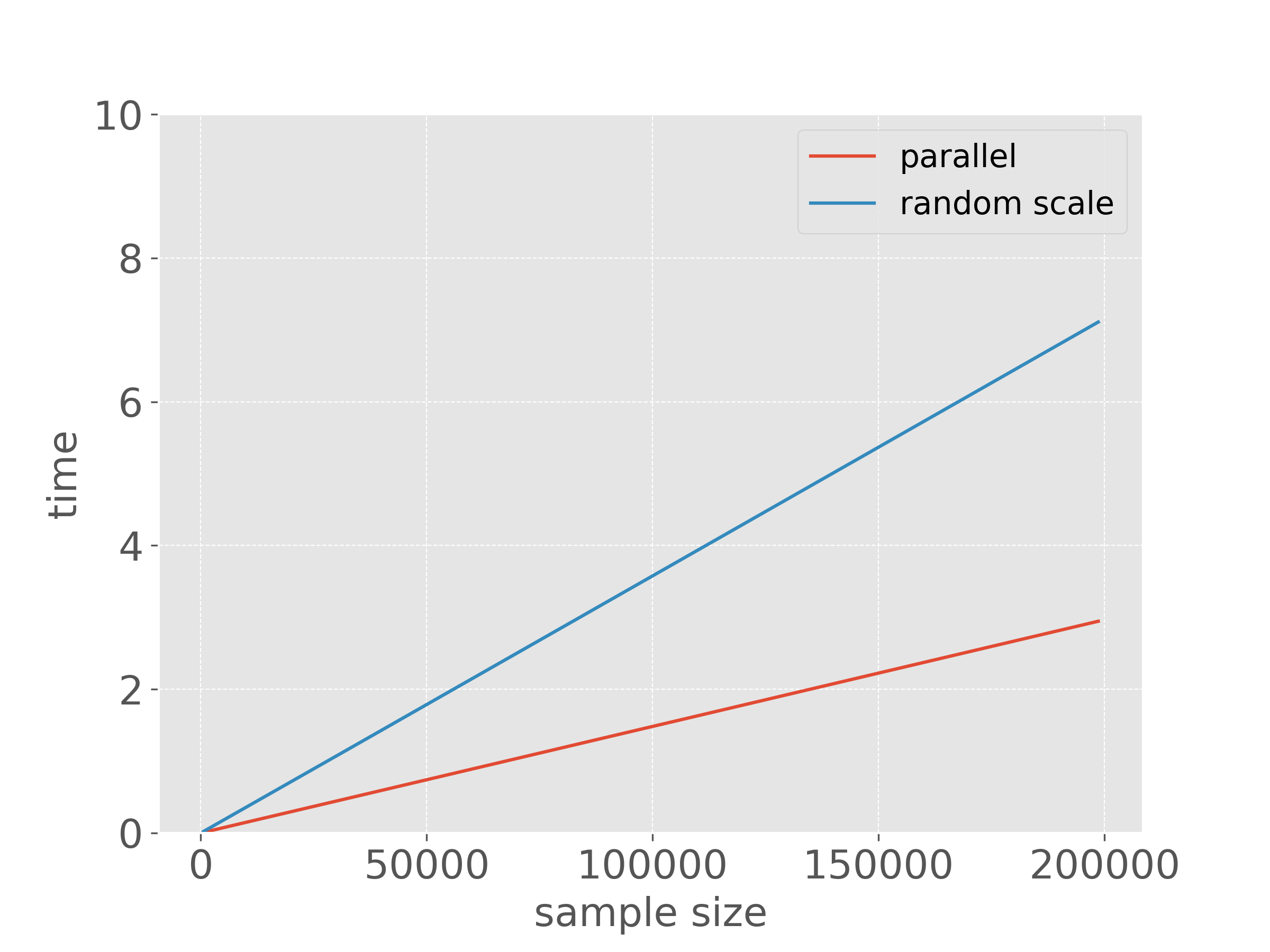}} 
	\caption{Computation time: d = 5}
	\label{fig:time_d5}
\end{figure}

In Table \ref{tb:cv}, we provide critical values used in the random scaling method. In Figure \ref{fig:linear_d5}, \ref{fig:logistic_d5}, \ref{fig:time_d5} and Table \ref{tb_d5}, we provide results for linear regression and logistic regression when $d = 5$ with same settings as described in Section \ref{sec:exp_convex}.

\begin{table}[ht]
	\centering
	\begin{tabular}{|c|c|c|c|}
		\hline
		Probability & 97.5\% & 99.5\%  &	99.95\% \\
		\hline
		Critical Value & 6.474   & 10.0544  & 14.76972 \\
		\hline
	\end{tabular}
	\caption{Asymptotic one-sided critical values for asymptotic pivotal statistic in the random scaling method \eqref{eqn:rs} via Monte Carlo simulation with 1,000,000 samples.}
	\label{tb:cv}
\end{table}

\begin{table}[ht]
	\vskip -0.1in
	\begin{center}
		\begin{small}
			\begin{sc}
				\begin{tabular}{lcccr}
					\midrule
					Linear  &  &     {\textbf{Coverage (Relative error)}}   &  {\textbf{Length (std)}}  & {\textbf{Time }}\\
					\midrule
					\multirow{2}{*}{\textbf{$\alpha = 0.05$}}   & Parallel &    0.9503 (0.006) & 0.021 (0.00300)& 0.918 \\ 
					& Random Scale  & 0.9454 (0.090)&0.021 (0.00384)& 2.262\\  
					\midrule
					\multirow{2}{*}{\textbf{$\alpha = 0.01$}} & Parallel &    0.9896 (0.042) & 0.032 (0.00467)& 0.918 \\ 
					& Random Scale  & 0.9885 (0.146)&0.031 (0.00572)& 2.262\\  
					\midrule
					\multirow{2}{*}{\textbf{$\alpha = 0.001$}}   & Parallel &    0.9991 (0.080) & 0.055 (0.00797)& 0.918 \\ 
					& Random Scale  & 0.9986 (0.420)&0.046 (0.00840)& 2.262\\  
					\midrule
				\end{tabular}
				\begin{tabular}{lcccr}
					\midrule
					Logistic &  &     {\textbf{Coverage (Relative error)}}   &  {\textbf{Length (std)}}  & {\textbf{Time }}\\
					\midrule
					\multirow{2}{*}{\textbf{$\alpha = 0.05$}}     & Parallel   &   0.9477 (0.046)& 0.026 (0.00383)&  2.861 \\ 
					& Random Scale & 0.9363 (0.274) &0.026 (0.00512)& 7.044   \\ 
					\midrule
					\multirow{2}{*}{\textbf{$\alpha = 0.01$}} & Parallel   &    0.9889 (0.106)& 0.041 (0.00601)&  2.861 \\ 
					& Random Scale & 0.9840 (0.598) &0.039 (0.00762)& 7.044   \\ 
					\midrule
					\multirow{2}{*}{\textbf{$\alpha = 0.001$}}  & Parallel   &    0.9990 (0.000)& 0.070 (0.01023)&  2.861 \\ 
					& Random Scale & 0.9976 (1.380) &0.057 (0.01120)& 7.044   \\ 
					\midrule
				\end{tabular}
			\end{sc}
		\end{small}
	\end{center} 
	\caption{Compare parallel method and random scaling method: $\alpha = \{0.001, 0.01, 0.05\}$ ($99.9\%$, $99 \%$ and $95\%$ confidence). Linear and Logistic regression $d = 5$.}
	\label{tb_d5}
\end{table}

\begin{table}[ht]
	\vskip -0.1in
	\begin{center}
		\begin{small}
			\begin{sc}
				\begin{tabular}{lcccr}
					\midrule
					Linear  &  &     {\textbf{Coverage (Relative error)}}   &  {\textbf{Length (std)}}  & {\textbf{Time }}\\
					\midrule
					\multirow{2}{*}{\textbf{$\alpha = 0.05$}}   & Parallel& 0.9509 (0.017)   & 0.022 (0.00218)&1.135  \\ 
					& Random Scale   & 0.9461 (0.078)&0.021 (0.00197)& 3.025\\  
					\midrule
					\multirow{2}{*}{\textbf{$\alpha = 0.01$}} & Parallel&    0.9903 (0.036) &0.034 (0.00340)&1.135 \\ 
					& Random Scale  & 0.9888 (0.124)&0.032 (0.00293)& 3.025\\  
					\midrule
					\multirow{2}{*}{\textbf{$\alpha = 0.001$}}   & Parallel  &    0.9990 (0.015) &0.058 (0.00563)&1.135 \\ 
					& Random Scale   & 0.9987 (0.310)&0.046 (0.00431)& 3.025\\  
					\midrule
				\end{tabular}
				\begin{tabular}{lcccr}
					\midrule
					Logistic &  &     {\textbf{Coverage (Relative error)}}   &  {\textbf{Length (std)}}  & {\textbf{Time }}\\
					\midrule
					\multirow{2}{*}{\textbf{$\alpha = 0.05$}}     & Parallel  &    0.9499 (0.024)&0.033 (0.00258)&  3.363\\ 
					& Random Scale & 0.9222 (0.554) & 0.032 (0.00337)& 9.384  \\ 
					\midrule
					\multirow{2}{*}{\textbf{$\alpha = 0.01$}} & Parallel  &    0.9898 (0.017)&0.051 (0.00405)&  3.363  \\ 
					& Random Scale  & 0.9784 (1.156) & 0.047 (0.00502)& 9.384   \\ 
					\midrule
					\multirow{2}{*}{\textbf{$\alpha = 0.001$}}  & Parallel  &    0.9991 (0.064)&0.088 (0.00689)&  3.363 \\ 
					& Random Scale  & 0.9960 (2.984) & 0.069 (0.00738)& 9.384 \\ 
					\midrule
				\end{tabular}
			\end{sc}
		\end{small}
	\end{center} 
	\caption{Compare parallel method and random scaling method: $\alpha = \{0.001, 0.01, 0.05\}$ ($99.9\%$, $99 \%$ and $95\%$ confidence). Linear and Logistic regression $d = 20$.}
	\label{tb_d20}
\end{table}

\end{document}